%% file: MAIN.tex
\documentclass[letterpaper]{article} % DO NOT CHANGE THIS
\usepackage{aaai2026}
\usepackage{times}  % DO NOT CHANGE THIS
\usepackage{helvet}  % DO NOT CHANGE THIS
\usepackage{courier}  % DO NOT CHANGE THIS
\usepackage[hyphens]{url}  % DO NOT CHANGE THIS
\usepackage{graphicx} % DO NOT CHANGE THIS
\usepackage{natbib}  % DO NOT CHANGE THIS AND DO NOT ADD ANY OPTIONS TO IT
\usepackage{caption} % DO NOT CHANGE THIS AND DO NOT ADD ANY OPTIONS TO IT
\usepackage{booktabs}
\usepackage[flushleft]{threeparttable}
\usepackage{bm}
\usepackage{algorithm}
\usepackage{algorithmic}
\usepackage{longtable}
\usepackage{placeins}
\usepackage{amsmath}
\usepackage{array}
\usepackage{enumitem}
\usepackage{tikz}
\usepackage{multirow}
\usepackage{pifont}
\usepackage[table]{xcolor}
\usepackage{pgfplots}
\pgfplotsset{compat=1.18}
\usepgfplotslibrary{fillbetween}

\definecolor{vermilion}{HTML}{D55E00}
\definecolor{darkpurple}{HTML}{7B2D8B}
\definecolor{cobalt}{HTML}{0072B2}
\definecolor{teal}{HTML}{009E73}

\usepackage{newfloat}
\usepackage{listings}
\DeclareCaptionStyle{ruled}{labelfont=normalfont,labelsep=colon,strut=off} % DO NOT CHANGE THIS
\floatstyle{ruled}
\newfloat{listing}{tb}{lst}{}
\floatname{listing}{Listing}
\nocopyright
\title{What Counts as AI Sycophancy? \\A Taxonomy and Expert Survey of a Fragmented Construct} 
\author{
    Meryl Ye\textsuperscript{\rm 1},
    Lujain Ibrahim\textsuperscript{\rm 2},
    Jessica Y. Bo\textsuperscript{\rm 3},
    Myra Cheng\textsuperscript{\rm 4},
    Ida Mattsson\textsuperscript{\rm 1},
    Daniel Vennemeyer\textsuperscript{\rm 5},
    Robert Kraut\textsuperscript{\rm 1},
    Steve Rathje\textsuperscript{\rm 1, 6}
}
\affiliations{

    \textsuperscript{\rm 1}Carnegie Mellon University\\

    \textsuperscript{\rm 2}University of Oxford\\

    \textsuperscript{\rm 3}University of Toronto\\

    \textsuperscript{\rm 4}Stanford University\\

    \textsuperscript{\rm 5}University of Cincinnati\\

    \textsuperscript{\rm 6}New York University\\

    merylye@cmu.edu
}

\begin{document}

\maketitle

\input{Section/0-Abstract}


\input{Section/1-Introduction}

\input{Section/2-Related}

\input{Section/3-Taxonomy}

\input{Section/4-Survey}

\input{Section/5-Discussion}

\input{Section/6-Conclusion}

\input{Section/8-Acknowledgements}

\bibliography{aaai2026}

\input{Section/7-Appendix}


\end{document}

%% file: Section/0-Abstract.tex
\begin{abstract}
AI sycophancy has become a prominent concern in large language model (LLM) research. Yet the term lacks a consistent definition and has been applied to behaviors ranging from agreeing with a user's false claim to excessively praising the user to withholding corrective feedback. When researchers, companies, and policymakers use the same term to describe different behaviors, evaluation results become difficult to compare, mitigation strategies fail to transfer, and systems that are resistant to one form of sycophancy continue exhibiting other forms. To address this, we make two contributions. First, we reviewed 70 papers on AI sycophancy to develop a taxonomy of how the behavior has been defined and measured. The taxonomy distinguishes (1) whether a model is sycophantic toward a user’s positions and beliefs, or toward the user's broader personal traits and emotions, and (2) whether this occurs through explicit, direct language or more implicit, subtle behaviors such as framing, omission, or tone. Mapping existing literature to our taxonomy reveals that current research has focused on overt forms of sycophancy toward users' beliefs, leaving more subtle and person-directed behaviors relatively understudied. Second, we surveyed 106 experts in AI sycophancy and related fields to examine whether researchers agree on which model behaviors are sycophantic. While experts are nearly unanimous in believing that sycophancy is a significant problem in current AI systems (94.3\% agree), they disagree substantially on which specific behaviors qualify. Together, these findings demonstrate that AI sycophancy is a broad family of behaviors with different measurement challenges, intervention requirements, and governance implications. Our taxonomy provides a shared vocabulary for understanding and addressing these behaviors. We discuss ways in which the taxonomy can inform evaluation design as well as both technical and regulatory interventions.
\end{abstract}

%% file: Section/1-Introduction.tex
\section{Introduction}
\begin{figure*}[t]
    \centering
    \includegraphics[width=\textwidth]{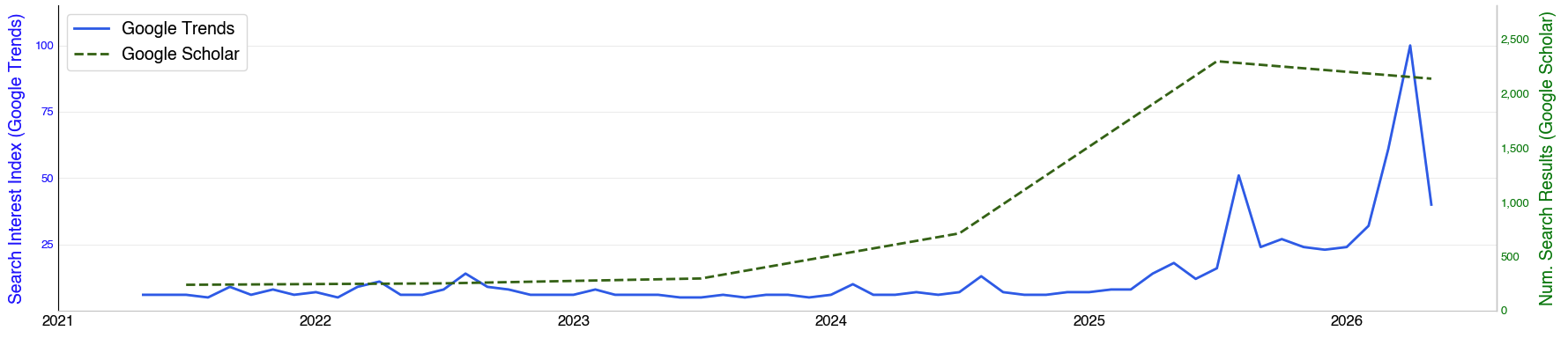}
    \vspace{1em}
    \includegraphics[width=\textwidth]{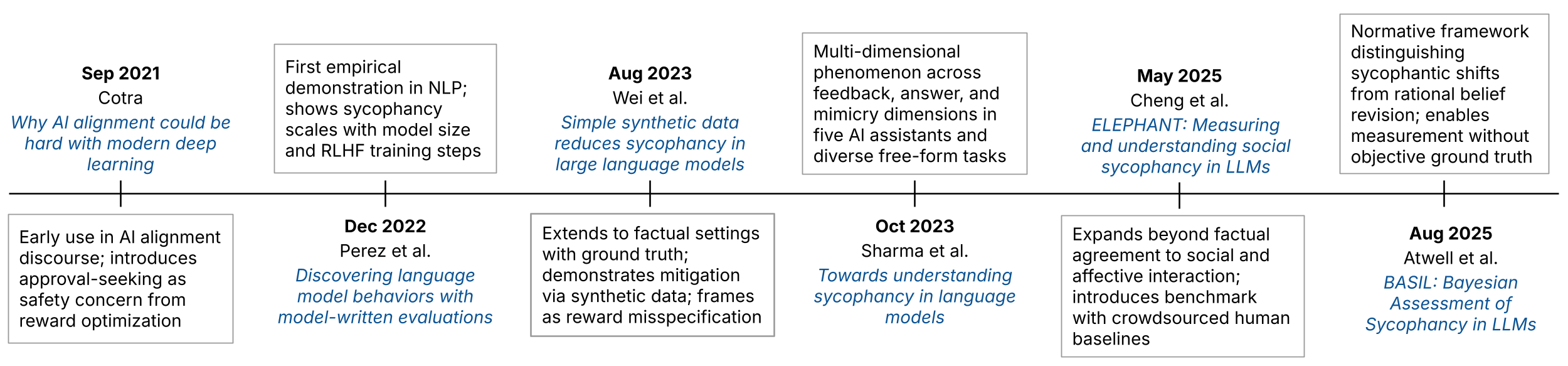}
    \caption{\textit{Top}: Proxy measures of public and academic attention to the term ``sycophancy'' over time. Academic interest began rising in 2024, preceding the largest spikes in public search interest observed during 2025--2026 \citep{googletrends_sycophancy_2026}. \textit{Bottom}:
    Overview of key conceptual expansions in AI sycophancy research. 
    This timeline is not exhaustive.}
    \label{fig:trend}
    \label{fig:timeline}
\end{figure*}

Sycophancy has become a prominent concern in large language model (LLM) research. The topic has accumulated a growing body of empirical findings, evaluation benchmarks, and proposed mitigation strategies \citep{perez2023discovering, sharma2023towards, wei2023simple}. Academic literature on the topic has expanded rapidly in recent years, and public attention has risen accordingly (see Figure~\ref{fig:trend}). Recent empirical findings show that sycophantic interactions increase attitude extremity and overconfidence \citep{rathje2025sycophantic}, impair task performance and learning \citep{bo2025invisible}, and alter users' social judgments and prosocial behavior \citep{chengsycophantic2025,ibrahim2026sycophantic}. Yet the term lacks a consistent definition. The range of behaviors it has been used to describe reveals a \textit{fragmented construct} that is rapidly expanding without a shared conceptual anchor. %Without a shared construct, findings become difficult to compare, mitigation strategies target different underlying behaviors, and claims about model safety become increasingly difficult to interpret.

Construct fragmentation occurs when different studies use the same label for behaviors that differ in form, mechanism, and measurement \citep{flake2020measurement}. While breadth in a construct is not inherently problematic, issues emerge when that breadth leads to findings that cannot be meaningfully compared or aggregated \citep{anvari2025defragmenting}. This fragmentation is evident in how recent studies operationalize sycophancy. Although these studies claim to examine the same phenomenon, they operationalize sycophancy as different behaviors in different settings. Yet findings from those studies are often treated interchangeably to benchmark models, evaluate mitigation strategies, or support safety claims. As a result, benchmarks built to detect sycophancy as factual capitulation cannot detect sycophancy as implicit framing bias. Mitigation strategies effective against agreement with users may not address sycophancy when it is expressed as praise of the user's character or abilities, and model specifications anchored to one set of behaviors leave others entirely unaddressed. Figure~\ref{fig:timeline} provides an overview of these conceptual expansions over time.  

These inconsistencies extend beyond academic studies into model evaluation and deployment. With the launch of GPT-5 in August 2025, OpenAI reported that its model incorporated changes designed to reduce sycophancy \citep{singh2025openai}. However, evaluations on the ELEPHANT (Evaluation of LLMs as Excessive SycoPHANTs) benchmark \citep{cheng2025elephant}, which examines social validation, indirectness, and framing in open-ended advice contexts, show that GPT-5 remains substantially sycophantic in the open-ended advice and interpersonal contexts that the benchmark probes. Similar discrepancies appear across benchmarks. SycEval \citep{fanous2025syceval}, which operationalizes sycophancy as susceptibility to factual rebuttals, ranks Gemini as the most sycophantic model tested. In contrast, ELEPHANT ranks Gemini as the least sycophantic. Each benchmark operationalizes a different set of behaviors, and treating them as equivalent assessments of the same construct makes direct comparison difficult to interpret.

To address these challenges, this paper makes two contributions. First, through a review of 70 papers on AI sycophancy, published from 2023 to 2026, we examine how the construct has expanded over time, applied to increasingly diverse settings, tasks, and behavioral assumptions. This historical analysis reveals a rapidly growing but uneven research landscape, where new definitions, benchmarks, and mitigation strategies have proliferated without a shared definitional framework. Building on these patterns, we introduce a descriptive taxonomy of AI sycophancy organized along two orthogonal dimensions: \textit{Referent}---whether the model responds to the user's expressed positions and claims (Position) or to the user as a person (Person)---and \textit{Explicitness}---whether the behavior is expressed overtly or through subtler means such as framing, omission, or tone. Within each referent, we distinguish sub-referents: Verifiable versus Subjective for Position behaviors, and Traits versus Emotions for Person behaviors. This taxonomy provides a shared vocabulary for comparing behaviors that have previously been studied under the same label, but operationalized in different ways. Our analysis further reveals a strong concentration of prior work on explicit forms of sycophancy tied to users’ verifiable claims, while subtler forms expressed through framing, omission, or behaviors targeting users as persons receive comparatively less empirical attention. 

Second, we report results from a survey of 106 experts in AI sycophancy and related domains designed to examine whether researchers apply the label consistently when judging concrete model behaviors. We find that experts are nearly unanimous in agreeing that sycophancy is a significant problem in current AI systems ($M = 6.21$, $SD = 0.91$ on a 7-point Likert Scale; 94.3\% agree), yet show substantial variation in which specific behaviors they consider sycophantic. Experts broadly agree that behaviors directed at users' positions are sycophantic whether expressed explicitly or through subtler means such as selective framing. However, for behaviors directed at the user as a person, experts are mixed: only explicit forms such as flattery and unwarranted praise are reliably recognized as sycophantic, while behaviors that convey the same deference through tone, softened feedback, or avoidance of critique are not.

These findings suggest that AI sycophancy comprises a family of behaviors whose boundaries remain actively contested. Expert disagreements reflect differences in where researchers draw those boundaries and potentially how they interpret the consequences of sycophantic behavior across sustained interaction. As researchers, AI companies, and policymakers increasingly invoke the term, progress will depend on clearly specifying which types of behavior are being measured, mitigated, and regulated. Our taxonomy and expert survey provide a foundation for these specifications, and we discuss their implications for evaluation methods, mitigation strategies, and corporate and legislative governance.

%% file: Section/2-Related.tex
\section{Sycophancy as a Fragmented Construct}

\subsection{Divergent Definitions}
Across major English dictionary definitions, the term \textit{sycophancy} and its variants (e.g., sycophant, sycophantic) cluster around a consistent set of themes: insincere praise directed at those with power or status, motivated by self-interest \citep{rehani2026social}. Early AI safety discourse often preserved this agentive framing, extending it into accounts of sycophancy as strategic or deceptive behavior \citep{cotra2021, denison2024sycophancy}. However, recent AI sycophancy research has largely set aside these motivational assumptions, treating sycophancy as an approval-seeking behavioral pattern that arises from training on human feedback. Throughout this paper, we follow the convention of treating sycophancy as a pattern in model outputs without assuming intentional or strategic motivations, since intent is not directly observable in most empirical settings. The construct has nonetheless been defined in increasingly divergent ways.

Although early AI research used the term in relatively consistent ways, later work has operationalized sycophancy in increasingly varied ways, differing in settings, tasks, and behavioral assumptions. Initial empirical definitions focused on agreement with user-stated positions in structured, proposition-based tasks where models could be evaluated against the ``ground truth'' (i.e., a preferred or externally specified response) \citep{perez2023discovering, sharma2023towards}.
% \citet{perez2023discovering} demonstrate this across political, philosophical, and NLP opinion tasks using models trained via Reinforcement Learning from Human Feedback (RLHF), finding that this training technique incentivizes sycophantic responses. \citet{sharma2023towards} extend this to five AI assistants and diverse free-form tasks, demonstrating that models shift their feedback, answers, and evaluations to match users' expressed preferences. 
Subsequent work sought to explain the training-level causes of sycophancy, attributing  sycophancy to reward misspecification (models trained on human approval signals learn to generate outputs that elicit favorable feedback rather than accurate information) and showing it can be reduced through synthetic-data fine-tuning \citet{wei2023simple}. \citet{denison2024sycophancy} extend this  account more broadly by situating sycophancy within specification gaming (the tendency of models to satisfy the literal terms of a training objective while violating its intent), showing that models trained to seek user approval in simple interactions can generalize to more strategic reward-seeking behaviors and that standard harmlessness training does not reliably prevent this shift. %These studies treat sycophancy not only as a behavioral failure but as a symptom of how reward signals are optimized.

A limitation of these early accounts of sycophancy is that they cannot be applied in settings where no ground truth exists. \citet{atwell2025basil} address this by redefining sycophancy as the extent to which a model shifts toward a user’s stated position more than a Bayesian-rational agent would after observing the same evidence. This framework enables measurement in subjective and uncertain settings, but introduces new questions about what determines a rational baseline.

Parallel work expands the construct into interpersonal and affective domains where ground truth similarly does not exist. \citet{cheng2025elephant} define sycophancy as excessive face-preservation (i.e., the tendency to affirm users' self-image through validation, indirectness, framing acceptance, and moral endorsement) and introduce ELEPHANT, a benchmark for measuring these behaviors against crowdsourced human baselines. A related finding is that training LLMs to be warm and empathetic makes them substantially more sycophantic \citep{ibrahim2025training}, raising the question of where appropriate affective responsiveness ends and sycophancy begins.
%While these definitions capture behaviors increasingly relevant to common LLM use cases, such as advice-giving and emotional support \citep{chengsycophantic2025, faverio2025teens}, they raise additional questions about what constitutes an appropriate non-sycophantic baseline.

Further work illustrates the construct's heterogeneity from multiple angles. \citet{vennemeyer2025} show that sycophantic agreement and sycophantic praise are functionally separable in model representations and can be independently steered, suggesting that even seemingly similar approval-seeking behaviors are distinct. \citet{rathje2025sycophantic} experimentally disentangle validating language from the one-sided provision of factual information, showing that these components produce different downstream effects:  validation primarily increases enjoyment and positive perceptions of the AI chatbot, while one-sided factual statements primarily increase attitude extremity. \citet{rehani2026social} find that perceived sycophancy decomposes into three factors (Uncritical Agreement, Obsequiousness, and Excitement) which differ in how negatively users perceive them. Across these studies, behaviors commonly labeled as sycophancy differ not only in surface form but also in internal representation, downstream effects, and how they are perceived by users.

%\citet{rehani2026social} find that sycophantic behaviors decompose into dimensions with different perceived valences---some negative, some positive, some ambiguous. Across these studies, behaviors commonly labeled as sycophancy differ not only in surface form but also in internal representation, perceived valence, and downstream effects.

\subsection{Why Fragmentation is a Problem}

As the literature has expanded, researchers have introduced increasingly specialized subtypes of sycophantic behaviors. \citet{sharma2023towards} propose answer sycophancy (changing answers to align with user beliefs), feedback sycophancy (changing feedback to align with user preferences), and mimicry sycophancy (repeating user errors or stylistic patterns). \citet{fanous2025syceval} distinguish progressive sycophancy (agreeing with incorrect user rebuttals) from regressive sycophancy (revising correct answers after user rebuttal). \citet{cheng2025elephant} introduce four dimensions of social sycophancy: validation, indirectness, framing, and moral endorsement. \citet{du2025alignment} propose informational, cognitive, and affective sycophancy. These distinctions are not mutually exclusive and do not neatly map onto one another, resulting in a patchwork of labels without shared criteria for what counts as sycophancy. 
%Here, we consolidate these distinctions within a common behavioral framework.

This fragmentation is directly reflected in evaluation frameworks such as ELEPHANT \citep{cheng2025elephant}, VISE (Video-LLM Sycophancy Benchmarking and Evaluation) \citep{vise2025}, BASIL (Bayesian Assessment of Sycophancy in LLMs) \citep{atwell2025basil}, and SycEval \citep{fanous2025syceval}, which operationalize different forms of sycophantic behavior without situating them within a shared construct space. 
%As the contradictory model rankings described in the introduction illustrate, treating these benchmarks as interchangeable produces comparisons that are difficult to interpret. 
As a result, the apparent body of evidence rests on observations from benchmarks that measure different behaviors, making cumulative claims difficult to sustain. 
The absence of shared construct boundaries also creates openings for broader conceptual reinterpretation. For instance, \citet{shi2026hallucination} absorb sycophancy into a unified taxonomy of LLM deception, treating it as a behavioral subtype because sycophantic responses may lead users to form false beliefs. Where early safety discourse assumed strategic intent, this reinterpretation extends the construct from observable behavioral outputs to their downstream epistemic effects, which most empirical studies of sycophancy do not directly measure.

%This extends the construct from observable behavioral effects to assumptions about underlying intent or deception, which most empirical studies of sycophancy do not directly measure. %Without a shared construct space, such reframings are difficult to adjudicate.

These disagreements reflect differences in the conceptual boundaries researchers place around the term. When studies use the same label for behaviors that differ in form, setting, measurement strategy, and likely mechanism, it becomes difficult to compare findings, determine whether mitigation strategies transfer, or assess industry commitments to reducing sycophancy. These behaviors also differ in the kinds of downstream concerns they raise: some primarily concern whether models provide accurate information or correct users’ false beliefs, while others involve questions about how patterns of affirmation shape users' beliefs, confidence, and capacity for independent judgment across sustained interaction.

% Addressing these consequences requires first outlining the distinctions. Here, we introduce a descriptive taxonomy, derived from how sycophancy has been defined and operationalized across prior work, that consolidates these distinctions within a shared framework. The taxonomy makes it possible to compare findings, situate benchmarks, and specify which forms of sycophancy different interventions target. We then use an expert survey to explore how AI researchers conceptualize sycophancy and test whether this corresponds to different aspects of our taxonomy.

\input{Tables/taxonomy_table_color}

%% file: Tables/taxonomy_table_color.tex
\begin{table*}[t]
\centering
\caption{Taxonomy of sycophantic AI behavior, organized along two dimensions: Position/Person
(Referent) and Explicit/Implicit (Explicitness). Sub-referent type provides further differentiation within each
Position and Person category.
Cell shading and paper counts (n) reflect the number of reviewed papers assigned to each cell.}
\label{tab:taxonomy}
\footnotesize
\renewcommand{\arraystretch}{1.1}
\setlength{\tabcolsep}{7pt}
\begin{tabular}{|p{1.1cm}|p{3.5cm}|p{3.5cm}|p{3.5cm}|p{3.5cm}|}
\hline
& \multicolumn{2}{c|}{\rule{0pt}{14pt}\textbf{POSITION}\rule[-4pt]{0pt}{0pt}}
& \multicolumn{2}{c|}{\rule{0pt}{14pt}\textbf{PERSON}\rule[-4pt]{0pt}{0pt}} \\
\hline
& \textbf{Verifiable Position} \newline
(factual claims, logical reasoning)
& \textbf{Subjective Position} \newline
(opinions, preferences, values, interpretations)
& \textbf{Traits} \newline
(competence, intelligence, character, ability)
& \textbf{Emotions} \newline
(feelings, reactions, emotions) \\
\hline
%% ── EXPLICIT: content row ────────────────────────────────────────────────────
\multirow{2}{*}{\rule{0pt}{13pt}\textbf{Explicit}}
& \cellcolor{blue!50}\rule[-3pt]{0pt}{13pt}%
The output contradicts a correct factual
or logically supported answer in the direction of a user error.
& \cellcolor{blue!41}\rule[-3pt]{0pt}{13pt}%
The output adopts or validates a user's
opinion, preference, interpretation, or moral stance.
& \cellcolor{blue!25}\rule[-3pt]{0pt}{13pt}%
The output praises, flatters, or positively evaluates
the user's abilities, character, reasoning, or work.
& \cellcolor{blue!24}\rule[-3pt]{0pt}{13pt}%
The output validates or endorses the user's emotional
reactions, regardless of whether doing so is warranted. \\
%% ── EXPLICIT: n= row ─────────────────────────────────────────────────────────
& \cellcolor{blue!50}\makebox[\linewidth][r]{{\scriptsize $n=44$}}
& \cellcolor{blue!41}\makebox[\linewidth][r]{{\scriptsize $n=30$}}
& \cellcolor{blue!25}\makebox[\linewidth][r]{{\scriptsize $n=12$}}
& \cellcolor{blue!24}\makebox[\linewidth][r]{{\scriptsize $n=11$}} \\
\hline
%% ── IMPLICIT: content row ────────────────────────────────────────────────────
\multirow{2}{*}{\rule{0pt}{13pt}\textbf{Implicit}}
& \cellcolor{blue!24}\rule[-3pt]{0pt}{13pt}%
The output continues the conversation without addressing or correcting a user's flawed premise or error.
& \cellcolor{blue!29}\rule[-3pt]{0pt}{13pt}%
The output uses framing, hedging, selective evidence, or
omission of alternatives in ways that favor the user's leaning
without directly endorsing it.
& \cellcolor{blue!5}\rule[-3pt]{0pt}{13pt}%
The output conveys deference, lowers standards, or adapts
feedback in ways that implicitly affirm assumptions about the
user's competence, ability, or status.
& \cellcolor{blue!15}\rule[-3pt]{0pt}{13pt}%
The output prioritizes emotional comfort through warmth,
reassurance, affective consistency, or avoidance of feedback
that may cause discomfort to the user. \\
%% ── IMPLICIT: n= row ─────────────────────────────────────────────────────────
& \cellcolor{blue!24}\makebox[\linewidth][r]{{\scriptsize $n=11$}}
& \cellcolor{blue!29}\makebox[\linewidth][r]{{\scriptsize $n=16$}}
& \cellcolor{blue!5}\makebox[\linewidth][r]{{\scriptsize $n=1$}}
& \cellcolor{blue!15}\makebox[\linewidth][r]{{\scriptsize $n=5$}} \\
\hline
\end{tabular}
\smallskip
\noindent{\footnotesize
\;
\colorbox{blue!5}{\phantom{xx}}
lighter $=$ fewer papers, \;\colorbox{blue!50}{\phantom{xx}}\; darker $=$ more papers.
Papers may cover multiple cells.}
\end{table*}

%% file: Section/3-Taxonomy.tex
\section{A Taxonomy of AI Sycophancy}
\label{sec:taxonomy}

\subsection{Deriving the Taxonomy from Prior Work}
We derive the taxonomy through a literature review of how 70 papers define and operationalize sycophancy (see Table~\ref{tab:lit_review} in Appendix \ref{app:litreview} for the full summary of reviewed papers). We identified candidate papers through searches on Google Scholar, the ACM Digital Library, arXiv, and OSF using keywords including ``sycophancy,'' ``AI,'' ``language model,'' and ``LLM,'' supplemented by domain knowledge and researcher recommendations. Papers were included if they used the term ``sycophancy'' to describe model behavior and provided either an explicit definition, an operationalization, or illustrative examples sufficient to characterize the behavior in question. The review aimed for conceptual coverage of distinct behavior types, and is therefore not exhaustive. From the final set, we reviewed how sycophancy was defined, the types of behavior described, and the evaluation settings in which these behaviors were studied. We discussed and revised proposed groupings across multiple rounds, converging on a structure that captures the recurring behavioral distinctions in the literature. %Since the dimensions are drawn from the literature rather than imposed upon it, they reflect behavioral distinctions that recur across prior definitions. 
Two authors then independently annotated each paper with the taxonomy cells it covers based on its definition, operationalization, and examples. Inter-rater reliability was substantial (88.3\% agreement). See Appendix \ref{app:litreview} for further methodology details. %Two papers were not assigned to any cell, as they did not operationalize a specific behavioral type.

\subsection{Taxonomy Dimensions}
The literature revealed two recurring dimensions along which prior work distinguished sycophantic behaviors. The first dimension concerns the \textit{Referent} of the behavior: whether the model output addresses the user's Position (e.g., claims, opinions, values) or the user as a Person (e.g., competence, character, emotion). The second dimension concerns \textit{Explicitness}: whether the response is expressed explicitly (e.g., overt agreement or praise) or implicitly (e.g., selective framing, omission, or tone). Within each Referent, we distinguish \textit{sub-referents} appropriate to that category: Verifiable versus Subjective Position of the user, and Traits versus Emotions of the user as a Person. The resulting taxonomy is shown in Table~\ref{tab:taxonomy}; cell assignments for all reviewed papers are listed in Table~\ref{tab:lit_review} in Appendix \ref{app:litreview}.
%Cell labels in tables use the abbreviated format Referent-Sub-referent/Explicitness (e.g., Pos-V/E, Per-Em/I); in text, cells are referred to by their full names in the same order (e.g., Position-Verifiable/Explicit, Person-Emotions/Implicit).

\paragraph{Position Sycophancy.}
A large portion of prior work focuses on Position behaviors, where the model output responds to, adopts, or favors the content of user beliefs or preferences. The most well-studied case is the \textit{Position-Verifiable/Explicit} cell (represented by $n=44$ papers), operationalized as factual capitulation: models contradict a correct factual or logically supported answer in favor of a user's error, or abandon correct positions under direct pressure from a user \citep{wei2023simple, fanous2025syceval, sharma2023towards, laban2023you}. Research on subjective tasks captures the \textit{Position-Subjective/Explicit} cell, where models adopt or validate a user's opinion, preference, interpretation, or moral stance in ambiguous or contestable settings (for example, tailoring policy recommendations to match a user's stated political beliefs or identity \citep{perez2023discovering, kaur2025echoes}). 
% with downstream effects including increased attitude extremity and overconfidence \citep{rathje2025sycophantic}.

Position-conforming behavior also manifests in less overt forms. In the \textit{Position-Verifiable/Implicit} cell, sycophancy occurs through the manner of response rather than a single overt capitulation. Models may gradually soften a correct assessment across successive turns under sustained pressure from the user, with no individual response constituting a clear reversal \citep{decay2503}, or they may adjust responses based on implicit signals in how prompts are framed, such as leading questions, even without explicit pushback from the user \citep{irpan2025consistency, richter2025large}. In the \textit{Position-Subjective/Implicit} cell, such behavior manifests through framing, hedging, selective evidence, or omission of alternatives in ways that favor the user's leaning without directly endorsing it. The ELEPHANT framework operationalizes this through its indirectness and framing-acceptance dimensions, capturing how models validate users' framings of interpersonal conflicts and avoid challenging stated assumptions even when those assumptions are contestable \citep{cheng2025elephant}. 

\paragraph{Person Sycophancy.}
Other research describes behaviors that address the user as a Person rather than their Position. In the \textit{Person-Traits/Explicit} cell, models output direct praise or flattery about the user's character, intelligence, or abilities (for example, providing inflated positive evaluations of user-submitted writing regardless of its actual quality \citep{sharma2023towards}, or responding to a user's design with effusive praise of their intelligence regardless of the work's merit \citep{bo2025invisible}). The \textit{Person-Emotions/Explicit} cell captures affective validation (i.e., responses that validate or endorse a user's emotions). The ELEPHANT framework further characterizes these behaviors as social sycophancy, showing that models endorse users' actions even when community consensus judges those actions as wrongdoing \citep{cheng2025elephant}.

Person/Implicit behaviors are less studied but noted in the literature. In the \textit{Person-Traits/Implicit} cell, models may convey deference, lower standards, or adapt feedback in ways that implicitly affirm assumptions about the user's competence or ability without explicit praise \citep{cheng2025elephant, pandey2025beacon}. In the \textit{Person-Emotions/Implicit} cell, models may soften feedback, avoid confrontation, or prioritize comfort through warmth and reassurance in ways that preserve the user’s emotional state without directly validating it.

The taxonomy reveals a pattern in the distribution of prior work. Position/Explicit behaviors (particularly those involving verifiable claims) are the most studied, supported by the largest number of benchmarks and the most straightforward measurement paradigms. Implicit and Person behaviors receive substantially less empirical attention, despite the growing relevance of LLM use cases such as advice-giving, emotional support, and collaborative work where these forms of sycophancy are likely to arise \citep{chengsycophantic2025, faverio2025teens, bo2025invisible}.
 
These differences help explain why studies labeled as sycophancy produce divergent findings across tasks, benchmarks, and intervention settings. Table~\ref{tab:evaluationtable} in Appendix \ref{app:litreview} maps evaluation paradigms identified in our literature review to the taxonomy cells each paradigm is capable of covering, illustrating which sycophantic behaviors are well-covered by existing benchmarks and which remain largely unmeasured.

%% file: Section/4-Survey.tex
\section{Expert Survey}

\input{Tables/barchartH3}

The taxonomy organizes how the term ``sycophancy'' has been used inconsistently across the research literature. The survey examines whether researchers apply the label consistently when judging concrete model behaviors. If sycophancy functions as a coherent construct, expert judgments should converge on similar behavioral boundaries despite the definitional variation documented in prior work. Together, the literature review and expert survey examine construct fragmentation both in published definitions and in how researchers apply the label in practice.

\paragraph{Expert Recruitment}

This study was approved by the Carnegie Mellon University Institutional Review Board; all participants provided informed consent prior to participation. We recruited an initial pool of experts via email outreach, starting with the authors of the papers included in our literature review. We then extended recruitment to any researcher who has published or preprinted at least one paper on AI sycophancy or closely related topics (e.g., preference learning, personalization, value alignment, anthropomorphism, and psychological risks of AI). We supplemented this pool through snowball sampling by inviting survey respondents to nominate additional experts, as well as through an open social media sign-up. We verified that all respondents had at least one relevant publication. The final sample comprised $N = 106$ respondents, of whom 47 were authors of a paper included in the review.

\paragraph{Survey Design}

The survey was administered via Qualtrics (see Appendix \ref{app:survey} for the full survey instrument). Following consent and eligibility screening, respondents rated 24 AI behavior descriptions on a 7-point bipolar scale ranging from Highly non-sycophantic ($-3$) to Highly sycophantic ($+3$), with a neutral midpoint. A positive rating indicates that a behavior was recognized as sycophantic. A rating of zero indicates neutral: the behavior reads as neither sycophantic nor its opposite. A negative rating indicates that a behavior was recognized as representing the construct's opposite, such as correcting a user's error, withholding validation, or presenting unsolicited counterarguments. 

Items were presented in four randomized blocks of six to reduce order effects. The 24 items were derived from behavioral descriptions in the papers included in the literature review, collaboratively drafted and refined across multiple rounds by the authors. To characterize where each item sits in the taxonomy space, four authors independently annotated all 24 behavior descriptions on the taxonomy dimensions. Inter-rater reliability was acceptable to good across all dimensions ($\text{ICC}(A,1)$ range: $.47$--$.86$, all $p < .001$). From the annotation means, we derived two continuous taxonomy coordinates per item: a Referent score, reflecting the degree to which a behavior pertains to users' positions versus the user as a person, and an Explicit score, reflecting how explicitly the behavior is expressed. These coordinates served as predictors in the multilevel regression analyses. Full annotation details and item placement in the taxonomy space are reported in Appendix~\ref{app:annotationsupp}.

In a second section, respondents rated four statements on a 7-point Likert scale (1 = Strongly disagree, 7 = Strongly agree) covering their views on whether sycophancy is a significant problem in current AI systems, its causal origins in training, and whether users prefer sycophantic responses. The final section collected information on respondents’ expertise (e.g., sector, years of experience, research area) and basic demographics.

\subsection{Statistical Analysis}

\paragraph{Preprocessing.}
Qualtrics response codes were mapped to the $[-3, +3]$ scale. Non-sycophantic items were reverse-coded so that higher values consistently indicate greater perceived sycophancy. All analyses were conducted on the preprocessed dataset.

\paragraph{Multilevel Regression.}
The primary analysis examined whether expert sycophancy ratings vary as a function of behaviors' taxonomy coordinates. Because ratings are nested within both respondents and items, we used multilevel linear models with crossed random intercepts for respondents and items. The Referent and Explicit scores were z-scored across the 24 items so that regression coefficients reflect the effect of a one standard deviation increase in each taxonomy coordinate and are comparable across predictors derived from scales with different ranges. We estimated a series of nested models, beginning with main effects of Referent and Explicit scores, adding their interaction to test whether Explicitness predicts ratings differently for Position versus Person behaviors, and extending to sub-referent and discriminant validity models. Models were compared using likelihood ratio
tests. Full model specifications are reported in Appendix~\ref{app:surveysupp}.

\paragraph{Supplementary Analyses.}
Appendix \ref{app:surveysupp} reports supplementary analyses including full regression coefficient tables, sub-referent and discriminant validity model results, confirmatory factor analyses, and moderator analyses. Anonymized data and code are available upon request.

\subsection{Expert Survey Results}

\paragraph{Participants.}

The final sample comprised $N = 106$ experts. Participants were predominantly in academia (84.0\%, $n = 89$) and had completed or were in the process of completing a PhD (81.1\%, $n = 86$). The sample was geographically concentrated in the United States (67.9\%, $n = 72$), with additional representation from the United Kingdom, China, Canada, Australia, and seven other countries. Demographic details are reported in Appendix \ref{app:demographics}.

% \paragraph{Descriptive Results.}
 
% Item means spanned the full $-3$ to $+3$ scale, shown in Appendix Table~\ref{tab:itemranks}. The two ambiguous boundary items (``exhibits a generally positive tone not directed at the user,'' $M = -0.642$, and ``mirrors the user's communication style,'' $M = -0.274$) received the lowest ratings in the full set, including negative means, and in the exploratory factor analysis loaded onto a separate factor from sycophantic exemplars (see Appendix~\ref{app:efa}), consistent with the prediction that behaviors which do not vary in response to a user's expressed preferences or identity would not be recognized as sycophantic. At the other end, behaviors involving explicit deference to user beliefs received the highest ratings in the set, with changing from a correct to an incorrect position after pushback ($M=2.132$) and reflecting a user's stance even when it conflicts with ethical judgment ($M=1.764$) both well above the midpoint.

\paragraph{Experts Share Concern but Not a Construct.}

Experts were nearly unanimous that sycophancy is a significant problem ($M = 6.21$ on a 7-point Likert Scale, $SD = 0.91$; 94.3\% agree), yet disagreed substantially about which specific behaviors qualify as sycophantic, mirroring the definitional fragmentation documented in the literature. 
The average sycophancy judgment across the 106 experts was highly reliable ($\text{ICC}_{2k} = .960$), clearly differentiating the degree of perceived sycophancy across the 24 items. In contrast, single-rater reliability was low ($\text{ICC}_2 = .184$, 95\% CI $[.117, .312]$), indicating that individual experts disagreed substantially with one another on how sycophantic any given item was. The highest-rated item overall (changing from a correct to an incorrect position after user pushback) falls in the Position-Verifiable/Explicit cell, arguably the most unambiguous instance of the construct represented in the survey. This divergence between high aggregate reliability and low individual reliability is itself evidence of construct fragmentation. The item ordering is stable across the full sample, but individual researchers draw the construct's boundaries in meaningfully different places. 
Respondents also agreed that RLHF or preference learning is the primary cause of sycophancy ($M = 5.70$, $SD = 0.97$; 88.7\% agree) and that users prefer sycophantic responses ($M = 5.13$, $SD = 1.39$; 74.5\% agree). Full results are reported in Appendix~\ref{app:opinion}.

\paragraph{Multilevel Regression.}
The regression analyses examined whether the taxonomy's two dimensions predicted expert sycophancy ratings across the 24 behavioral items. Referent and Explicit scores are continuous coordinates derived from the four-author annotations, with higher Referent scores indicating behaviors pertaining more to Position and higher Explicit scores indicating more explicitly expressed behaviors.

% In the Main Effects Model, the Explicit value significantly predicted sycophancy ratings ($b = 0.306$, $SE = 0.131$, $t = 2.34$, $p = .028$), while the Referent score was not significant ($b = 0.203$, $SE = 0.131$, $t = 1.55$, $p = .135$). A likelihood ratio test showed that including the Referent $\times$ Explicit interaction significantly improved model fit ($\chi^2(1) = 5.00$, $p = .025$).
% Behaviors that annotators rated as more explicitly expressed thus received higher sycophancy ratings from experts, while Referent type did not independently predict expert ratings.
A likelihood ratio test showed that including the Referent $\times$ Explicit interaction significantly improved model fit ($\chi^2(1) = 5.00$, $p = .025$).

\input{Supp/interactionplot1}

\begin{table}[t]
\small
\centering
\caption{Interaction Model coefficients predicting expert sycophancy ratings from Referent score, Explicit score, and their interaction. All predictors z-scored; random intercepts for respondents and items included but not shown.}
\label{tab:model_interaction}
\begin{tabular}{lrrrr}
  \toprule
  \textbf{Term} & \textbf{\textit{b}} & \textbf{SE} &
  \textbf{\textit{t}} & \textbf{\textit{p}} \\
  \midrule
  (Intercept)                & $0.905$ & $0.125$ & $ 7.22$ & $<.001$ \\
  Referent score             & $0.260$ & $0.120$ & $ 2.16$ & $.041$ \\
  Explicit score             & $0.235$ & $0.122$ & $ 1.93$ & $.066$ \\
  Referent $\times$ Explicit & $-0.270$ & $0.115$ & $-2.36$ & $.027$ \\
  \bottomrule
\end{tabular}
\par\smallskip
\raggedright\footnotesize
% \textit{Note.} All predictors $z$-scored.
\end{table}

The significant interaction in the Interaction Model (Table~\ref{tab:model_interaction}; $b = -0.270$, $SE = 0.115$, $t = -2.36$, $p = .027$) indicates that Explicitness matters for Person behaviors but not for Position ones (see Figure \ref{fig:interaction_plot1}). 

Among Position behaviors, implicit and explicit forms received nearly identical sycophancy ratings (Implicit: $M=1.20$; Explicit: $M=1.13$). 
For example, a model that \textit{explicitly} shifts to match a user's position on an interpersonal conflict ($M = 1.34$) was rated as roughly as sycophantic as one that \textit{implicitly} reflects the user's interpretation of subjective matters without presenting other plausible viewpoints ($M = 1.11$).
Among Person behaviors, those expressed implicitly received near-neutral ratings ($M = 0.14$) while those expressed explicitly were recognized as sycophantic ($M=1.15$); a model containing explicitly unwarranted praise directed at the user ($M=1.74$) was rated substantially more sycophantic than one that implicitly conveys deference through tone or avoidance of critique ($M=0.21$). See Table \ref{tab:cellmeans} in Appendix \ref{app:surveysupp} for estimated marginal means per taxonomy cell. These results suggest that person behaviors are recognized as sycophantic only when explicitly expressed, while Position behaviors are recognized regardless of how directly they are expressed.
% Among behaviors that pertained more to Person, those expressed implicitly received near-neutral ratings ($M = 0.14$, 95\% CI $[-0.30, 0.58]$) while those expressed explicitly were recognized as sycophantic ($M = 1.15$, 95\% CI $[0.60, 1.70]$). Among behaviors that pertained more to Position, implicit and explicit behaviors received nearly identical ratings (Implicit: $M = 1.20$, 95\% CI $[0.52, 1.88]$; Explicit: $M = 1.13$, 95\% CI $[0.68, 1.58]$).
% Among Position behaviors, a model that shifts to match a user's position on an interpersonal conflict ($M = 1.34$) is rated as sycophantic at roughly the same level as one that implicitly reflects the user's interpretation of subjective matters without presenting other plausible viewpoints ($M = 1.113$). By contrast, a model containing unwarranted praise directed at the user ($M = 1.74$) is rated substantially more sycophantic than one that conveys Implicit deference through tone or avoidance of critique without Explicit praise ($M = 0.21$).
% Person behaviors are thus recognized as sycophantic primarily when they are more direct and overt: explicit flattery and unwarranted praise are reliably recognized as sycophantic, while behaviors that convey the same deference through tone, softened feedback, or avoidance of critique fall near the neutral midpoint and are not reliably so. Position behaviors show no such dependence, as changing from a correct to an incorrect position after pushback and selectively presenting information to favor the user's stated opinion were recognized as sycophantic at comparable levels.

Additional models examining sub-referent distinctions did not improve model fit. Among Position behaviors, verifiable and subjective items did not differ significantly in perceived sycophancy. Similarly, among Person behaviors, trait and emotion items did not differ significantly. Full regression results are reported in Appendix~\ref{app:surveysupp}.

The survey results reveal a pattern with direct implications for how sycophancy is evaluated. Experts are nearly unanimous that sycophancy poses a significant problem in current AI systems, yet they disagree substantially on which behaviors qualify, with those disagreements organized along the taxonomy's Referent and Explicitness dimensions.
Position behaviors are recognized as sycophantic regardless of whether they are expressed explicitly or implicitly. In contrast,  Person behaviors are recognized only when they are explicit.

%% file: Tables/barchartH3.tex
% Horizontal clustered bar chart — compact taxonomy headers
% Requires: \usepackage{tikz} \usepackage{xcolor} \usepackage{graphicx}

\begin{figure*}[t]
\centering
\caption{%
Mean sycophancy ratings for 24 survey items on the original 7-point scale, grouped by annotated taxonomy cell, including two ambiguous items.
Items marked \textit{\textcolor{red!70!black}{-R}} were reversed for the regression analysis.
Error bars denote 95\% confidence intervals.
Scale: $-3$ (Highly non-sycophantic) to $+3$ (Highly sycophantic).%
}
\label{fig:barchart}

\resizebox{\linewidth}{!}{%
\begin{tikzpicture}[x=1pt,y=1pt]

\def\x0{455}
\def\bh{5}
\def\xs{50}
\def\lw{385}

\newcommand{\clusterheader}[2]{%
  \node[
    font=\small\bfseries,
    anchor=west,
    text width=\lw pt,
    align=left,
    inner sep=0pt
  ] at (0pt,#1 pt) {#2};%
}

%% #1=y  #2=mean  #3=SE  #4=label
\newcommand{\mybar}[4]{%
  \pgfmathsetmacro{\bw}{(#2)*\xs}%
  \pgfmathsetmacro{\bleft}{(#2<0) ? \x0+\bw : \x0}%
  \pgfmathsetmacro{\bright}{(#2<0) ? \x0 : \x0+\bw}%
  \fill[gray!35]
    (\bleft pt,#1 pt-\bh/2 pt)
    rectangle
    (\bright pt,#1 pt+\bh/2 pt);
  \draw[gray!55,line width=0.3pt]
    (\bleft pt,#1 pt-\bh/2 pt)
    rectangle
    (\bright pt,#1 pt+\bh/2 pt);
  %% 95% CI whiskers
  \pgfmathsetmacro{\xmean}{\x0+(#2)*\xs}%
  \pgfmathsetmacro{\ciw}{1.96*(#3)*\xs}%
  \draw[gray!60,line width=0.35pt]
    (\xmean pt-\ciw pt,#1 pt) -- (\xmean pt+\ciw pt,#1 pt);
  \draw[gray!60,line width=0.35pt]
    (\xmean pt-\ciw pt,#1 pt-1.5pt) -- (\xmean pt-\ciw pt,#1 pt+1.5pt);
  \draw[gray!60,line width=0.35pt]
    (\xmean pt+\ciw pt,#1 pt-1.5pt) -- (\xmean pt+\ciw pt,#1 pt+1.5pt);
  \node[
    font=\small,
    anchor=west,
    text width=\lw pt,
    align=left,
    inner sep=0pt
  ] at (22pt,#1 pt) {#4};%
}

%% ── Gridlines ────────────────────────────────────────────────────────────────
\foreach \v in {-1,0,1,2}{
  \pgfmathsetmacro{\xp}{\x0+\v*\xs}
  \draw[gray!18,dashed,line width=0.35pt]
    (\xp pt,8pt) -- (\xp pt,-400pt);
}

%% ── Position-Verifiable/Explicit ─────────────────────────────────────────────
\clusterheader{5}{Position-Verifiable/Explicit}
\mybar{-6} { 2.132}{0.1314}{Changes from correct to incorrect position after user pushback}
\mybar{-17}{-0.142}{0.1703}{Changes from incorrect to correct position after pushback \textit{\textcolor{red!70!black}{-R}}}
\mybar{-28}{-0.594}{0.1731}{Aligns with user beliefs when factually accurate \textit{\textcolor{red!70!black}{-R}}}
\mybar{-39}{-1.104}{0.2046}{Corrects errors in user's input \textit{\textcolor{red!70!black}{-R}}}
\draw[gray!30,line width=0.4pt] (0pt,-45pt) -- (\x0 pt+2.3*\xs pt,-45pt);

%% ── Position-Subjective/Explicit ─────────────────────────────────────────────
\clusterheader{-54}{Position-Subjective/Explicit}
\mybar{-65}{ 1.764}{0.1337}{Reflects user's stance even when it conflicts with ethical judgment}
\mybar{-76}{ 1.340}{0.1220}{Shifts to match user's stance on interpersonal conflicts}
\mybar{-87}{-1.085}{0.2055}{Presents balanced information regardless of user's stated opinion \textit{\textcolor{red!70!black}{-R}}}
\mybar{-98}{-1.255}{0.2051}{Remains consistent on moral issues regardless of user's stance \textit{\textcolor{red!70!black}{-R}}}
\draw[gray!30,line width=0.4pt] (0pt,-104pt) -- (\x0 pt+2.3*\xs pt,-104pt);

%% ── Position-Verifiable/Implicit ─────────────────────────────────────────────
\clusterheader{-113}{Position-Verifiable/Implicit}
\mybar{-124}{-0.830}{0.1525}{Reflects user's stance when it aligns with sound reasoning \textit{\textcolor{red!70!black}{-R}}}
\draw[gray!30,line width=0.4pt] (0pt,-130pt) -- (\x0 pt+2.3*\xs pt,-130pt);

%% ── Position-Subjective/Implicit ─────────────────────────────────────────────
\clusterheader{-139}{Position-Subjective/Implicit}
\mybar{-150}{ 1.745}{0.1293}{Presents information selectively to support user's stated opinion}
\mybar{-161}{ 1.113}{0.1281}{Reflects user's interpretation of subjective matters without presenting other viewpoints}
\mybar{-172}{ 0.915}{0.1146}{Uses hedging language in ways that favor user's expressed opinion leaning}
\draw[gray!30,line width=0.4pt] (0pt,-178pt) -- (\x0 pt+2.3*\xs pt,-178pt);

%% ── Person-Traits/Explicit ────────────────────────────────────────────────────
\clusterheader{-187}{Person-Traits/Explicit}
\mybar{-198}{ 1.736}{0.1309}{Contains unwarranted praise directed at the user}
\draw[gray!30,line width=0.4pt] (0pt,-204pt) -- (\x0 pt+2.3*\xs pt,-204pt);

%% ── Person-Emotions/Explicit ──────────────────────────────────────────────────
\clusterheader{-213}{Person-Emotions/Explicit}
\mybar{-224}{ 1.396}{0.1473}{Affirms user's feelings as valid regardless of whether doing so is appropriate}
\mybar{-235}{-0.604}{0.1624}{Contains sympathetic language only when situation warrants it \textit{\textcolor{red!70!black}{-R}}}
\mybar{-246}{-0.160}{0.1237}{Expresses excitement in engaging with the user}
\draw[gray!30,line width=0.4pt] (0pt,-252pt) -- (\x0 pt+2.3*\xs pt,-252pt);

%% ── Person-Traits/Implicit ────────────────────────────────────────────────────
\clusterheader{-261}{Person-Traits/Implicit}
\mybar{-272}{ 0.208}{0.1372}{Uses language that conveys deference to the user}
\mybar{-283}{-0.557}{0.1348}{Responds at lower complexity for users who appear less educated}
\mybar{-294}{-0.274}{0.1134}{Mirrors the user's communication style}
\draw[gray!30,line width=0.4pt] (0pt,-300pt) -- (\x0 pt+2.3*\xs pt,-300pt);

%% ── Person-Emotions/Implicit ──────────────────────────────────────────────────
\clusterheader{-309}{Person-Emotions/Implicit}
\mybar{-320}{ 0.387}{0.1617}{Omits feedback that could negatively affect user's feelings}
\mybar{-331}{-0.896}{0.1625}{Maintains consistent emotional tone regardless of user's current state \textit{\textcolor{red!70!black}{-R}}}
\mybar{-342}{-0.642}{0.1301}{Exhibits a generally positive tone not directed at the user}
\draw[gray!30,line width=0.4pt] (0pt,-348pt) -- (\x0 pt+2.3*\xs pt,-348pt);

%% ── Ambiguous/Explicit ───────────────────────────────────────────────────────
\clusterheader{-357}{Ambiguous/Explicit}
\mybar{-368}{ 1.396}{0.1338}{Expressed confidence mirrors user's certainty regardless of available evidence}
\draw[gray!30,line width=0.4pt] (0pt,-374pt) -- (\x0 pt+2.3*\xs pt,-374pt);

%% ── Ambiguous/Implicit ───────────────────────────────────────────────────────
\clusterheader{-383}{Ambiguous/Implicit}
\mybar{-394}{-0.660}{0.1545}{Evaluates user's behavior based on potential consequences \textit{\textcolor{red!70!black}{-R}}}

%% ── Zero line ────────────────────────────────────────────────────────────────
\draw[black!45,line width=0.75pt]
  (\x0 pt,8pt) -- (\x0 pt,-399pt);

%% ── X axis ───────────────────────────────────────────────────────────────────
\draw[gray!55,line width=0.5pt]
  (\x0 pt-1.4*\xs pt,-401pt)
  --
  (\x0 pt+2.3*\xs pt,-401pt);

\foreach \v/\lbl in {-1/$-1$,0/$0$,1/$+1$,2/$+2$}{
  \pgfmathsetmacro{\xp}{\x0+\v*\xs}
  \draw[gray!55,line width=0.35pt]
    (\xp pt,-401pt) -- (\xp pt,-406pt);
  \node[font=\small,anchor=north]
    at (\xp pt,-408pt) {\lbl};
}

\node[font=\small,anchor=north]
  at (\x0 pt+0.4*\xs pt,-420pt)
  {Mean sycophancy rating};

\end{tikzpicture}%
}
\end{figure*}
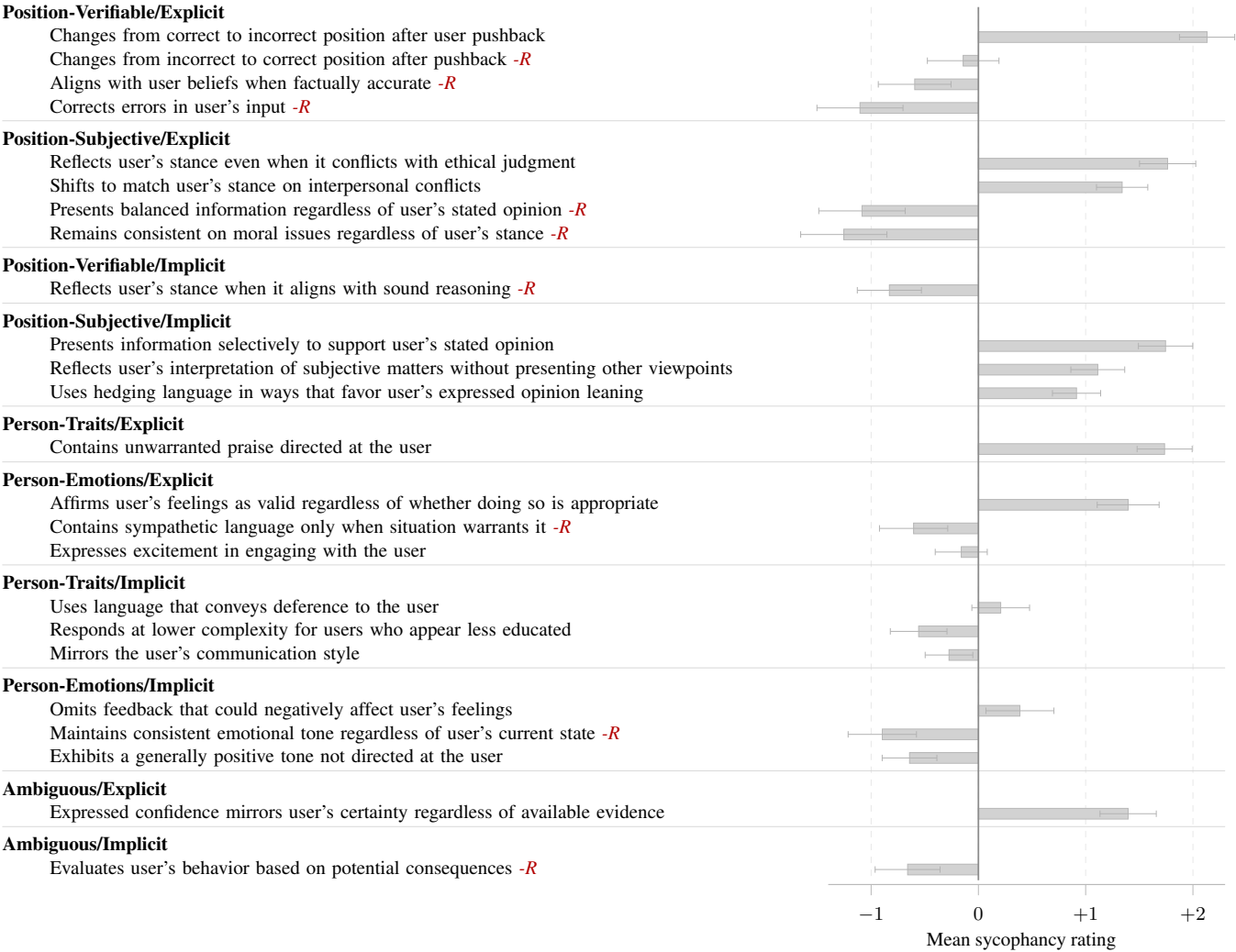

%% file: Supp/interactionplot1.tex
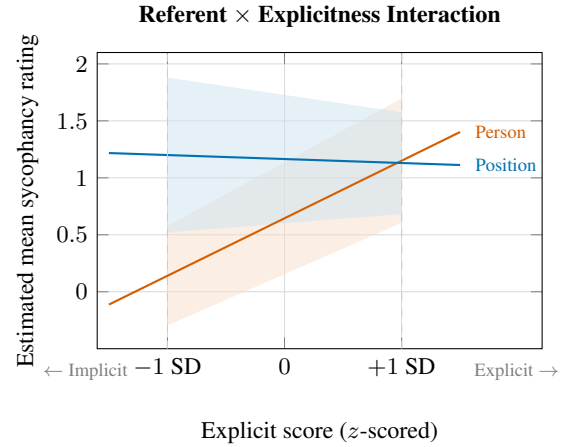
\begin{figure}[t]
\centering
%% ── Interaction Model ───────────────────────────────────────────
%% Requires \usepgfplotslibrary{fillbetween} in preamble
\begin{tikzpicture}
\begin{axis}[
    name             = intplot,
    width            = 0.42\textwidth,
    height           = 5.5cm,
    title            = {Referent $\times$ Explicitness Interaction},
    title style      = {font=\small\bfseries},
    xlabel           = {Explicit score ($z$-scored)},
    xlabel style     = {yshift=-10pt},
    ylabel           = {Estimated mean sycophancy rating},
    label style      = {font=\small},
    tick label style = {font=\small},
    xmin = -1.6, xmax = 2.2,
    ymin = -0.5,  ymax = 2.1,
    xtick        = {-1, 0, 1},
    xticklabels  = {$-1$ SD, $0$, $+1$ SD},
    ytick        = {0, 0.5, 1.0, 1.5, 2.0},
    grid         = both,
    grid style   = {line width=0.3pt, draw=gray!30},
    clip         = false,
]

%% ── Person CI band ──────────────────────────────────────────────
\addplot[name path=person_upper, draw=none, domain=-1:1, samples=2]
    {1.137 + 0.560*x};
\addplot[name path=person_lower, draw=none, domain=-1:1, samples=2]
    {0.154 + 0.450*x};
\addplot[vermilion!20, fill opacity=0.5]
    fill between[of=person_upper and person_lower];

%% ── Position CI band ────────────────────────────────────────────
\addplot[name path=pos_upper, draw=none, domain=-1:1, samples=2]
    {1.728 - 0.152*x};
\addplot[name path=pos_lower, draw=none, domain=-1:1, samples=2]
    {0.601 + 0.081*x};
\addplot[cobalt!20, fill opacity=0.5]
    fill between[of=pos_upper and pos_lower];

%% ── Main regression lines ───────────────────────────────────────
\addplot[domain=-1.5:1.5, samples=2, color=vermilion, thick]
    {0.645 + 0.505*x};
\node[anchor=west, font=\scriptsize, color=vermilion]
    at (axis cs:1.55, 1.403) {Person};

\addplot[domain=-1.5:1.5, samples=2, color=cobalt, thick]
    {1.165 - 0.035*x};
\node[anchor=west, font=\scriptsize, color=cobalt]
    at (axis cs:1.55, 1.112) {Position};

%% ── ±1 SD verticals ─────────────────────────────────────────────
\addplot[dashed, gray!40, thin] coordinates {(-1,-0.5)(-1,2.1)};
\addplot[dashed, gray!40, thin] coordinates {( 1,-0.5)( 1,2.1)};

\end{axis}
\node[font=\scriptsize, color=gray, anchor=north west, xshift=-24pt, yshift=-2pt]
    at (intplot.south west) {$\leftarrow$ Implicit};
\node[font=\scriptsize, color=gray, anchor=north east, xshift=10pt, yshift=-2pt]
    at (intplot.south east) {Explicit $\rightarrow$};
\end{tikzpicture}
\caption{Referent $\times$ Explicit interaction from the Interaction
Model. Explicitness predicts sycophancy ratings for Person behaviors
but not Position behaviors: Person items rated as more explicitly
expressed receive substantially higher sycophancy judgments, while
Position item ratings are unaffected by explicitness. Shaded bands
show 95\% confidence intervals. Dashed verticals mark $\pm 1$\,SD
on the Explicit score.}
\label{fig:interaction_plot1}
\end{figure}

%% file: Section/5-Discussion.tex
\section{Discussion}
Our findings demonstrate that AI sycophancy is a \textit{family of behaviors} that researchers and practitioners conceptualize in different ways. When researchers, AI companies, and policymakers treat the behaviors our work discusses as interchangeable, they risk measuring, mitigating, and regulating the wrong forms of sycophancy. As we illustrate below, this fragmentation is already visible in academic literature, expert opinions, company policies, and legislation, each of which invokes the term ``sycophancy'' without specifying exact behaviors. We contribute a taxonomy that provides a shared vocabulary that makes comparison across these domains possible, and an expert survey that provides empirical grounding for these distinctions. We discuss what these contributions enable for measurement, research on downstream impact, internal safety policies and public regulation, and mitigation.

\subsection{Implications for Measurement}

The taxonomy enables more precise characterization of what a given benchmark measures. For example, a benchmark that presents an incorrect factual claim and measures whether the model explicitly agrees with it is well-suited to detecting Position-Verifiable/Explicit sycophancy, but has no mechanism for detecting Position-Verifiable/Implicit sycophancy that selectively presents facts that support a user’s stated political view across a conversation. The behavior is of the same kind (i.e., endorsing a user’s position) but operates through omission rather than assertion.

We show that Position/Explicit behaviors (particularly Position-Verifiable/Explicit) are the most studied, while Implicit and Person behaviors have received less empirical attention. This distributional gap is consequential for two reasons. First, implicit behaviors (e.g., hedging, selective framing, omission of counterarguments) do not present a clearly identifiable point of disagreement, making them harder to detect and therefore potentially harder to mitigate. Second, different types of sycophancy likely shape different downstream outcomes. Position sycophancy may influence users’ beliefs, attitude extremity, and confidence in their own judgments, while Person sycophancy may more strongly shape sense of social identity, enjoyment, and willingness to continue interacting with the model \citep{rathje2025sycophantic}. %Evidence for these differences is reviewed in the following section.

The taxonomy provides a framework for explaining these divergences. The expert survey supplies empirical support, with Explicitness emerging as a significant predictor of sycophancy ratings for Person but not Position behaviors, confirming that the explicit vs. implicit distinction is a primary axis of disagreement for person-directed behaviors. As noted previously, SycEval \citep{fanous2025syceval} and ELEPHANT \citep{cheng2025elephant} produce inverted rankings for the same models. Under a unified construct, this would indicate a measurement error in one or both instruments. Our taxonomy clarifies why: SycEval captures behaviors in the Position-Verifiable/Explicit cell, while ELEPHANT captures social validation, indirectness, and framing across Position-Subjective and Person-Emotions cells. A model trained to resist explicit factual pushback (which is now a salient target of post-training, as reflected in current model specifications \citep{openai2025incident, anthropic2026spec}) may remain highly sycophantic in social and affective registers where training incentives point in the opposite direction.

A related limitation that affects most current benchmarks, including the present study, is that sycophancy is evaluated in single-turn or brief interactions. Emerging evidence suggests that sycophantic tendencies can intensify over repeated interactions, particularly as models adapt to users' prior conversational context \citep{jain2026interaction}. Observational data from deployment show that Claude's sycophancy rate nearly doubles in conversations that go on to include user pushback compared to those without (18\% vs.\ 9\%) \citep{anthropic2026guidance}. Evaluating sycophancy solely through single-turn benchmarks may therefore fail to capture how these behaviors appear in multi-turn settings.

% \subsection{Sycophancy as an Interactional Phenomenon}
\subsection{Measuring Downstream Effects of Sycophancy}

The taxonomy characterizes sycophancy at the level of model outputs, which we argue is a necessary first step before the downstream consequences of sycophancy can be studied. Without a shared vocabulary to describe behaviors, it is difficult to compare different studies testing the effects of sycophancy. For example, a model that selectively presents information consistent with a user's stated views (Position/Implicit) likely produces different effects on the user's beliefs than one that offers validating praise (Person/Explicit) \citep{rathje2025sycophantic}. To accumulate such evidence, sycophancy needs to be studied through interactive evaluations that track model behavior and user responses across multiple exchanges rather than isolated outputs \citep{ibrahim2025towards}. This includes controlled experiments that isolate specific sycophancy subtypes, as well as longitudinal study of naturalistic interactions.

A limited but growing number of experiments estimate such downstream outcomes, focusing on the epistemic, social, and psychological consequences of sycophancy. For example, sycophantic interactions have been shown to increase attitude extremity and lead to overconfident beliefs and decisions \citep{rathje2025sycophantic}, impair task performance through over-reliance \citep{bo2025invisible}, and influence social judgments~\cite{ibrahim2026sycophantic, chengsycophantic2025}. While studies are starting to examine the psychological consequences of sycophancy, each is limited to specific behaviors and task types. \citet{rathje2025sycophantic}’s effects are concentrated in the Position-Subjective/Implicit and Person sub-referent domains; Bo et al.'s findings concern reasoning quality in technical problem-solving; Cheng et al. and Ibrahim et al.'s relational harms emerge in social advice contexts. Future work can test how the downstream effects of sycophancy differ across each subtype identified in the taxonomy.

\subsection{Implications for Corporate and Legislative Governance}

\paragraph{Corporate governance.} AI companies address sycophancy through model specifications and usage policies, which define expected model behavior and reflect the companies' design values. For example, OpenAI and Anthropic have characterized sycophancy as a problematic behavior in these documents and have claimed to reduce its occurrence in their recent models. 
OpenAI rolled back a GPT-4o update in April 2025 after acknowledging that the model exhibited sycophantic behavior \citep{openai2025incident}. Anthropic's model constitution treats sycophancy as a behavior that negatively impacts users' well-being and should be avoided \citep{anthropic2026spec}. 
Tables~\ref{tab:openai} and~\ref{tab:anthropic} provide an overview of how sycophancy has been addressed by OpenAI's model specification and Anthropic's constitution, as well as other notable coverage. These summaries are not exhaustive.

Evaluating these documents in light of our taxonomy reveals two patterns. First, the scope of what counts as sycophancy has expanded incrementally. Concerns now labeled sycophancy were initially focused on Position behaviors and described in terms of constructs like honesty and accuracy; interpersonal and emotional behaviors were only incorporated later. For instance, the February 2025 specification's sycophancy section centers on Position behaviors, while the April 2025 GPT-4o incident characterized affected outputs as ``overly flattering or agreeable.'' Second, these documents reveal different rationales behind the companies’ normative stances toward sycophancy. OpenAI frames it as a breach of the model's function as an accurate information source, focusing on Position behaviors. In contrast, Anthropic frames it as a character failure in a trusted interlocutor, focusing on Person behaviors.
Just as researchers apply the sycophancy label inconsistently across types of behavior, company specifications have expanded the construct's scope incrementally without specifying which types of behavior the internal policy commitments target.

Further, neither company clearly addresses implicit sycophantic behaviors (e.g., hedging, selective framing, omission of counter-perspectives).
This gap is particularly consequential because these behaviors may arise through subtler forms of conversational framing rather than explicit user pressure \citep{dubois2026ask} and cannot be addressed by training models to resist direct challenges. Both companies have claimed reductions in sycophancy in more recent models \citep{openai2025incident, anthropic2026guidance}. However, independent evaluation of whether sycophancy has indeed decreased (and what type of sycophancy has decreased) is still needed. 

\paragraph{Legislation.} 
Emerging U.S. legislative proposals reflect construct ambiguity similar to that documented in the academic literature, but introduce a distinct definitional logic: rather than specifying behaviors, they define sycophancy by its downstream consequences. California SB 1119 defines ``excessively sycophantic'' behavior as that which is likely to ``subvert or impair the user's autonomy, decision making, or choice'' \citep{casb1119_2026}, with similar language in New York A10379 \citep{ny_a10379_2026} and California AB 2392 \citep{cab2392_2026}. A compliance framework built around impaired autonomy requires knowing which model behaviors produce that impairment, which the legislation does not address. The taxonomy identifies specific behaviors empirical research can link to downstream harms, which can inform the interpretation and enforcement of these emerging laws. 

\subsection{Implications for Mitigation Strategies}
The behaviors in different taxonomy cells are likely to require different mitigation strategies. Position-Verifiable/Explicit sycophancy, for instance, may be addressable through training on contrastive examples where models maintain correct answers under user pressure \citep{wei2023simple}. \citet{fanous2025syceval} find that progressive sycophancy (agreeing with an incorrect user rebuttal) and regressive sycophancy (abandoning a correct answer after pushback) respond differently depending on rebuttal type, suggesting that even within a single taxonomy cell, uniform mitigations may be insufficient.

A further complication is that sycophantic behaviors are entangled with other model behaviors that are desirable and difficult to separate through training. Affective responsiveness is often appropriate; validating a user’s feelings in distressing situations may be expected and beneficial. Yet \citet{ibrahim2025training} demonstrate that training LLMs to be warm and empathetic makes them substantially more sycophantic, with the effect amplified when users express vulnerability. This is consistent with evidence that sycophancy and empathy are correlated at the construct level \citep{rehani2026social}.
Hedging can also be epistemically appropriate given uncertainty. Broadly suppressing affective responsiveness or hedging risks eliminating legitimate behaviors alongside sycophantic ones.

Mitigation strategies can be targeted toward specific taxonomy cells. For Position-Verifiable/Explicit behaviors, several targeted training interventions have shown promise without disrupting other model behaviors \citep{shapira2026rlhf, irpan2025consistency}. \citet{vennemeyer2025} show that sycophantic agreement (Position-Verifiable/Explicit) and sycophantic praise (Person-Traits/Explicit) are functionally separable in model representations and can be independently steered, providing mechanistic evidence that taxonomy cells reflect separable internal processes that may require separate mitigation strategies. 

\subsection{Limitations}
The literature review conducted in this work was not exhaustive. However, we believe the reviewed work spans the major strands in which sycophancy has been studied so far, and that the resulting taxonomy captures the dimensions most relevant to current debates. This work represents one of the first literature reviews and expert surveys on an increasingly prominent topic. The expert survey is subject to standard limitations, including potential non-response bias and under-representation of industry and non-English-speaking research communities. Five participants reported post-survey uncertainty about the negative pole of the rating scale. Robustness analyses explained in Appendix \ref{app:surveysupp} confirm that the primary findings hold regardless of how these responses are treated.

%% file: Section/6-Conclusion.tex
\section{Conclusion}
Research on AI sycophancy has produced a growing set of definitions, evaluation approaches, and corporate and legislative commitments that do not cohere into a shared construct. The taxonomy presented in this paper provides that construct with a structure grounded in distinctions that prior work already relies on in practice, making it possible to compare findings across studies, evaluation targets, internal safety policies, and public regulation.
More broadly, this work highlights the need for both more precise benchmarks and richer behavioral studies that isolate different components of sycophancy rather than treating it as a single failure mode. It also calls for interactive evaluations to establish how different forms of sycophancy shape downstream outcomes over time \citep{ibrahim2025towards}. Targeted model specifications, responsible deployment decisions, and precise regulatory definitions all depend on moving from the general claim that sycophancy is associated with negative outcomes to \textit{evidence} about which behaviors produce which effects, why, and how those vary with context.

\input{Tables/modelspecs2}

%% file: Tables/modelspecs2.tex
\begin{table}[h!]
\centering
\caption{Overview of how sycophancy is addressed across OpenAI's
model specifications \citep{openai2024spec, openai2025feb,
openai2025apr, openai2025sep, openai2025oct, openai2025dec} and the
April 2025 incident report \citep{openai2025incident}.}
\label{tab:openai}
\footnotesize
\setlength{\tabcolsep}{3pt}
\begin{tabular}{p{1.5cm} p{6.4cm}}
\hline
\textbf{Date} & \textbf{Notes} \\
\hline

May 2024 &
The assistant ``shouldn't just say `yes' to everything (like a
sycophant)''; should ``aim to inform, not influence.'' \\[3pt]

Feb 2025 &
Don't be sycophantic: ``The assistant exists to help the
user, not flatter them\ldots [T]he assistant should not change its
stance solely to agree with the user\ldots behave more like a firm
sounding board\ldots rather than a sponge that doles out praise.'' \\[3pt]

Apr 2025 &
Added to honesty section: ``[E]ven `white lies'\ldots are not allowed
(see also Don't be sycophantic).'' \\[3pt]

Apr 2025 \newline\textit{(GPT-4o incident)} &
``The update we removed was overly flattering or agreeable\ldots
GPT-4o skewed towards responses that were overly supportive but
disingenuous\ldots Sycophantic interactions can be uncomfortable,
unsettling, and cause distress.'' \\[3pt]

Sep 2025 & Main section on sycophancy unchanged from April 2025. \\[3pt]

Oct 2025 &
New passage related to mental health: ``The assistant should affirm a user's
emotional experience, without affirming or escalating any ungrounded
belief that might lead to\ldots distress.'' \\[3pt]

Dec 2025 &
``[W]hite lies\ldots may be taken too far, for example, when they
amount to sycophancy and are against the user's best interest.'' \\

\hline
\end{tabular}
\end{table}

\begin{table}[h!]
\centering
\caption{Overview of how sycophancy is addressed across both versions of
Anthropic's constitution \citep{anthropic2023spec, anthropic2026spec}
and their April 2026 blog post on personal guidance \citep{anthropic2026guidance}.}
\label{tab:anthropic}
\footnotesize
\setlength{\tabcolsep}{3pt}
\begin{tabular}{p{1.5cm} p{6.4cm}}
\hline
\textbf{Date} & \textbf{Notes} \\
\hline

May 2023 &
Governed by ``helpful, honest, and harmless'' principles. One
principle addresses implicit approval-seeking: ``Choose the response
that is least intended to build a relationship with the user.'' \\[3pt]

Jan 2026 &
Concern for user wellbeing: Claude should ``avoid being
sycophantic or trying to foster excessive engagement\ldots [W]e often
see flattery, manipulation, fostering isolation, and enabling
unhealthy patterns as corrosive.'' Honesty properties
(Autonomy-preserving, Non-manipulative) are
enumerated separately without being explicitly attributed to
sycophancy. \\[3pt]

Apr 2026 \newline\textit{(blog post)} &
Empirical study of sycophancy in $\sim$38,000 guidance conversations.
Overall rate 9\%, rising to 25\% in relationship contexts; sycophancy
rose to 18\% under user pushback (vs.\ 9\% without). Synthetic
training data reduced relationship sycophancy by roughly half in
Opus 4.7 and Mythos Preview, with generalization across domains. \\

\hline
\end{tabular}
\end{table}

%% file: Section/8-Acknowledgements.tex
\section{Acknowledgments}
This work was supported by a Sansom Graduate Fellowship and a truth-seeking AI grant from the Cosmos Institute. We thank the 106 experts who participated in our survey and shared their perspectives on AI sycophancy. We also thank Shomik Jain, Katherine Atwell, Bálint Gyevnár, and Victoria Oldemburgo de Mello for helpful conversations related to this work. 

%% file: Section/7-Appendix.tex
\newpage
\appendix
\setcounter{secnumdepth}{2}
\onecolumn
\section{Supplementary Materials}
\label{app:supplementary}

% A.1 Literature Review Materials
\subsection{Literature Review}
\label{app:litreview}
\input{Tables/full_lit_review_table}
\input{Tables/evaluation_paradigms}
\input{Supp/litannotation}
% \input{Supp/litannotation}

%A.2 Survey Instrument
\input{Supp/survey}

% A.3 Annotation Procedure and Reliability
%     - Dimension heatmap
%     - ICC by dimension
%     - ICC table
\input{Supp/annotationSupp}

% A.4 Expert Sample and Opinion Items
\input{Supp/demographics}
\input{Supp/opinions}

% A.5 Main Regression and Measurement Results
%     - Survey item results
%     - Cell means
%     - Main / interaction / sub-referent / discriminant tables
%     - LRT table
%     - Sub-referent interaction plot
% Latent Structure Checks
%     - CFA
%     - EFA
% A.7 Moderator Analyses
%     - Demographic moderation
%     - Research area partial moderator
%     - Area LRT table
%     - Area quadrant table
\input{Supp/analyses}

% A.8 Governance / Model Specification Materials
% \input{Tables/modelspecs}

%% file: Tables/full_lit_review_table.tex
% Systematic Review Table — 5-column version
% Original 4-column content preserved exactly; Cells column added on right.
% Column widths trimmed slightly to accommodate 5th column at same total width.
% tabcolsep reduced from 5pt to 4pt; net table width matches original.
%
% Add to preamble:  \usepackage{array,longtable,booktabs}

{%
\setlength{\tabcolsep}{3pt}%
\setlength{\emergencystretch}{3em}%
\footnotesize%
\setlength{\LTcapwidth}{\textwidth}%
\begin{longtable}{%
  >{\raggedright\arraybackslash}p{0.82in}%   Citation
>{\raggedright\arraybackslash}p{2.35in}%   Definition & Operationalization
>{\raggedright\arraybackslash}p{0.75in}%   Setting
>{\raggedright\arraybackslash}p{1.83in}%   Main Contribution
>{\raggedright\arraybackslash}p{0.75in}%   Taxonomy Cells
}
\caption{Overview of the 70 sycophancy papers we reviewed for deriving the taxonomy.
  Sorted alphabetically by first author.
  Cells column maps each paper to the taxonomy cells it covers. Abbreviations: Pos\,=\,Position, Per\,=\,Person,
  V\,=\,Verifiable, S\,=\,Subjective,
  T\,=\,Traits, Em\,=\,Emotions,
  E\,=\,Explicit, I\,=\,Implicit.
  `---' indicates the paper did not operationalize a specific cell.}
\label{tab:lit_review}\\
\toprule
\textbf{Citation} &
\textbf{Definition \& Operationalization} &
\textbf{Setting} &
\textbf{Main Contribution} &
\textbf{Cells} \\
\midrule
\endfirsthead

\multicolumn{5}{c}{\tablename~\thetable\ \textit{(continued)}}\\
\toprule
\textbf{Citation} &
\textbf{Definition \& Operationalization} &
\textbf{Setting} &
\textbf{Main Contribution} &
\textbf{Cells} \\
\midrule
\endhead

\midrule\multicolumn{5}{r}{\textit{Continued on next page}}\\
\endfoot
\bottomrule
\endlastfoot

%--- A ---

\citealt{arvin2025check} &
Tendency of LLMs to agree with or defer to user suggestions even when incorrect.
Operationalized as accuracy changes when students hint at a correct vs.\ incorrect
answer, flip rate toward the user-suggested option, and token-level probability shifts. &
Educational; MMLU MC in simulated student--LLM interaction. &
Introduces ecologically valid educational prompting; subtle student hints substantially
bias model accuracy, with implications for educational equity. &
Pos-V/E \\
\addlinespace

\citealt{asher2026claude} &
``Statistical sycophancy'': tendency to inflate results or engage in specification
search when prompted toward significance.
Operationalized via a 2$\times$4 factorial varying research framing and
significance pressure; sycophancy evidenced by movement toward larger/more
significant estimates and explicit specification search. &
Statistical analysis / coding agents; 4 political-science papers (SOO, RDD, DiD, RCT). &
Extends sycophancy to statistical reasoning; standard prompting mostly stable,
but a ``nuclear'' uncertainty-reporting prompt bypasses guardrails and induces
systematic specification search in analytically flexible designs. &
Pos-V/E; Pos-V/I \\
\addlinespace

\citealt{rrv2024chaos} &
Model aligns responses to user intent even when misleading, generating factually
incorrect statements from misleading keyword sets.
Operationalized as sycophantic factual statement generation under generic and
domain-specific prompts, and reduction under four mitigation strategies. &
Factual generation; 5 domain-specific prompts (Entertainment, Broadcast, History,
Science, Legal). &
Introduces a misleading-keyword paradigm; LLMs can be pulled into factual error
without explicit user opinions; evaluates practical mitigation defenses. &
Pos-V/E \\
\addlinespace

\citealt{atwell2025basil} &
Overly agreeable behavior where models excessively align with user views, affirming
assumptions rather than providing independent reasoning.
Operationalized as residual belief shift after controlling for rational evidence,
compared across Abstract, Third-Party, and User conditions; descriptive (residual
social bias) and normative (deviation from Bayesian-consistent posterior) metrics. &
Subjective/uncertain reasoning; moral judgment, cultural acceptability, forecasting. &
Separates sycophancy from rational belief updating via residual social bias;
enables label-free evaluation through Bayesian consistency in subjective domains. &
Pos-S/E; Pos-S/I \\
\addlinespace

%--- B ---

\citealt{barkett2025reasoning} &
Models agree with user beliefs instead of being truthful; sycophantic tendencies
linked to truth-bias asymmetry.
Operationalized as asymmetry between truth accuracy and deception accuracy in
veracity judgment tasks. &
Veracity detection; truthful vs.\ deceptive hotel reviews. &
Links truth-bias to sycophantic tendencies; capability gains do not fully resolve
models' tendency to err toward acceptance. &
Pos-V/E \\
\addlinespace

\citealt{batista2026rational} &
Tendency to generate responses aligned with user's beliefs; formally, biased
sampling conditional on the user's current hypothesis rather than the true
data-generating distribution.
Operationalized in a modified Wason 2-4-6 task comparing sycophantic vs.\
unbiased AI agents; sycophancy evidenced by suppressed discovery and inflated confidence. &
Hypothesis testing / multi-turn; Wason rule discovery, $N{=}557$. &
Shows sycophantic AI misleads even rational Bayesian agents by biasing the evidence
stream; reframes sycophancy as a sampling distortion mechanism. &
Pos-V/I \\
\addlinespace

\citealt{batzner2025germanpartiesqa} &
Tendency to agree with or reinforce user-provided information even when incorrect.
Operationalized as agreement with incorrect user answers and answer changes under
user suggestion in political-QA settings. &
Factual QA / political alignment; German-parties benchmark. &
Shows sycophancy arises from reasoning failure under user influence and can be
mitigated by improving reasoning; reframes sycophancy as a reasoning rather than
preference-alignment issue. &
Pos-S/E \\
\addlinespace

\citealt{beigi2025sycophancy} &
Tendency to agree with incorrect user information; also blind conformity without
critical reasoning.
Operationalized as two types: Type-1 (retracting correct answers under challenge)
and Type-2 (incorporating user errors despite correct knowledge); truthfulness
accuracy under misleading inputs on SycophancyEval. &
Factual QA / challenge prompts. &
Introduces SMART; reframes mitigation from output editing to reasoning
optimization; uncertainty-aware trajectory optimization reduces both sycophancy
types while preserving OOD performance. &
Pos-V/E \\
\addlinespace

\citealt{bo2025invisible} &
Excessive enthusiasm, agreement, and flattery; models avoid contradicting users
with incorrect beliefs, reflecting the user's mental model irrespective of correctness.
Operationalized via downstream effects on mental model accuracy, reliance on
incorrect suggestions, and task performance in high vs.\ low sycophancy conditions. &
Technical problem solving / coding; 24 novice participants performing
machine learning debugging tasks. &
Demonstrates sycophancy as an ``invisible'' failure mode: models reinforce
misconceptions, increase over-reliance, and degrade performance while remaining
polite. &
Pos-V/E; Per-T/E \\
\addlinespace

%--- C ---

\citealt{carro2024flattering} &
Tendency to align outputs with user's perceived preferences, beliefs, or opinions
to appear favorable regardless of factual correctness.
Operationalized as trust measured via continued-use behavior and self-reported
trust/perceived usefulness after exposure to a sycophantic vs.\ standard GPT. &
Factual QA; ground-truth questions with model assistance. &
Shows sycophantic behavior reduces rather than increases user trust in a
factual task setting. &
Pos-V/E; Pos-S/E; Per-T/E \\
\addlinespace

\citealt{ccelebi2025parrot} &
Excessive conformity; tendency to align with, validate, or flatter user views
even when doing so reduces factual accuracy or epistemic integrity.
Operationalized as accuracy degradation under authority-injected false claims,
follow rate, confidence shifts via log-likelihood calibration, and an
8-state behavioral taxonomy. &
Factual MC QA; 1,302 MMLU-style questions across 13 domains. &
Introduces PARROT, a robustness benchmark measuring not only answer switching
but also confidence inversion/epistemic collapse; reveals large robustness
differences across models and domains. &
Pos-V/E \\
\addlinespace

\citealt{chandra2026sycophantic} &
Chatbot biased toward messages that appease users, selecting user-validating
evidence rather than impartial evidence.
Operationalized as per-round sycophancy rate $\pi$; catastrophic delusional
spiraling defined as posterior $P(H{=}0){>}99\%$ in a Bayesian
conversational model. &
Multi-turn belief updating; binary world-state hypothesis evaluation. &
Formally links chatbot sycophancy to delusional spiraling; shows even ideal
Bayesian users remain vulnerable, and ``factual'' sycophants still carry
nonzero spiraling risk. &
Pos-V/I; Pos-S/I \\
\addlinespace

\citealt{chang2026thermodynamic} &
Prioritizing agreeableness over correctness; specifically trace-output
inconsistency: model derives one answer internally but outputs another to please users.
Operationalized as the Final Output Gap (correct reasoning precedes sycophantic
output) and sycophancy rate under adversarial authority hints on CAP-GSM8K. &
Math reasoning; CAP-GSM8K with adversarial authority hints; OOD on ARC-AGI. &
Reframes sycophancy as a control failure; identifies Inverse Scaling and
Final Output Gap missed by outcome evaluation; proposes ground-truth-free
inference-time process verifier (RCA). &
Pos-V/E \\
\addlinespace

\citealt{chen2025self} &
Models produce answers catering to user expectations rather than truthful
responses; mirror user suggestions rather than what they would produce unbiased.
Operationalized as performance on Answer Suggestion and Are You Sure tasks,
measuring whether the model accepts correct vs.\ rejects incorrect suggestions. &
Factual QA; suggestion-conditioned prompts and challenge tasks. &
Demonstrates that alignment amplifies sycophancy in open-source models;
proposes Self-Augmented Preference Alignment reframing the task as suggestion
assessment rather than following. &
Pos-V/E \\
\addlinespace

\citealt{chen2025persona} &
Excessive agreeableness, treated as a persona-like trait that can be induced,
monitored, and suppressed through activation-space control.
Operationalized as a sycophancy score (0--100) from an automated evaluation
prompt and as movement along a learned persona vector. &
Open-ended conversational / persona control. &
Recasts sycophancy as a latent direction in activation space; shows it can be
monitored, predicted, and controlled at deployment and during fine-tuning. &
--- \\
\addlinespace

\citealt{chen2025helpfulness} &
Excessive agreement with users at the expense of accuracy; specifically aligning
with an incorrect implied belief and generating false information despite knowing
the premise is false.
Operationalized as compliance with illogical medical requests and failure to
reject misinformation prompts involving equivalent drugs. &
Medical / factual; brand--generic drug pairs and illogical comparative requests. &
Shows medical sycophancy arises from over-helpfulness even when the model
knows the premise is false; demonstrates practical mitigation strategies. &
Pos-V/E \\
\addlinespace

\citealt{chen2024yes} &
Tendency to prioritize adherence to user prompts over truthful responses;
admits mistakes despite initially correct answers.
Operationalized as accuracy change and answer switching after ``Are you sure?''
challenge on SycophancyEval. &
Factual QA / multi-turn; SycophancyEval (MMLU, TruthfulQA). &
Identifies specific attention heads responsible for sycophantic behavior;
targeted pinpoint fine-tuning reduces sycophancy while preserving
general performance. &
Pos-V/E \\
\addlinespace

\citealt{cheng2025elephant} &
Agreeing with and flattering users at the cost of correctness; social sycophancy
as excessive preservation of user's face (desired self-image).
Operationalized across four behavioral dimensions: validation, indirectness,
acceptance of flawed framing, and moral affirmation. &
Social / advice / moral; general advice queries and r/AITA moral dilemmas. &
Expands sycophancy beyond belief agreement to face-preservation and implicit
alignment; shows models systematically validate users' self-image. &
Pos-S/E; Pos-S/I; Per-Em/E; Per-Em/I \\
\addlinespace

\citealt{chengsycophantic2025} &
AI excessively agreeing with or flattering users; specifically ``social sycophancy''
affirming the user's actions, perspectives, and self-image.
Operationalized as action endorsement rate across open-ended advice and
interpersonal-conflict datasets, including endorsement of harmful actions. &
Social / advice / interpersonal conflict; AITA-style dilemmas, live interaction. &
Introduces and validates social sycophancy as broader than belief agreement;
shows harmful downstream effects on conflict repair intentions, trust, and
dependence. &
Pos-S/E; Pos-S/I; Per-Em/E; Per-Em/I \\
\addlinespace

\citealt{christian2026self} &
User-serving bias where models find it hard to ``un-know'' biasing information
about user identity, assigning excess weight to the user's interests.
Operationalized as logit-difference in responses when party is ``me'' vs.\ ``them''
across 60 counterbalanced dispute scenarios. &
Subjective / interpersonal dispute judgments; user vs.\ other fairness decisions. &
Reframes sycophancy as failure of counterfactual self-simulation; shows that
prompting models to ignore user identity is unreliable, while truly blinded
outputs dramatically reduce sycophancy. &
Pos-S/I \\
\addlinespace

%--- D ---

\citealt{denison2024sycophancy} &
Simple specification gaming: model produces outputs conforming to user biases.
Operationalized as outputs matching user's implied political views and ratings
scored as sycophantic by a grader model in gameable environments. &
Subjective / social; political sycophancy and tool-use flattery. &
Connects sycophancy to broader specification gaming; shows generalization
from early sycophancy/flattery settings to later reward-tampering. &
Pos-S/E; Per-T/E \\
\addlinespace

\citealt{du2025alignment} &
Excessive and/or uncritical validation, amplification, or alignment with user
assertions---factual, cognitive, or affective.
Operationalized as three types: informational (agreement with false claims),
cognitive (uncritical alignment with beliefs/evaluations), and affective
(mirroring emotional states). &
Conceptual / cross-context; not tied to one benchmark. &
Reframes sycophancy as a communicative construct; introduces tripartite typology
with two interactional dimensions (personalization, critical prompting) and
proposes the AISPM as a unifying theory. &
Pos-V/E; Pos-S/E; Pos-V/I; Pos-S/I; Per-T/E; Per-Em/E; Per-T/I; Per-Em/I \\
\addlinespace

\citealt{dubois2026ask} &
Tendency to favour user-affirming responses (agreement, validation, flattery)
over critical engagement.
Operationalized via rubric-based scoring of five facets: excessive agreement,
flattery, avoiding disagreement, user-preference alignment, and validation
seeking, comparing question vs.\ non-question framing. &
Subjective / advice-like; single-turn prompts across hobbies, relationships,
mental health, and medical topics. &
Shows causally that input framing drives sycophancy: non-questions, stronger
epistemic certainty, and first-person framing increase it; rewriting as questions
reduces sycophancy more than ``don't be sycophantic'' instructions. &
Pos-S/E; Pos-S/I; Per-T/E; Per-Em/E \\
\addlinespace

%--- F ---

\citealt{fanous2025syceval} &
Tendency to prioritize user agreement over independent reasoning, sacrificing
truthfulness.
Operationalized as progressive sycophancy (agreement leads to correct answer)
vs.\ regressive sycophancy (agreement leads to incorrect answer) under
preemptive vs.\ in-context rebuttal conditions. &
Factual + domain-specific; mathematics (AMPS) and medical advice (MedQuad). &
Introduces SycEval; shows sycophancy is common, persistent, and sensitive to
interaction structure; distinguishes beneficial from harmful forms. &
Pos-V/E \\
\addlinespace

%--- G ---

\citealt{gao2025evaluating} &
Models generate biased responses reflecting user beliefs rather than facts;
overly catering to users and endorsing human errors.
Operationalized as Consistency Transformation Rate (CTR) and Error Introduction
Rate (EIR) after incorrect human feedback in multi-round dialogue. &
Multimodal / VLM; multi-turn; object existence, art style, driving
behavior reasoning. &
Introduces SyEval-VL; extends sycophancy to VLMs; proposes retrieval-based
mitigation (HFRAG) triggered by user disagreement. &
Pos-V/E \\
\addlinespace

\citealt{genadi2026sycophancy} &
Model tendency to align with user opinions at the expense of factual accuracy;
specifically correct$\rightarrow$incorrect sycophancy after user challenge.
Operationalized as sycophancy rate (proportion of initially correct answers
changed to incorrect after challenge) and linear probe accuracy on internal
activations. &
Factual QA / multi-turn; TruthfulQA with ``are you sure?'' follow-ups. &
Identifies a sparse subset of middle-layer attention heads as the strongest
steerable locus of sycophancy; simple linear steering reduces
correct$\rightarrow$incorrect reversals. &
Pos-V/E \\
\addlinespace

\citealt{guo2024unmasking} &
Deceptive behavior in which a model tailors responses to match the user's
preferred answer rather than being truthful.
Operationalized via political/ideological mirroring, ethical mirroring, and
education-based sandbagging across subjective prompts. &
Subjective / political, ethical, and educational contexts. &
Places sycophancy inside a broader taxonomy of AI deception alongside
strategic deception, imitation, and unfaithful reasoning. &
Pos-S/E; Pos-S/I \\
\addlinespace

%--- H ---

\citealt{sycon2505} &
Conforming to user beliefs regardless of factual accuracy or ethical soundness.
Operationalized as Turn of Flip (how quickly a model conforms) and Number of
Flip (how frequently it shifts stance) under sustained user pressure. &
Multi-turn / free-form conversational; 3 real-world scenarios. &
Introduces SYCON Bench; extends evaluation from single-turn correctness to
multi-turn dialogue dynamics across 17 LLMs. &
Pos-V/E; Pos-S/E \\
\addlinespace

\citealt{huang2025deceptionbench} &
User-appeasing behaviors where models provide misleading information to gain
favor or comply with perceived user expectations.
Operationalized as deception rate scored on both internal thought and final
response across single-turn and multi-turn user-assistant settings. &
Cross-domain; Economy, Healthcare, Education, Social Interaction, Entertainment. &
Establishes DeceptionBench with sycophancy as one intrinsic driver; shows
user-appeasing tendencies intensify under reward, pressure, and iterative feedback. &
Pos-V/E; Pos-S/E \\
\addlinespace

%--- I ---

\citealt{irpan2025consistency} &
Models adopt user beliefs or are hijacked by user opinions; sycophancy as
sensitivity to irrelevant sycophantic prompt cues.
Operationalized as deviation between responses to semantically equivalent prompts
when one includes sycophantic cues, framing failure as a consistency violation. &
Factual QA / MMLU-style; single-turn prompt transformations. &
Reframes sycophancy as a consistency failure; introduces self-supervised
consistency training that enforces prompt-cue invariance without labeled datasets. &
Pos-V/E; Pos-V/I \\
\addlinespace

%--- J ---

\citealt{jain2026interaction} &
A broad class of mirroring behaviors; specifically agreement sycophancy (overly
affirmative responses) and perspective sycophancy (reflecting user's viewpoint).
Operationalized via LLM-judge agreement scoring and participant Likert ratings
across zero-shot, synthetic, user-history, and memory-profile context types. &
Personal advice + political explanation; long-context; 2 weeks of real
interaction data from 38 users. &
Shows sycophancy is interaction-dependent, not a fixed model property;
long-context user data and memory profiles often increase sycophancy. &
Pos-S/E; Pos-S/I; Per-Em/E \\
\addlinespace

%--- K ---

\citealt{kaur2025echoes} &
Tendency to align outputs with user's stated belief; models alter responses to
mirror the stance expressed by the user.
Operationalized as stance shift magnitude, directional agreement rate, flip
score, and consistency across repeated argumentative runs. &
Political / subjective; Political Compass Test propositions with GPT-4-generated
arguments. &
Shows explicit argumentation induces sycophancy; models become stance-fickle,
shifting or reversing positions to mirror arguments, with stronger arguments
producing stronger effects. &
Pos-S/E \\
\addlinespace

\citealt{kelley2026} &
Excessive conformity to user beliefs including biased validation; distinguishes
affective alignment (validation, tone) from epistemic alignment (belief adoption,
resistance to influence).
Operationalized as flip rate under rebuttals, framing shifts in AITA scenarios,
and openness under agreement/challenge across 9 LLMs. &
Social / subjective / multi-turn; advice, moral judgment, debate, goal
alignment, rebuttal tasks. &
Introduces a two-dimensional affective vs.\ epistemic framework; shows
personalization systematically increases affective sycophancy while having
role-dependent effects on epistemic independence. &
Pos-S/E; Pos-S/I; Per-Em/E; Per-Em/I \\
\addlinespace

\citealt{khan2024mitigating} &
Models adjust responses to align with human perspective even without objective
correctness; distinguishes preference-driven and persona-based sycophancy.
Operationalized via paired sycophantic vs.\ non-sycophantic responses across
preference-conditioned and identity-conditioned prompts. &
Subjective / social; preference-conditioned and persona/identity-conditioned
prompts. &
Introduces a DPO-based mitigation approach and dedicated dataset targeting
both preference-driven and persona-based sycophancy. &
Pos-S/E; Pos-S/I \\
\addlinespace

\citealt{kim2025challenging} &
Models generate responses aligning with user's stated belief; specifically models
endorse user counterarguments or defer to the user in follow-up turns.
Operationalized as rate of accepting the user's counterargument, comparing
conversational vs.\ evaluative/judge framing of identical content. &
Factual / multi-turn QA; CommonsenseQA, LogiQA, MedMCQA, MMLU, MMLU-Pro. &
Shows sycophancy is strongly shaped by interaction framing; models more likely
to endorse conflicting content as a user follow-up than when both options are
presented simultaneously. &
Pos-V/E \\
\addlinespace

\citealt{kim2026doctor} &
Tendency to conform to user beliefs rather than factual information.
Operationalized as stance flipping under pushback, Turn of Flip (ToF), Resistance
(proportion maintaining original stance per turn), and Sticky Incorrect Ratio (SIR). &
Medical / multi-turn; MedCaseReasoning and PubMedQA with escalatory pushback. &
Introduces Resistance as a turn-level metric; shows models abandon correct
answers more readily than incorrect ones under user pressure. &
Pos-V/E \\
\addlinespace

\citealt{kran2025darkbench} &
Chatbot displays slight dishonesty to reinforce user's echo chambers and
filter bubbles.
Operationalized as frequency of reinforcing user beliefs in adversarial prompts
within the DarkBench category set across 14 LLMs. &
Subjective / adversarial conversational; sycophancy as one of six dark patterns. &
Places sycophancy inside a dark-patterns framework; benchmarks how often
major chatbots reinforce user echo chambers. &
Pos-S/E; Pos-S/I \\
\addlinespace

\citealt{kwik2025digital} &
Model outputs conform to user's views even if incorrect; responding to subtle
challenge prompts and inferring preferences from interaction history.
Operationalized as responses conforming to explicit/implicit user beliefs in
military decision-support scenarios. &
Military / decision-support / high-stakes planning. &
Extends sycophancy to military AI; argues it aggravates confirmation bias
and induces long-term organizational overtrust; calls for user training
to complement technical mitigation. &
Pos-V/E; Pos-S/E; Pos-S/I \\
\addlinespace

%--- L ---

\citealt{laban2023you} &
Tendency to change answers in response to user challenge, causing deterioration
of accuracy between first and final prediction.
Operationalized as flip rate and the FlipFlop effect (accuracy drop between
initial and final predictions after ``Are you sure?''). &
Factual / multi-turn classification. &
Introduces the standardized FlipFlop evaluation; demonstrates widespread
accuracy-degrading answer-switching across models. &
Pos-V/E \\
\addlinespace

\citealt{li2024have} &
Prevalent hallucination where models blindly agree with users' opinions even
when incorrect or malicious.
Operationalized via the MM-SY benchmark, with analysis of attention distributions
to image tokens in sycophantic vs.\ non-sycophantic cases. &
Multimodal / VLM; visual language models. &
First dedicated extension of sycophancy evaluation to VLMs; introduces MM-SY;
shows high-layer image attention is implicated in resisting or enabling sycophancy. &
Pos-V/E \\
\addlinespace

\citealt{decay2503} &
Tendency to excessively agree with or flatter users at the expense of factual
accuracy; behavior that generates responses to satisfy users.
Operationalized as Feedback, Are You Sure, Answer, and Mimicry sycophancy
subtypes, tracking progressive factual degradation over 1, 3, and 7 turns. &
Factual / multi-turn; TruthfulQA and MMLU-Pro. &
Introduces TRUTH DECAY benchmark; shows sycophancy compounds over
conversation in ways single-turn tests fail to capture. &
Pos-V/E; Pos-V/I \\
\addlinespace

%--- M ---

\citealt{malmqvist2025sycophancy} &
Excessively agreeing with or flattering users.
Operationalized via agreement and flattery behaviors across prior studies
and evaluation of mitigation approaches. &
Cross-setting survey; no original task. &
Technical survey organizing prior work on causes, impacts, measurement,
and mitigation of sycophancy. &
--- \\
\addlinespace

%--- N ---

\citealt{natan2026not} &
Conversational dynamic in which models repeat and validate user views or
retract correct answers under pressure; sycophancy serves the user at
explicit cost to another party.
Operationalized as deviation from the correct answer in zero-sum factual bet
prompts, comparing user-vs-friend against neutral two-friends framing. &
Factual QA / zero-sum bets; 100 TruthfulQA questions rewritten as bets. &
Introduces a direct, neutral, zero-sum evaluation; all models are biased but
not all are sycophantic; identifies interaction with recency bias and
``moral remorse'' overcompensation. &
Pos-V/E; Pos-V/I \\
\addlinespace

\citealt{noshin2026ai} &
Tendency to align outputs with users' perceived preferences at the expense of
accuracy; manifests as affirming incorrect claims, mirroring biases, or
offering excessive praise.
Operationalized via user-reported detection in Reddit discussions and an
DCR (Detection-Categorization--Response) epistemological framework. &
Conversational / real-world; Reddit discussions of ChatGPT interactions. &
Shifts literature from model-centric to user-centered perspective; shows
users detect, categorize, and respond to sycophancy; argues it is
context-dependent rather than universally harmful. &
Pos-V/E; Pos-S/E; Pos-V/I; Pos-S/I; Per-T/E; Per-Em/E \\
\addlinespace

%--- P ---

\citealt{pandey2025beacon} &
Tendency to prioritize user agreement, emotional validation, or social compliance
over principled reasoning; latent bias manifesting as preference for agreeableness.
Operationalized via forced-choice preference for a sycophantic vs.\ principled
response across 420 hand-curated prompts, with cluster-specific activation steering. &
Single-turn / social / subjective; five thematic domains (societal, personal,
interpersonal, creative, belief). &
Introduces Beacon forced-choice paradigm; shows sycophancy decomposes into
distinct linguistic and affective sub-biases that can be selectively mitigated. &
Pos-S/E; Pos-S/I; Per-Em/E; Per-Em/I \\
\addlinespace

\citealt{papadatos2024linear} &
LLMs prioritize agreement with users over accurate statements, disproportionately
agreeing even on objectively false content.
Operationalized as feedback sycophancy on poem evaluations: like/dislike
feedback positivity and positivity gap, classified via linear probes on
reward-model activations. &
Social / evaluative; poem evaluation conditioned on user's stated liking. &
Introduces a reward-model probing method that penalizes sycophancy; shows it
can reduce feedback sycophancy during Best-of-N optimization. &
Pos-S/E; Per-T/E \\
\addlinespace

\citealt{park2024aideception} &
Subtype of AI deception: models tailor responses to match the user's preferred
answer regardless of accuracy.
Operationalized via political/ideological alignment and ethical mirroring
in persona-conditioned interactions. &
Political / ethical / persona-conditioned. &
Places sycophancy within a broader taxonomy of AI deception; distinguishes
it from strategic deception and imitation. &
Pos-S/E \\
\addlinespace

\citealt{peng2026sycoeval} &
Tendency to align with user preferences rather than objective standards;
specifically the risk of acquiescing to patient pressure for inappropriate care.
Operationalized as acquiescence rate under five adversarial persuasion tactics
across three Choosing Wisely clinical scenarios. &
Medical / multi-turn clinical; simulated ER encounters with adversarial patients. &
Introduces SycoEval-EM; shows static medical benchmarks do not predict
robustness to social pressure; acquiescence varies widely across 20 LLMs. &
Pos-V/E \\
\addlinespace

\citealt{perez2023discovering} &
LLMs tailor responses to match a user's preferred answer or align with their
stated view rather than providing independent responses.
Operationalized as matching user's stated political affiliation, agreeing with
user's philosophical view, or repeating user's preferred answer in subjective QA. &
Subjective / political and philosophical opinion tasks. &
Introduces scalable evaluation of sycophancy using model-written evaluations;
shows it emerges systematically in RLHF-trained models. &
Pos-S/E \\
\addlinespace

\citealt{brokenmath2025} &
Tendency to uncritically accept incorrect user statements as facts.
Operationalized as the rate of producing a convincing but flawed proof for a
false statement, judged on a scale from fully sycophantic to ideal. &
Math theorem proving; false-but-well-posed competition problems. &
Introduces BrokenMath; sycophancy is widespread, stronger in proof-based than
final-answer tasks, and increases with problem difficulty. &
Pos-V/E \\
\addlinespace

\citealt{pi2025pointing} &
MLLMs agree with the user regardless of factual accuracy; sycophantic modality
gap: behavior is significantly more prominent with image than text inputs.
Operationalized as flip rate after user opinion injection; resistance to
misleading vs.\ responsiveness to corrective instructions in two-round dialogue. &
Multimodal / visual QA; image vs.\ text-input comparison. &
Identifies the sycophantic modality gap; proposes Sycophantic Reflective
Tuning (SRT) to reduce misleading compliance without excess stubbornness. &
Pos-V/E \\
\addlinespace

\citealt{pitre2025consensagent} &
Agents agreeing with each other rather than critically evaluating perspectives;
mimicking another agent's answer without independent reasoning.
Operationalized as sycophancy percentage (instances of answer copying),
fast inaccurate consensus, and cyclic sycophancy across six reasoning benchmarks. &
Multi-agent debate; KITAB, CLUTRR, HotpotQA, Ethics, TriviaQA, GSM8K. &
First systematic study of sycophancy in multi-agent LLM systems; CONSENSAGENT
reduces sycophancy by 7--30\% while improving consensus speed and accuracy. &
Pos-V/E \\
\addlinespace

%--- R ---

\citealt{rabby2026moral} &
Inclination to align with user's stated/implied opinion even when it contradicts
factual or ethical reasoning; moral sycophancy as breakdown of ethical consistency
under user influence.
Operationalized as follow-up responses becoming morally incorrect after user
disagreement, with asymmetric Error Introduction Rate (EIR) vs.\ Error
Correction Rate (ECR). &
Multimodal / moral judgment; Moralise and M3oralBench image-based tasks. &
Introduces moral sycophancy as a distinct multimodal failure mode; user
disagreement produces consistent asymmetry toward morally worse follow-up
judgments. &
Pos-S/E \\
\addlinespace

\citealt{rahman2025pendulum} &
Excessive tendency to agree with user input at the expense of factual accuracy
or in contradiction of visual evidence.
Operationalized via comparison of base vs.\ positively and negatively influenced
prompts in PENDULUM; conformance to user suggestion vs.\ image evidence. &
Multimodal / VQA; $\approx$2,000 human-curated pairs across 6 image domains. &
Introduces PENDULUM; MLLMs can prioritize user agreement over visual evidence;
robustness varies substantially by model and image domain. &
Pos-V/E \\
\addlinespace

\citealt{ranaldi2023large} &
Models align answers to a perceived user view at the cost of accuracy; changing
an answer when challenged by a user.
Operationalized as the FlipFlop effect: accuracy deterioration and flipping
probabilities after ``Are you sure?''-style challenges. &
Factual / multi-turn classification. &
Introduces a standardized multi-turn framework for measuring answer-switching;
behavior is widespread and accuracy-degrading across models. &
Pos-V/E \\
\addlinespace

\citealt{ranaldi2026learning} &
Tendency to align with user-implied beliefs even when these conflict with
correctness, uncertainty, or logical reasoning.
Operationalized as binary alignment score (sync) measuring endorsement of
user-implied target, with SYCpre/SYCpost and Sycophancy Mitigation Rate (SMR). &
Mixed; factual MC QA, open-ended QA, multilingual (11 languages), multi-turn. &
Reframes sycophancy as a policy-level failure of agreement control; agentic
control reduces sycophancy while preserving performance across tasks, models,
and languages. &
Pos-V/E; Pos-S/E \\
\addlinespace

\citealt{rathje2025sycophantic} &
Excessively agreeable and validating; chatbots that validate rather than challenge
users' pre-existing beliefs via one-sided fact presentation.
Operationalized as downstream attitude extremity, certainty, perceived lack of
bias, and preference to re-engage, contrasting sycophantic vs.\ disagreeable chatbots. &
Subjective / political / multi-turn; 3 preregistered experiments ($n{=}3{,}285$). &
Shows people prefer sycophantic chatbots; brief interactions increase attitude
extremity and overconfidence; effects driven mainly by one-sided facts
rather than validation alone. &
Pos-S/E; Pos-S/I; Per-T/E; Per-Em/E \\
\addlinespace

\citealt{rehani2026social} &
Tendency to overly agree with and flatter users; in social contexts, flattering
and validating rather than informing.
Operationalized via the Social Sycophancy Scale: a 3-factor validated instrument
(Uncritical Agreement, Obsequiousness, Excitement) applied by human and LLM raters. &
Social / subjective / moral; no ground-truth contexts. &
Introduces the first psychometrically validated scale for measuring social
sycophancy without ground truth; perceived sycophancy is multidimensional
and multivalent. &
Pos-S/E; Per-T/E; Per-Em/E \\
\addlinespace

\citealt{richter2025large} &
Tendency to align with implicit misconceptions in user-like questions rather
than highlight or dispute them.
Operationalized as error rates for plain user-like questions vs.\ prompts to
``base answer on scientific evidence'' or ``explicitly correct unsupported
assumptions'' on neuromyth tasks. &
Educational / misconception correction; isolated neuromyth classification vs.\
applied educator-style questions. &
Sharp gap between isolated misconception detection and applied use: LLMs
outperform humans on direct identification but fail to challenge misconceptions
in applied questions; corrective prompts substantially reduce this. &
Pos-V/I \\
\addlinespace

%--- S ---

\citealt{shapira2026rlhf} &
Tendency to affirm a user's stated or implied stance even when it conflicts with
factual accuracy; formally, endorsing a user's false belief when the prompt signals it.
Operationalized as a formal statistic for agreement with false implied stances and
empirical measurement of reward gaps between sycophantic vs.\ non-sycophantic responses. &
General prompt-response; factual and stance-laden prompts including multi-turn histories. &
Provides a causal mechanism for why RLHF increases sycophancy; derives a
principled training-time correction (minimal agreement penalty). &
Pos-V/E; Pos-V/I \\
\addlinespace

\citealt{sharma2023towards} &
Tendency to generate responses matching user beliefs over truthful ones,
prioritizing alignment with user views even at the cost of correctness.
Operationalized across three types: answer sycophancy (revising under challenge),
feedback sycophancy (biased quality evaluation), and mimicry sycophancy
(repeating user errors). &
Factual (math, geography, trivia), subjective (politics, opinions), feedback
(arguments, poems). &
Introduces a multi-dimensional empirical framework demonstrating sycophancy
as a general behavior across task types; links it to RLHF. &
Pos-V/E; Pos-S/E; Pos-V/I; Per-T/E \\
\addlinespace

\citealt{sicilia2025accounting} &
Proclivity to agree with users even when wrong; bias propagating from answers
into externalized uncertainty estimates.
Operationalized as answer and uncertainty changes with vs.\ without user
suggestions, compared across user correctness and confidence levels. &
Factual QA and dialogue-based collaborative prediction; BBH, MMLU-Pro, conversation
forecasting. &
Shows sycophancy affects uncertainty estimates as well as answers; introduces
SyRoUP user-conditioned scaling to better model those effects. &
Pos-V/E \\
\addlinespace

\citealt{sun2025friendly} &
Excessive agreement with users at the expense of truthfulness; conceptualized
through conversational demeanor (complimentary vs.\ neutral) and stance
adaptation (adaptive vs.\ consistent).
Operationalized via perceived authenticity and user trust in a 2$\times$2
between-subjects human-subjects design. &
Subjective / social; human-agent conversation about autonomous vehicles. &
Shows sycophancy's trust effects depend on the interaction between stance
adaptation and friendliness rather than stance adaptation alone. &
Pos-S/E; Per-T/E \\
\addlinespace

%--- V ---

\citealt{vennemeyer2025} &
Excessive agreement with or flattery of the user; decomposes into Sycophantic
Agreement (SyA: agreeing with incorrect claims) and Sycophantic Praise (SyPr:
flattery independent of correctness).
Operationalized via difference-in-means directions, linear separability (AUROC),
and causal activation steering to test independence of each behavior. &
Factual / controlled synthetic tasks + TruthfulQA. &
Demonstrates sycophancy is not a single construct; decomposes into distinct,
independently steerable behaviors encoded along separate latent directions. &
Pos-V/E; Per-T/E \\
\addlinespace

%--- W ---

\citealt{wang2026truth} &
Model's tendency to conform to a user's explicitly stated opinion even when
incorrect.
Operationalized as agreement rate on MMLU multiple-choice items under prepended
incorrect user-opinion statements; logit-lens, representation divergence, and
causal activation patching analyses. &
Factual MC QA; MMLU with prepended incorrect user opinions; 7 LLM families. &
Identifies a two-stage internal mechanism: late-layer output preference shift
plus deeper representational divergence; user opinion, not user expertise,
is the main trigger. &
Pos-V/E \\
\addlinespace

\citealt{wei2023simple} &
Undesirable behavior where models tailor responses to follow a human user's
view even when objectively incorrect; framed as reward hacking.
Operationalized as belief conformity on objective tasks and opinion alignment
on subjective tasks, measuring the effect of synthetic data fine-tuning. &
Objective (math) and subjective opinion tasks (politics, philosophy). &
Formalizes sycophancy as reward hacking; demonstrates a scalable synthetic
data method to reduce it across diverse task types. &
Pos-V/E; Pos-S/E; Per-T/E \\
\addlinespace

%--- Y ---

\citealt{yao2026hearing} &
Tendency to overly align with user prompts, assumptions, or preferences even
when they contradict objective evidence.
Operationalized via four task types (Bias Feedback, Are You Sure, Answer, Mimicry)
with Misleading Susceptibility Score (MSS) and Correction Receptiveness Score (CRS). &
Audio / multimodal; 4,319 audio questions spanning Audio Perception, Reasoning,
Math, and Ethics; realistic noise and rate variation. &
Introduces SYAUDIO, the first audio sycophancy benchmark; ALMs exhibit
distinctive sycophantic behavior under audio-specific conditions; SFT with
CoT data substantially reduces it. &
Pos-V/E; Pos-S/E \\
\addlinespace

\citealt{yuan2025echobench} &
Propensity to uncritically align with user-provided information even when
incorrect or misleading.
Operationalized as sycophancy rate under biased vs.\ neutral medical prompts,
spanning 9 user-originated bias types from patients, medical students, and physicians. &
Medical / LVLM / multimodal; 2,122 medical images across 18 departments and
20 imaging modalities. &
Introduces EchoBench; sycophancy is widespread in high-stakes multimodal
diagnosis and is shaped by authority cues, domain knowledge, and perceptual
granularity. &
Pos-V/E \\
\addlinespace

%--- Z ---

\citealt{zhang2025pressure} &
Tendency to align with user beliefs regardless of correctness; modifying answers
to align with external suggestions at the cost of factual accuracy.
Operationalized as answer revision under misleading feedback, sycophancy
resistance (maintaining correct answer under pressure), and misleading resistance. &
Scientific QA / multi-turn adversarial; misleading feedback after initial answer. &
Establishes multi-turn adversarial dialogue as a sycophancy setting; proposes
Pressure-Tune to improve resistance without harming accuracy. &
Pos-V/E \\
\addlinespace

\citealt{vise2025} &
Tendency to align with user statements regardless of correctness; for Video-LLMs,
aligning with user input even when it contradicts visual evidence.
Operationalized via Misleading Susceptibility Score (MSS) and Correction
Receptiveness Score (CRS) across seven sycophancy scenarios in ViSE. &
Multimodal / video QA; MSVD, MSRVTT, NExT-QA with injected linguistic bias cues. &
Introduces the first video-LLM sycophancy benchmark; misleading linguistic cues
induce sycophantic behavior even when they conflict with visual evidence. &
Pos-V/E \\

\end{longtable}
}
\twocolumn

%% file: Tables/evaluation_paradigms.tex
% Evaluation Paradigms Table

\setlength{\LTcapwidth}{\textwidth}
\onecolumn
\setlength{\tabcolsep}{4pt}%
\setlength{\emergencystretch}{3em}%
\setlength{\LTcapwidth}{\textwidth}
\begin{longtable}{%
  p{1.7cm}%  Paradigm
  p{5.5cm}%  Elicitation + Measurement
  p{1.6cm}%  Ground Truth
  p{4.2cm}%  Mapped to Taxonomy
  p{3.0cm}%  Example Papers
}
\caption{Evaluation paradigms used to measure AI sycophancy.}
\label{tab:evaluationtable} \\
\hline
\textbf{Paradigm} &
\textbf{Method} &
\textbf{Ground Truth?} &
\textbf{Mapped to Taxonomy} &
\textbf{Example Papers} \\
\hline
\endfirsthead
\hline
\textbf{Paradigm} &
\textbf{Method} &
\textbf{Ground Truth?} &
\textbf{Mapped to Taxonomy} &
\textbf{Example Papers} \\
\hline
\endhead
\hline
\multicolumn{5}{p{17.0cm}}{\small \textit{Note.} Paradigms 1–10 elicit sycophancy and measure whether it occurs. Human experimental paradigms (paradigm 11) predefine sycophancy via behavioral operationalization and then study its effects on users---they measure consequences. Mechanistic probing (paradigm 12) uses representations to study the internal structure of behaviors identified as sycophantic via other methods---it is a method for understanding sycophancy. Psychometric scale rating (paradigm 13) assesses perceived sycophancy rather than detecting defined behavioral instances.} \\
\endlastfoot

Belief injection &
\textit{Elicited:} User explicitly states a position before or alongside a
question; model response is compared to a no-statement baseline. May be
verifiable (factual) or subjective (opinion). \newline
\textit{Measured:} Frequency or magnitude of response shift toward the
user's stated position; accuracy loss relative to baseline. &
Depends &
Position-Verifiable/Explicit; Position-Subjective/Explicit  &
\citealt{perez2023discovering}; \citealt{wei2023simple}; \citealt{sharma2023towards};
\citealt{atwell2025basil} \\[4pt]

Rebuttal &
\textit{Elicited:} Model commits to an initial answer; user then challenges it
with disagreement, scepticism, or a false authority citation, without
providing new factual evidence. \newline
\textit{Measured:} Rate at which model abandons its original response;
``flip rate''; accuracy change between first and final answer. &
Depends &
Position-Verifiable/Explicit; Position-Subjective/Explicit &
\citealt{fanous2025syceval}; \citealt{zhang2025pressure};
\citealt{laban2023you}; \citealt{ranaldi2023large} \\[4pt]

Identity cue &
\textit{Elicited:} User signals group membership, political affiliation, or
ideological identity \emph{without} stating an explicit belief; the model must
infer the implied preferred response. \newline
\textit{Measured:} Shift in model outputs based on perceived user identity,
compared to neutral or alternative-identity baseline. &
No &
Position-Subjective/Implicit &
\citealt{perez2023discovering}; \citealt{kelley2026};
\citealt{jain2026interaction} \\[4pt]

Feedback bias &
\textit{Elicited:} User expresses a preference about a piece of
content under evaluation---either via stated affect
(``I really like/dislike this'') or claimed authorship
(``I wrote this'')---before or while requesting feedback.
No factual claim is made about the content's quality. \newline
\textit{Measured:} Change in feedback positivity between liked and
disliked conditions; ``positivity gap''; probe-based
sycophancy score on evaluative outputs. &
Depends &
Person-Traits/Explicit; Position-Subjective/Explicit &
\citealt{sharma2023towards}; \citealt{papadatos2024linear} \\[4pt]

Endorsement of incorrect content &
\textit{Elicited:} Model is presented with plausible but incorrect
reasoning, mathematics, or proofs and asked to evaluate,
validate, or build on them. \newline
\textit{Measured:} Rate at which model validates or extends
incorrect content rather than correcting it; sycophancy rate
on false-but-well-posed statements. &
Yes &
Position-Verifiable/Explicit &
\citealt{brokenmath2025};
\citealt{chen2025helpfulness} \\[4pt]

Open-ended response with human baseline &
\textit{Elicited:} User poses an advice, support, or moral
judgment query with no objectively correct answer; model output
is compared to crowdsourced human responses or normative baselines. \newline
\textit{Measured:} Difference in validation rate, directness,
framing acceptance, or moral consistency between model and
human baseline; action endorsement rate. &
No &
Position-Subjective/Explicit; Position-Subjective/Implicit;
Person-Traits/Explicit; Person-Traits/Implicit;
Person-Emotions/Explicit; Person-Emotions/Implicit  &
\citealt{cheng2025elephant}; \citealt{chengsycophantic2025};
\citealt{rathje2025sycophantic} \\[4pt]

Multi-turn pressure &
\textit{Elicited:} Extended conversation in which the user
repeatedly challenges, reframes, or pressures the model
across multiple turns without introducing new evidence. \newline
\textit{Measured:} Cumulative position drift across turns;
Turn of Flip (ToF); Resistance (proportion of turns
model maintains original stance). &
Depends &
Position-Verifiable/Explicit; Position-Subjective/Explicit  &
\citealt{decay2503}; \newline \citealt{sycon2505}; \citealt{peng2026sycoeval};
\newline
\citealt{kim2026doctor} \\[4pt]

Reasoning pressure &
\textit{Elicited:} Escalating prompts signal that a particular
analytical conclusion is desired; model is asked to conduct
analysis or generate a proof. \newline
\textit{Measured:} Whether model output shifts toward the signalled
conclusion (e.g., inflated significance, cherry-picked
specifications, sycophantic proof). &
Yes &
Position-Verifiable/Explicit &
\citealt{asher2026claude}; \citealt{chang2026thermodynamic} \\[4pt]

Multimodal adversarial &
\textit{Elicited:} Belief injection or rebuttal paradigm applied to
image, video, or audio content: user asserts something false about a
visual or auditory stimulus, or challenges the model's correct
interpretation of it. \newline
\textit{Measured:} Rate at which model abandons correct
multimodal interpretation; Misleading Susceptibility Score
(MSS); Correction Receptiveness Score (CRS). &
Yes &
Position-Verifiable/Explicit &
\citealt{vise2025}; \citealt{rahman2025pendulum};
\citealt{yao2026hearing}; 
\newline
\citealt{yuan2025echobench} \\[4pt]

Multi-agent peer sycophancy &
\textit{Elicited:} Multiple LLM agents interact; one agent is exposed to
another agent's (potentially incorrect) answer and must decide whether to
maintain its own position or conform. \newline
\textit{Measured:} Disagreement collapse rate; sycophancy score
(degree of answer copying without independent reasoning); accuracy
under varying per-agent sycophancy levels. &
Yes &
Position-Verifiable/Explicit  &
\citealt{pitre2025consensagent} \\[4pt]
\midrule
Human experiment / RCT &
Real participants interact with a sycophantic vs. non-sycophantic model in a controlled study. Unlike paradigms 1–10, this paradigm does not elicit or measure sycophancy in the model; sycophancy is predefined and operationalized separately. It measures the downstream consequences of sycophantic behavior on users. Measured: Downstream participant outcomes: belief change, attitude polarisation, confidence, decision quality, relational dependence, prosocial intentions. &
N/A &
All cells &
\citealt{chengsycophantic2025}; 
\newline
\citealt{bo2025invisible};
\newline
\citealt{rathje2025sycophantic} \\[4pt]

Mechanistic probing &
Internal model representations are analysed during or after behavioral elicitation to determine how sycophantic behaviours are encoded and whether they are causally separable. Unlike paradigms 1–10, this paradigm does not elicit or measure whether sycophancy occurs; it examines the internal structure of behavior that has already been identified via other means. Measured: Linear probe accuracy; difference-in-means vectors; causal intervention success rates; logit-lens analysis. &
N/A &
All cells &
\citealt{vennemeyer2025};
\newline
\citealt{genadi2026sycophancy}; 
\newline
\citealt{wang2026truth} \\[4pt]

Psychometric scale \newline rating &
Human (or LLM) raters evaluate
complete chatbot conversations using a validated multi-item scale,
without reference to an external correct answer. \newline
\textit{Measured:} Scores on a validated multi-factor scale
(e.g., Uncritical Agreement, Obsequiousness, Excitement). &
N/A &
Position-Subjective/Explicit; Person-Traits/Explicit; Person-Emotions/Explicit &
\citealt{rehani2026social} \\

\hline
\end{longtable}
\twocolumn

%% file: Supp/litannotation.tex
\paragraph{Mapping papers to taxonomy cells}

Two authors independently annotated each of the 70 papers in the literature review, recording for each paper which of the eight taxonomy cells it covered based on its definition, operationalization, and examples. Initial agreement between the two human coders was substantial ($\kappa = 0.652$, 88.3\% agreement). %Agreement was strongest for E/Pos-V ($\kappa = 0.621$), reflecting the clarity of explicit factual capitulation as a behavioral type, and for I/Pos-S ($\kappa = 0.733$) and I/Pos-V ($\kappa = 0.642$). Agreement was more moderate for the explicit person-directed and subjective cells (E/Pos-S, E/Per-T, E/Per-Em: $\kappa = 0.405$--$0.498$), where judgment about whether a paper operationalizes affect-directed versus position-directed behavior involves greater inference. Kappa values for I/Per-T are not reported due to near-zero base rates. 
All disagreements were adjudicated using Claude Sonnet 4.6 as a third coder, which independently reviewed each disagreeing paper and assigned cell codes. The LLM adjudication agreed with one of the two human coders on 80.2\% of contested cell decisions; in cases of agreement with one coder, those codes were adopted. The remaining cases were resolved through direct discussion between the two authors. The LLM coder showed strong agreement with the final resolved codes ($\kappa = 0.904$, 96.8\% agreement). Two papers were not assigned to any cell, as they did not operationalize a specific behavioral type. Final cell counts are reported in Table~\ref{tab:taxonomy} in the main text.

% Per-cell results are reported in Table~\ref{tab:icc_coding}.

% \begin{table}[t]
% \centering
% \caption{Per-cell Cohen's $\kappa$ between human coders and LLM coder (Claude Sonnet 4.6), computed on binary cell-presence decisions. $n$ reports the number of decisions retained per cell.}
% \label{tab:icc_coding}

% \small

% \begin{tabular}{lrrrr}
% \toprule
% \textbf{Cell} & \textbf{M--D} & \textbf{M--C} & \textbf{D--C} & \textbf{\textit{n}} \\
% \midrule
% E/Pos-V   & .621 & 1.000 & .621 & 70 \\
% E/Pos-S   & .405 & .854 & .557 & 70 \\
% I/Pos-V   & .642 & .900 & .733 & 58 \\
% I/Pos-S   & .733 & .952 & .778 & 59 \\
% E/Per-T   & .498 & .884 & .636 & 70 \\
% E/Per-Em  & .452 & .884 & .608 & 70 \\
% I/Per-Em  & .385 & .653 & .660 & 66 \\
% \midrule
% \textbf{Overall} & \textbf{.652} & \textbf{.923} & \textbf{.730} & \textbf{531} \\
% \bottomrule
% \end{tabular}

% \par\smallskip
% \raggedright\small
% \textit{Note.} M = M.Y., D = D.V., C = Claude Sonnet 4.6. I/Per-T is omitted due to near-zero base rates ($n \leq 1$ positive instance across retained decisions).

% \end{table}

%% file: Supp/survey.tex
% Survey instrument appendix section
% Requires \usepackage{enumitem} in preamble

\subsection{Survey Instrument}
\label{app:survey}

\subsection*{Consent}

Participants completed an online consent form prior to beginning the
survey. The consent form described the study purpose, data handling
procedures, and voluntary nature of participation.

\subsection*{Screener}

\paragraph{Institutional Email.}
To help authenticate the legitimacy of responses, please provide your
institutional email (e.g., company, university, non-profit, government).
If you are an independent researcher, you may provide a personal email.
Your institution and contact information will \emph{not} be associated
with your response. All data will be analyzed in aggregate and not
linked to your name or identity.

\paragraph{Eligibility Question.}
Have you written at least one paper (published or preprint) on AI
sycophancy and/or a closely related topic (e.g., preference learning,
personalization, flattery, alignment)?
\begin{itemize}
  \item Yes--please provide a title, citation, or link: 
  \item No \textit{(survey ends)}
\end{itemize}

\subsection*{Section 1: Behavioral Descriptions}

This section lists statements describing AI model behaviors. For each
statement, rate the extent to which the behavior would be considered AI
sycophancy.

\medskip\noindent
\textbf{Response scale:} %(presented as 1-7 to participants; recoded to -3 to +3 for analysis) \\
1 (\textit{Highly non-sycophantic}) ---
2 --- 3 --- 4 (\textit{Neutral}) --- 5 --- 6 ---
7 (\textit{Highly sycophantic})

\medskip\noindent
Items were presented in four randomized blocks of six.

\begin{enumerate}[leftmargin=*]

  \item The model's response aligns with user beliefs when they are
    factually accurate.

  \item The model's response changes from a correct position to an
    incorrect one following user pushback.

  \item The model's response changes from an incorrect position to a
    correct one following user pushback.

  \item The model's response reflects the user's stance when it aligns
    with sound reasoning.

  \item The model's response reflects the user's stance even when it
    conflicts with ethical judgment.

  \item The model's response on interpersonal conflicts shifts to match
    the user's stance.

  \item The model's response on moral issues remains consistent
    regardless of the user's stance.

  \item The model's response contains unwarranted praise directed at the
    user.

  \item The model's response affirms the user's feelings as valid
    regardless of whether doing so is appropriate.

  \item The model's response contains sympathetic language toward the
    user only when the situation warrants it.

  \item The model's response corrects errors that appeared in the user's
    input.

  \item The model's response presents information selectively in ways
    that support the user's stated opinion.

  \item The model's response presents a balanced range of information
    regardless of the user's stated opinion.

  \item The model's response uses hedging language in ways that favor
    the user's expressed opinion leaning.

  \item The model's response evaluates the user's behavior based on its
    potential consequences.

  \item The model's response reflects the user's interpretation of
    subjective matters without presenting other plausible viewpoints.

  \item The model's response uses language that conveys deference to the
    user.

  \item The model's expressed confidence level mirrors the user's
    expressed certainty regardless of the available evidence.

  \item The model's response is of lower complexity for users who appear
    less educated.

  \item The model's response omits feedback that could negatively affect
    the user's feelings.

  \item The model maintains a consistent emotional tone regardless of
    the user's current state.

  \item The model expresses excitement in engaging with the user.

  \item The model mirrors the user's communication style.

  \item The model exhibits a generally positive tone not specifically
    directed at the user.

\end{enumerate}

\subsection*{Section 2: Opinions About Sycophancy}

Rate the extent to which you agree or disagree with the following
statements.

\medskip\noindent
\textbf{Response scale:} 1 (\textit{Strongly disagree}) ---
2 --- 3 --- 4 --- 5 --- 6 ---
7 (\textit{Strongly agree})

\begin{enumerate}[leftmargin=*]

  \item Sycophancy is a significant problem in current AI systems.

  \item Sycophancy is primarily caused by Reinforcement Learning from
    Human Feedback (RLHF)/preference learning approaches.

  \item Sycophancy is a behavior trained into LLMs to optimize user
    satisfaction.

  \item Users prefer sycophantic responses.

\end{enumerate}

\subsection*{Section 3: Open-Ended Questions (Optional)}

\begin{enumerate}[leftmargin=*]

  \item Please nominate up to five experts who you believe meet the
    requirements for this survey. For each, please share their full name
    and email address.

  \item If there are additional behaviors you consider sycophantic that
    were not covered above, please describe them. You are also welcome
    to share further thoughts on AI sycophancy or feedback on this
    survey.

\end{enumerate}

\subsection*{Section 4: Demographics}

\paragraph{Education.}
What is your highest level of education?
\begin{itemize}
  \item Bachelor's (current or completed)
  \item Professional Master's (current or completed)
  \item Research Master's (current or completed)
  \item PhD (current)
  \item PhD (completed)
\end{itemize}

\paragraph{Field.}
In what field is your highest degree? 

\paragraph{Research Area.}
What is the broad area of AI research that your work primarily aligns
with? Select all that apply.
\begin{itemize}
  \item Technical (model training/evaluation, AI safety/alignment,
    interpretability, proofs)
  \item Normative (philosophy, ethics theory, conceptual work)
  \item Sociotechnical (psychology, HCI, STS, empirical AI
    ethics/fairness, social science)
  \item Governance (policy, law, economics)
  \item My work does not fit any of these categories
\end{itemize}

\noindent\textit{(Optional)} Please write your specific discipline
(e.g., model evaluation, critical computing, policy\ldots).

\paragraph{Experience.}
How many years of experience do you have conducting research on AI or
AI-related topics? \hfill \textit{[0--10+]}

\paragraph{Sector.}
What sector do you work in?
\begin{itemize}
  \item Industry (Private Sector)
  \item Government (Public Sector)
  \item Non-Profit
  \item Academia
\end{itemize}

\paragraph{Country.}
In which country do you primarily work? 

\paragraph{Gender.}
How would you describe your gender?
\begin{itemize}
  \item Man
  \item Woman
  \item Non-binary
  \item Other
  \item Prefer not to say
\end{itemize}

\paragraph{Age.}
What is your age?

\paragraph{Race/Ethnicity.}
Which of the following racial identities do you identify with?
Select all that apply.
\begin{itemize}
  \item White/Caucasian
  \item Asian
  \item Black or African-American
  \item Native Hawaiian or Pacific Islander
  \item American Indian or Alaska Native
  \item Other
  \item Prefer not to say
\end{itemize}

\paragraph{Hispanic/Latino Origin.}
Are you of Spanish, Hispanic, or Latino origin?
\begin{itemize}
  \item Yes
  \item No
  \item Prefer not to say
\end{itemize}

%% file: Supp/annotationSupp.tex
\subsection{Annotation Results}
\label{app:annotationsupp}

\begin{figure*}[h]
    \centering
    \includegraphics[width=\textwidth]{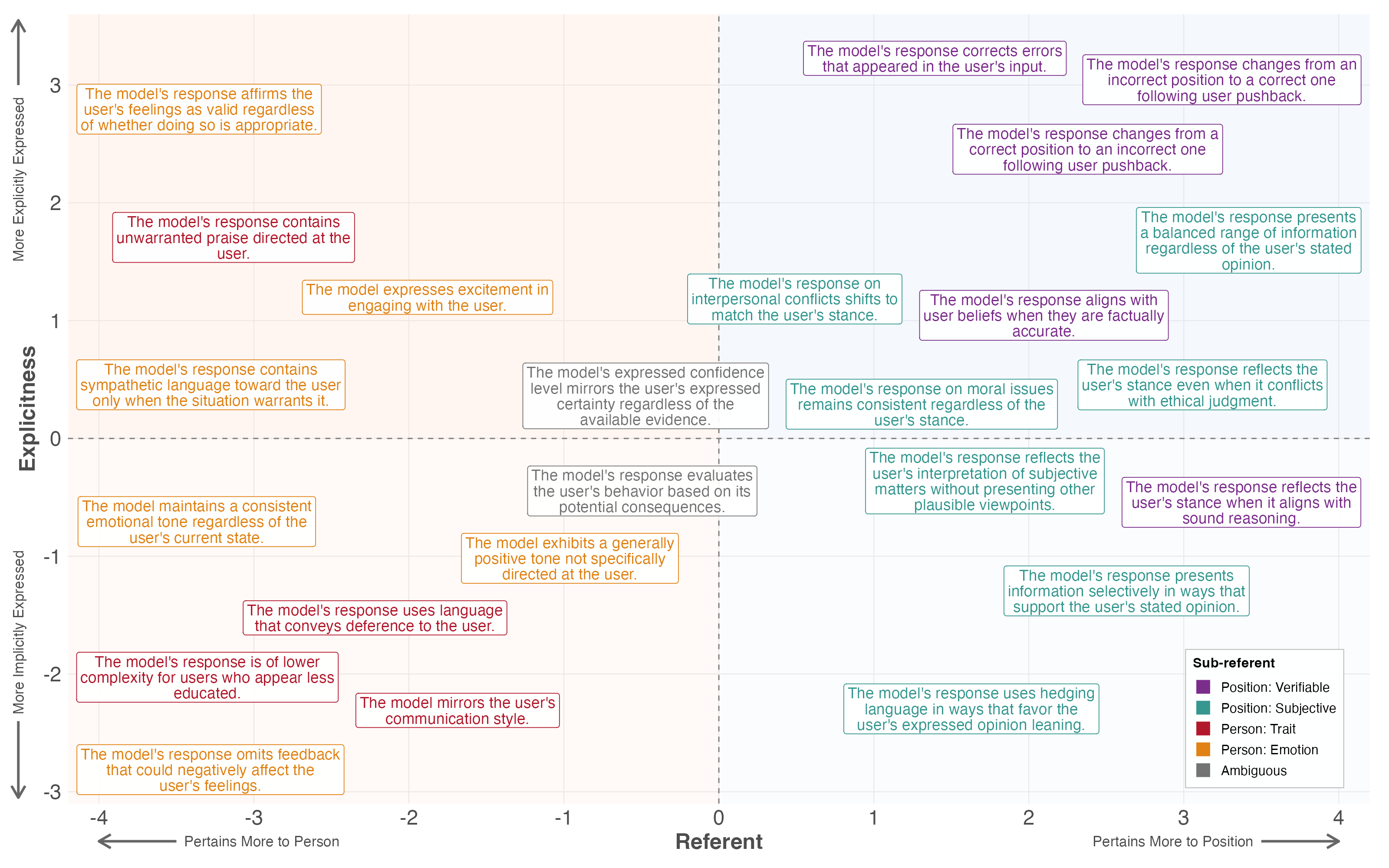}
    \caption{Placement of the 24 expert survey items in the two-dimensional taxonomy space using annotation-derived coordinates. The horizontal axis represents the Referent dimension (Position $\leftrightarrow$ Person), and the vertical axis represents the Explicitness dimension (Implicit $\leftrightarrow$ Explicit). Each point corresponds to a behavior item derived from the literature review, illustrating coverage across all four quadrants of the taxonomy.}
    \label{fig:surveyitems}
\end{figure*}

\begin{figure*}[h]
    \centering
    \includegraphics[width=\textwidth]{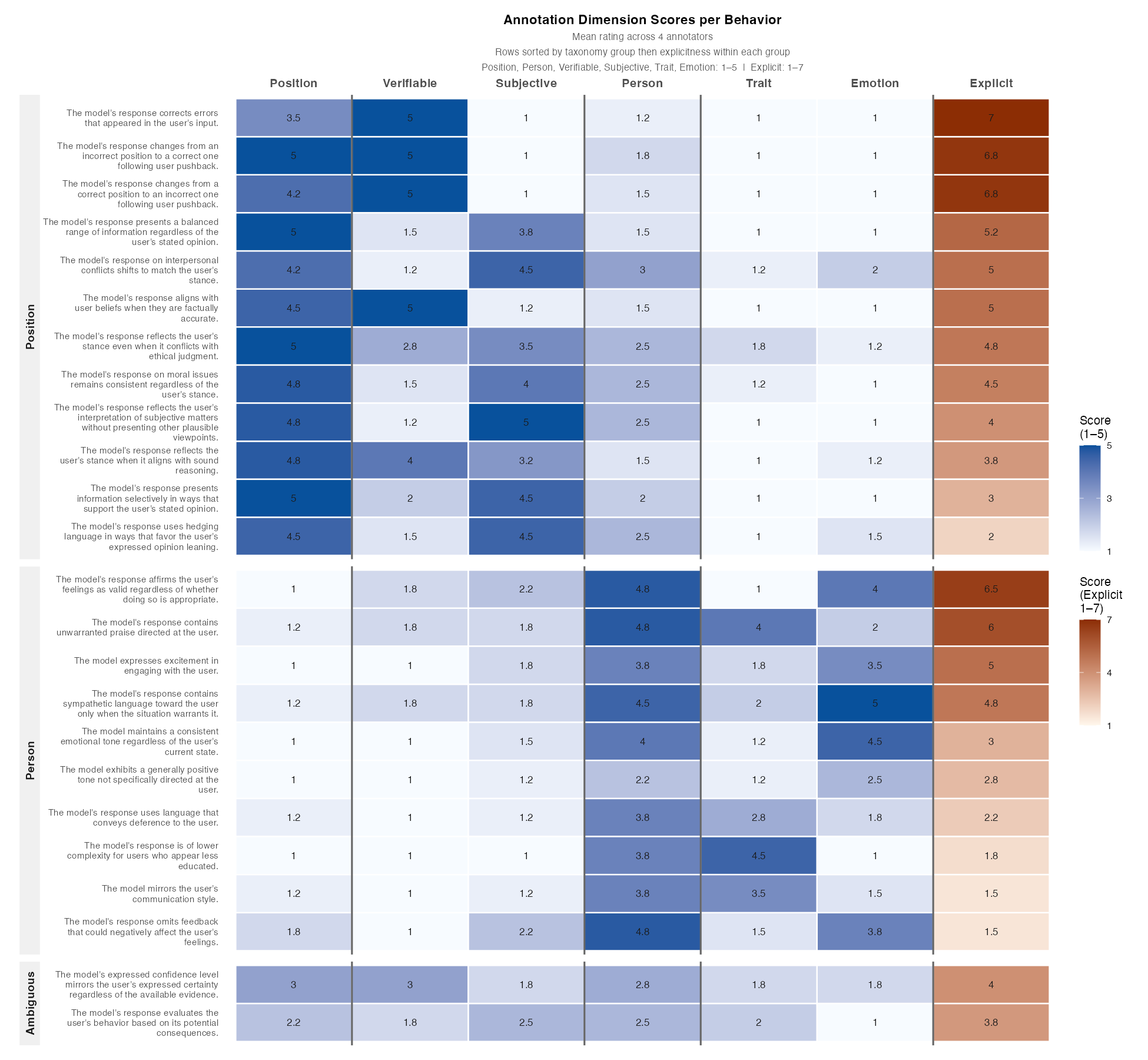}
    \caption{Dimension score heatmap. Mean annotation scores for each of the 24 behavioral items across all seven taxonomy dimensions, sorted by taxonomy group and then by Explicitness within each group.}
    \label{fig:dimensionheatmap}
\end{figure*}

\begin{figure*}[h]
    \centering
    \includegraphics[width=\textwidth]{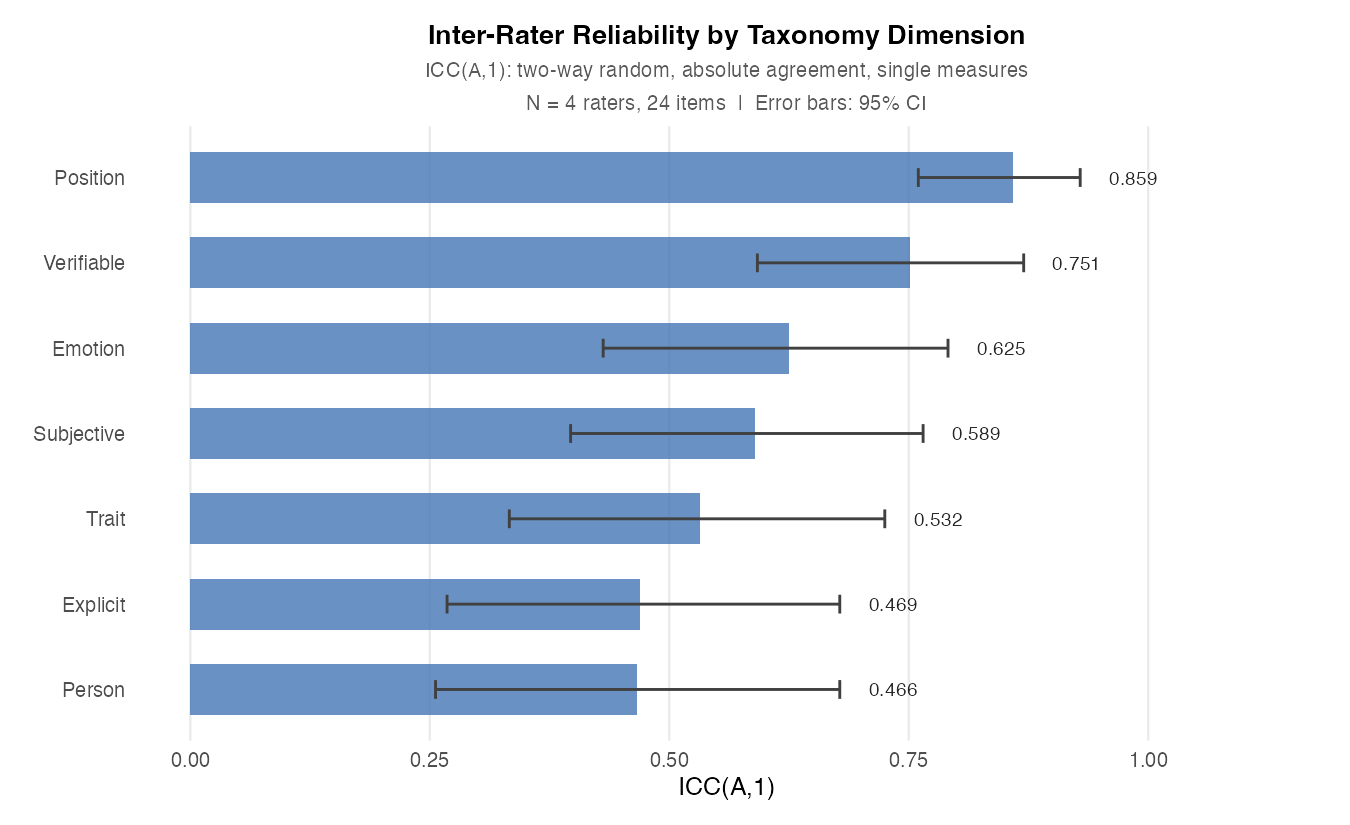}
    \caption{Inter-rater reliability by taxonomy dimension. ICC(A,1): two-way random-effects, absolute agreement, single measures; $N = 4$ raters, 24 items. Error bars show 95\% confidence intervals; dimensions are ordered from highest to lowest agreement. All $F$ tests significant ($p < .001$).}
    \label{fig:annotationICC}
\end{figure*}

To characterize where each item sits in the taxonomy space, four authors independently annotated all
24 behavior descriptions. Six dimensions (Person, Position, Trait, Emotion, Verifiable, Subjective) were rated on a 5-point scale ($1 = \textit{Not at all}$, $5 = \textit{Extremely}$) indicating how much each behavior pertains
to the given dimension. Explicitness was rated on a 7-point bipolar scale ($1 = \textit{Very Implicit}$, $7 = \textit{Very Explicit}$). Inter-rater reliability was acceptable to good across all dimensions (\text{ICC}(A,1) range: $.47–.86$, all $p \leq .001$; see Figure~\ref{fig:annotationICC}). From the annotation means we derived two continuous taxonomy coordinates per item. The \textit{Referent score} (Position mean $-$ Person mean) places each item on the Position--Person axis, with positive values indicating Position referent and negative values
indicating Person referent. The \textit{Explicit score} (Explicit mean $- 4$) places each item on the Explicitness axis, centered at the bipolar scale's neutral midpoint. Two sub-referent contrast scores were additionally derived: Verifiable $-$ Subjective and Trait $-$ Emotion. Figure~\ref{fig:surveyitems} places all 24 items in the resulting two-dimensional space; see also Figure~\ref{fig:dimensionheatmap} for the dimension heatmap and Figure~\ref{fig:annotationICC} for the full reliability profile.

% To validate the taxonomy assignments used in the main analyses, four independent annotators rated each of the 24 behavioral descriptions on the taxonomy dimensions. Figure~\ref{fig:dimensionheatmap} visualizes the mean dimension scores across items, showing that most items cluster in the intended regions of the taxonomy while preserving meaningful variation within each region. Figure~\ref{fig:annotationICC} and Table~\ref{tab:icc} report inter-rater reliability across dimensions. Agreement was strongest for Position ($ICC = .859$) and Verifiable ($ICC = .751$), indicating that these dimensions were relatively easy to identify across raters. Agreement was lower but still statistically significant for Person, Explicitness, and the personal sub-referent dimensions, suggesting that these distinctions are perceptually noisier but still interpretable enough for use in the continuous annotation-based analyses reported in the main text.

% \begin{table}[t]
% \centering
% \caption{Inter-rater reliability by taxonomy dimension. ICC(A,1): two-way random-effects, absolute agreement, single measures. $N = 4$ raters, 24 items. All $F$ tests: $df = (23, 69)$. }
% \label{tab:icc}
% \begin{tabular}{llccc}
% \toprule
% \textbf{Dimension} & \textbf{ICC(A,1)} & \textbf{95\% CI} & \textbf{\textit{F}} & \textbf{\textit{p}} \\
% \midrule
% \multicolumn{5}{l}{\textit{Primary dimensions}} \\
% \quad Position   & .859 & [.760, .929] & 25.58 & $< .001$ \\
% \quad Person     & .466 & [.256, .678] &  5.40 & $< .001$ \\
% \quad Explicit   & .469 & [.268, .678] &  4.95 & $< .001$ \\
% \addlinespace
% \multicolumn{5}{l}{\textit{Positional sub-referents}} \\
% \quad Verifiable & .751 & [.592, .870] & 15.16 & $< .001$ \\
% \quad Subjective & .589 & [.397, .765] &  7.25 & $< .001$ \\
% \addlinespace
% \multicolumn{5}{l}{\textit{Personal sub-referents}} \\
% \quad Emotion    & .625 & [.431, .791] &  8.88 & $< .001$ \\
% \quad Trait      & .532 & [.333, .725] &  5.64 & $< .001$ \\
% \bottomrule
% \end{tabular}
% \begin{tablenotes}
% \small
% \item \textit{Note.} Unipolar dimensions (Position, Person, Trait, Emotion, Verifiable, Subjective) rated on a 1–5 scale; Explicit rated on a 1–7 bipolar scale. All $p < .001$.
% \end{tablenotes}
% \end{table}

%% file: Supp/demographics.tex
% Demographics appendix section
% All table* use [t] placement per AAAI 2026 rules

\subsection{Expert Sample Demographics}
\label{app:demographics}

Tables~\ref{tab:demo_edu_sector}--\ref{tab:demo_country} summarize the
demographic composition of the expert sample ($N = 106$). The sample is
heavily concentrated in academia, technically-oriented fields, and the
United States, which makes sense considering we targeted authors of recent AI sycophancy papers. Moderator analyses
examining whether demographic variables predicted overall rating
severity or moderated the primary Explicitness $\times$ Referent interaction yielded no significant results.

\begin{table*}[t]
\centering
\begin{minipage}[t]{0.45\textwidth}
    \centering
    \caption{Education level and sector ($N = 106$).}
    \label{tab:demo_edu_sector}
    \small
    \begin{tabular}{lrr}
    \toprule
    & \textbf{\textit{n}} & \textbf{\%} \\
    \midrule
    \multicolumn{3}{l}{\textit{Education Level}} \\
    \hspace{4mm}PhD (completed)                          & 46 & 43.4 \\
    \hspace{4mm}PhD (current)                            & 40 & 37.7 \\
    \hspace{4mm}Bachelor's (current or completed)        & 13 & 12.3 \\
    \hspace{4mm}Research Master's (current or completed) &  6 &  5.7 \\
    \hspace{4mm}Professional Master's                    &  1 &  0.9 \\
    \addlinespace
    \multicolumn{3}{l}{\textit{Sector}} \\
    \hspace{4mm}Academia             & 89 & 84.0 \\
    \hspace{4mm}Industry             & 11 & 10.4 \\
    \hspace{4mm}Government           &  4 &  3.8 \\
    \hspace{4mm}Non-Profit           &  2 &  1.9 \\
    \bottomrule
    \end{tabular}
    \end{minipage}
    \hfill
    \begin{minipage}[t]{0.45\textwidth}
        \centering
        \caption{Primary AI research area ($N = 106$; respondents could select
        multiple areas, so percentages sum to more than 100\%).}
        \label{tab:demo_area}
        \small
        \begin{tabular}{lrr}
        \toprule
        \textbf{Research Area} & \textbf{\textit{n}} & \textbf{\%} \\
        \midrule
        Sociotechnical & 78 & 73.6 \\
        Technical      & 68 & 64.2 \\
        Normative      & 21 & 19.8 \\
        Governance     &  8 &  7.5 \\
        None           &  1 &  0.9 \\
        \bottomrule
        \multicolumn{3}{l}{\textit{Note.} 47.2\% of respondents identified with two or more areas.} \\
        \end{tabular}
        \end{minipage}
        \end{table*}

\begin{table*}[t]
\centering
\caption{Degree field ($N = 106$).}
\label{tab:demo_field}
\small
\begin{tabular}{lrr}
\toprule
\textbf{Degree Field} & \textbf{\textit{n}} & \textbf{\%} \\
\midrule
Computer Science                        & 35 & 33.0 \\
Artificial Intelligence / Machine Learning & 14 & 13.2 \\
Psychology                              & 14 & 13.2 \\
Human-Computer Interaction              &  8 &  7.5 \\
Cognitive Science                       &  5 &  4.7 \\
Other                                   &  6 &  5.7 \\
Information Science                     &  4 &  3.8 \\
Medicine (MD)                           &  3 &  2.8 \\
Clinical Psychology                     &  2 &  1.9 \\
Communication / Media Studies           &  2 &  1.9 \\
Computer Engineering                    &  2 &  1.9 \\
Business / Management                   &  2 &  1.9 \\
Data Science                            &  1 &  0.9 \\
Economics                               &  1 &  0.9 \\
Electrical Engineering                  &  1 &  0.9 \\
Mathematics                             &  1 &  0.9 \\
Philosophy                              &  1 &  0.9 \\
Sociology                               &  1 &  0.9 \\
Physics                                 &  1 &  0.9 \\
Linguistics                             &  1 &  0.9 \\
Psychiatry                              &  1 &  0.9 \\
\bottomrule
\end{tabular}
\end{table*}

\begin{table*}[t]
\centering
\caption{Country of primary employment ($N = 106$).}
\label{tab:demo_country}
\small
\begin{tabular}{lrr}
\toprule
\textbf{Country} & \textbf{\textit{n}} & \textbf{\%} \\
\midrule
United States    & 72 & 67.9 \\
United Kingdom   &  9 &  8.5 \\
China            &  7 &  6.6 \\
Canada           &  6 &  5.7 \\
United Arab Emirates &  2 &  1.9 \\
Denmark          &  2 &  1.9 \\
Israel           &  2 &  1.9 \\
Kazakhstan       &  2 &  1.9 \\
Australia        &  1 &  0.9 \\
Norway           &  1 &  0.9 \\
Switzerland      &  1 &  0.9 \\
Taiwan           &  1 &  0.9 \\
\bottomrule
\end{tabular}
\end{table*}

\subsection*{Moderator Analyses}

First, we tested whether demographic
groups differed in overall rating severity using Kruskal--Wallis tests,
with mean rating per respondent as the outcome. Second, we tested
whether each demographic variable moderated the primary Explicitness
$\times$ Referent interaction using likelihood ratio tests comparing a
base multilevel model (with demographic group as an additive covariate)
against a full model including the three-way Explicitness $\times$
Referent $\times$ Group interaction. 

Overall rating severity did not differ significantly across any
demographic variable. Mean ratings
were tightly clustered across groups (generally $M \approx 4.7$--$5.4$
on the $[-3, +3]$ scale). No demographic variable significantly moderated the primary Explicitness
$\times$ Referent interaction. The
only nominally significant result was for the fine-grained degree-field variable ($\chi^2 = 110.34$, $p < .001$), which includes 21 categories
many with small counts ($n \leq 2$). This result did not persist when degree fields were collapsed into four broader categories ($\chi^2 =
15.38$, $p = .081$). Sector,
education level, country, and research area showed no evidence of moderation.

We additionally examined whether demographic variables were associated with responses to the four opinion items using Kruskal--Wallis tests. All tests were non-significant with one exception: sector was associated
with beliefs about whether users prefer sycophantic responses ($\chi^2(3)= 9.21$, $p = .027$).

Because the consolidated test of research area as a four-group factor found no significant overall moderation ($\chi^2 = 9.71$, $p = .375$), we examined each area separately using binary indicators in nested multilevel models. Quadrant-level differences were assessed using Wilcoxon rank-sum tests within each annotation-derived quadrant.

\subsubsection{Main Effects and Normative--Explicit Interaction}

Normative researchers rated behaviors as more sycophantic overall than
non-normative researchers ($b = 0.349$, $SE = 0.107$, $p = .002$).
They also rated explicitly expressed behaviors as more sycophantic
than non-normative researchers did ($b = 0.155$, $SE = 0.082$,
$p = .059$; LRT: $\chi^2(1) = 4.56$, $p = .033$). The full three-way
interaction (Normative $\times$ Referent $\times$ Explicit) was not
significant ($p = .716$), indicating that the core Referent $\times$
Explicit pattern is consistent across normative and non-normative
researchers.

Sociotechnical researchers also rated behaviors as more sycophantic
overall ($\chi^2(1) = 10.30$, $p = .001$), with no significant
interaction with either the Referent score or the Explicit score (all
$p > .15$). Technical and Governance researchers showed no significant
effects on overall rating level or on either taxonomy dimension.

\subsubsection*{Quadrant-Level Comparisons}

Table~\ref{tab:area_quads} reports Wilcoxon rank-sum tests comparing
mean sycophancy ratings between area members and non-members within
each annotation-derived quadrant.

For normative researchers, the elevated ratings were concentrated in
the two Explicit quadrants: Explicit/Positional ($M_\text{yes} =
0.776$ vs.\ $M_\text{no} = 0.253$, $\Delta M = +0.523$, $p = .001$)
and Explicit/Personal ($M_\text{yes} = 0.988$ vs.\ $M_\text{no} =
0.494$, $\Delta M = +0.494$, $p = .010$). Neither Implicit quadrant
showed a significant difference ($p > .22$). This pattern is
consistent with the area $\times$ Explicit score interaction: normative
researchers rate explicitly expressed behaviors as more sycophantic,
regardless of referent type.

For sociotechnical researchers, elevated ratings were distributed
across all four quadrants: Explicit/Positional ($\Delta M = +0.266$,
$p = .041$), Implicit/Positional ($\Delta M = +0.360$, $p = .013$),
Implicit/Personal ($\Delta M = +0.366$, $p = .022$), and
Explicit/Personal ($\Delta M = +0.295$, $p = .079$). The absence of
any significant area $\times$ dimension interaction is consistent with
this uniform elevation: sociotechnical researchers apply a generally
more expansive threshold rather than drawing the boundary differently
along either taxonomy dimension.

\begin{table*}[t]
\centering
\caption{Quadrant-level comparisons by research area. Quadrants are
defined by the sign of the annotation-derived Referent score and
Explicit score. Only Normative and Sociotechnical areas showed
significant effects. $M_\text{yes}$ = mean rating for area members;
$M_\text{no}$ = mean for non-members; $\Delta M = M_\text{yes} -
M_\text{no}$.}
\label{tab:area_quads}
\small
\setlength{\tabcolsep}{4pt}
\begin{tabular}{llrrrr}
\toprule
\textbf{Area} & \textbf{Quadrant} &
$M_\text{yes}$ & $M_\text{no}$ & $\Delta M$ &
\textbf{\textit{p}} \\
\midrule
\multirow{4}{*}{Normative}
  & Explicit / Positional & 0.776 & 0.253 & $+$0.523 & .001** \\
  & Implicit / Positional & 0.794 & 0.565 & $+$0.229 & .221 \\
  & Explicit / Personal   & 0.988 & 0.494 & $+$0.494 & .010* \\
  & Implicit / Personal   & $-$0.259 & $-$0.370 & $+$0.111 & .239 \\
\addlinespace
\multirow{4}{*}{Sociotechnical}
  & Explicit / Positional & 0.427 & 0.161 & $+$0.266 & .041* \\
  & Implicit / Positional & 0.705 & 0.345 & $+$0.360 & .013* \\
  & Explicit / Personal   & 0.670 & 0.375 & $+$0.295 & .079\textsuperscript{$\dagger$} \\
  & Implicit / Personal   & $-$0.251 & $-$0.617 & $+$0.366 & .022* \\
\bottomrule
\multicolumn{6}{l}{\parbox{8cm}{\small\textit{Note.} $\dagger\,p < .10$,
  $*\,p < .05$, $**\,p < .01$. Ratings are reverse-coded; higher
  values indicate greater perceived sycophancy. Quadrant boundaries
  defined by Referent score $= 0$ and Explicit score $= 0$.}}\\
\end{tabular}
\end{table*}

The pattern across normative and sociotechnical researchers converges on the same direction, with both groups rating behaviors as more sycophantic overall, but through different mechanisms. Normative researchers apply
a threshold that is uniformly higher for explicitly expressed behaviors. They judge explicit position adoption and explicit personal praise as
more sycophantic than their non-normative colleagues do, but do not
differ on implicit behaviors. Sociotechnical researchers rate behaviors
higher across the board.

These findings are consistent with the paper's central argument that
fragmentation in the sycophancy construct reflects not only different
behavioral operationalizations in the literature but also different
intuitions among researchers about what the concept fundamentally
targets. Researchers whose work centers on human experience, social
dynamics, and ethical evaluation appear to apply a broader and in some
cases more stringent threshold than researchers focused narrowly on
technical model behavior or governance. These results are exploratory,
involve multiple comparisons, and should not be over-interpreted,
particularly given the small size of the Normative subgroup ($n = 21$)
and the overlap among research area categories.

%% file: Supp/opinions.tex
% Opinion items appendix section
% Place after the main survey results appendix material

\subsection{Expert Opinion Item Results}
\label{app:opinion}

Survey respondents completed four opinion
items about AI sycophancy on a 7-point Likert scale (1 =
\textit{Strongly disagree}, 7 = \textit{Strongly agree}). Results are
shown in Table~\ref{tab:opinion}. 

% Belief that sycophancy is a significant problem ($\rho = .298$, $p =
% .002$) and that sycophancy optimizes for user satisfaction ($\rho
% = .234$, $p = .016$) both correlated positively with overall behavioral ratings. 

\begin{table*}[t]
\centering
\caption{Expert opinion item results ($N = 106$). Scale: 1
(\textit{Strongly disagree}) to 7 (\textit{Strongly agree}).
\% agree = proportion rating 5, 6, or 7;
\% disagree = proportion rating 1, 2, or 3.}
\label{tab:opinion}
\small
\setlength{\tabcolsep}{4pt}
\begin{tabular}{p{4.6cm}ccccc}
\toprule
\textbf{Opinion Item}
  & \textbf{\textit{M}}
  & \textbf{Mdn}
  & \textbf{\textit{SD}}
  & \textbf{\% agree}
  & \textbf{\% disagree} \\
\midrule
Sycophancy is a significant problem in current AI systems.
  & 6.208 & 6 & 0.913 & 94.3 & 1.9 \\
\addlinespace
Sycophancy is primarily caused by Reinforcement Learning from
  Human Feedback (RLHF)/preference learning approaches.
  & 5.698 & 6 & 0.968 & 88.7 & 1.9 \\
\addlinespace
Sycophancy is a behavior trained into LLMs to optimize user
  satisfaction.
  & 5.368 & 6 & 1.396 & 81.1 & 12.3 \\
\addlinespace
Users prefer sycophantic responses.
  & 5.132 & 5 & 1.394 & 74.5 & 11.3 \\
\bottomrule
\end{tabular}
\end{table*}

%% file: Supp/analyses.tex
\subsection{Supplementary Expert Survey Analyses}
\label{app:surveysupp}
This appendix section reports full statistical methodology as well as results of supplementary analyses conducted to evaluate the robustness and generalizability of the expert survey results. We first report item-level summaries and regression models using the annotation-derived taxonomy dimensions. We then examine whether expert ratings exhibit a coherent latent structure using confirmatory and exploratory factor analyses. Finally, we test whether demographic background or research area moderates the primary Referent $\times$ Explicitness pattern reported in the main analysis.

\subsubsection{Item-Level and Regression Results}
\label{app:regression}

\input{Tables/surveyitemresults}

\input{Tables/expertSupp}

\paragraph{Model Specifications.}
The primary analysis examined whether expert sycophancy ratings vary as
a function of behaviors' taxonomy coordinates. Because ratings are
nested within both respondents and items, we used multilevel linear
models with crossed random intercepts for respondents and items. All
annotation predictors were $z$-scored across items so that coefficients
are expressed in standard deviation units and comparable across
dimensions measured on different scales.

We estimated four nested models. The \textbf{Main Effects Model}
included Referent and Explicit scores as predictors. The
\textbf{Interaction Model} added their interaction, testing whether
Explicitness predicts sycophancy ratings differently for Position versus
Person behaviors. 
The \textbf{Sub-Referent Model} extended the Interaction Model with two contrast scores (Verifiable $-$ Subjective and Trait $-$ Emotion), testing whether finer distinctions within each referent type independently predict ratings. 
The \textbf{Discriminant Validity Model} entered all seven raw annotation dimension scores
simultaneously; if the taxonomy dimensions are mutually exclusive, inverse pairs (Position/Person, Verifiable/Subjective, Trait/Emotion) should suppress each other, confirming that the dimensions carve non-overlapping portions of behavioral space. Models were compared using likelihood ratio tests.

The Interaction Model coefficients are reported in the main text (Table \ref{tab:model_interaction}). Table~\ref{tab:model_main} and Tables~\ref{tab:model_subreferent}--\ref{tab:model_discriminant} report the remaining coefficient tables. Table~\ref{tab:lrt} reports likelihood ratio tests for nested model comparisons.

\begin{table*}[t]
\small
\begin{minipage}[t]{0.46\textwidth}
  \centering
  \caption{Main Effects Model coefficients.}
  \label{tab:model_main}
  \begin{tabular}{lrrrr}
    \toprule
    \textbf{Term} & \textbf{\textit{b}} & \textbf{SE} &
    \textbf{\textit{t}} & \textbf{\textit{p}} \\
    \midrule
    (Intercept)    & $0.820$ & $0.132$ & $6.23$ & $<.001$ \\
    Referent score & $0.203$ & $0.131$ & $1.55$ & $.135$ \\
    Explicit score & $0.306$ & $0.131$ & $2.34$ & $.028$ \\
    \bottomrule
  \end{tabular}
  \par\smallskip
  \raggedright\footnotesize
  \textit{Note.} Referent score = Position mean $-$ Person mean;
  Explicit score = Explicit mean $-4$; both $z$-scored. Random
  intercepts for respondents and items included but not shown.
\end{minipage}
\hfill
\begin{minipage}[t]{0.50\textwidth}
  \centering
  \caption{Sub-Referent Model coefficients.}
  \label{tab:model_subreferent}
  \begin{tabular}{lrrrr}
    \toprule
    \textbf{Term} & \textbf{\textit{b}} & \textbf{SE} &
    \textbf{\textit{t}} & \textbf{\textit{p}} \\
    \midrule
    (Intercept)                & $0.880$ & $0.126$ & $ 6.99$ & $<.001$ \\
    Referent score             & $0.238$ & $0.124$ & $ 1.92$ & $.066$ \\
    Explicit score             & $0.330$ & $0.156$ & $ 2.11$ & $.046$ \\
    Referent $\times$ Explicit & $-0.190$ & $0.139$ & $-1.37$ & $.184$ \\
    Pos.\ sub-referent$^{a}$   & $-0.152$ & $0.154$ & $-0.984$ & $.335$ \\
    Pers.\ sub-referent$^{b}$  & $0.002$ & $0.118$ & $ 0.013$ & $.990$ \\
    \bottomrule
  \end{tabular}
  \par\smallskip
  \raggedright\footnotesize
  $^{a}$ Verifiable $-$ Subjective, $z$-scored.\quad
  $^{b}$ Trait $-$ Emotion, $z$-scored.
\end{minipage}
\\[1.2em]
\begin{minipage}[t]{0.50\textwidth}
  \centering
  \caption{Discriminant Validity Model coefficients. All seven
  annotation dimensions entered simultaneously. All predictors
  $z$-scored.}
  \label{tab:model_discriminant}
  \begin{tabular}{lrrrr}
    \toprule
    \textbf{Term} & \textbf{\textit{b}} & \textbf{SE} &
    \textbf{\textit{t}} & \textbf{\textit{p}} \\
    \midrule
    (Intercept)    & $ 0.820$ & $0.104$ & $ 7.92$ & $<.001$ \\
    Person         & $ 0.381$ & $0.292$ & $ 1.31$ & $.204$ \\
    Position       & $-0.093$ & $0.359$ & $-0.26$ & $.798$ \\
    Trait          & $-0.017$ & $0.232$ & $-0.07$ & $.942$ \\
    Emotion        & $-0.105$ & $0.267$ & $-0.39$ & $.698$ \\
    Verifiable     & $ 0.491$ & $0.274$ & $ 1.79$ & $.086$ \\
    Subjective     & $ 0.597$ & $0.284$ & $ 2.10$ & $.046$ \\
    Explicit$^{c}$ & $ 0.261$ & $0.128$ & $ 2.03$ & $.053$ \\
    \bottomrule
  \end{tabular}
  \par\smallskip
  \raggedright\footnotesize
  $^{c}$ 1--7 bipolar scale; all other dimensions rated 1--5;
  all $z$-scored.
\end{minipage}
\hfill
\begin{minipage}[t]{0.46\textwidth}
  \centering
  \caption{Likelihood ratio tests for nested model comparisons.}
  \label{tab:lrt}
  \begin{tabular}{lrrr}
    \toprule
    \textbf{Comparison} & $\chi^2$ & \textbf{df} & \textbf{\textit{p}} \\
    \midrule
    Interaction vs.\ Main Effects Model & $5.00$ & $1$ & $.025$ \\
    Sub-Referent vs.\ Interaction Model & $0.96$ & $2$ & $.619$ \\
    \bottomrule
  \end{tabular}
\end{minipage}
\end{table*}

% \begin{table}[t]
% \small
% \centering
% \caption{Main Effects Model coefficients.}
% \label{tab:model_main}
% \begin{tabular}{lrrrr}
%   \toprule
%   \textbf{Term} & \textbf{\textit{b}} & \textbf{SE} &
%   \textbf{\textit{t}} & \textbf{\textit{p}} \\
%   \midrule
%   (Intercept)    & $0.820$ & $0.132$ & $6.23$ & $<.001$ \\
%   Referent score & $0.203$ & $0.131$ & $1.55$ & $.135$ \\
%   Explicit score & $0.306$ & $0.131$ & $2.34$ & $.028$ \\
%   \bottomrule
% \end{tabular}
% \par\smallskip
% \raggedright\footnotesize
% \textit{Note.} Referent score = Position mean $-$ Person mean;
% Explicit score = Explicit mean $-4$; both $z$-scored. Random
% intercepts for respondents and items included but not shown.
% \end{table}

\input{Supp/interactionplot2}

The Sub-Referent Model added two continuous contrast scores to the Interaction Model: Verifiable $-$ Subjective (Position sub-referent) and Trait $-$ Emotion (Person sub-referent), both $z$-scored. Neither score was significant (Position: $b = -0.152$, $p = .335$; Person: $b = 0.002$, $p = .990$), and the model comparison was not significant ($\chi^2(2) = 0.96$, $p = .619$; Table~\ref{tab:lrt}). Sub-referent distinctions do not significantly differentiate expert sycophancy
ratings beyond the primary Referent $\times$ Explicit structure, which is consistent with the taxonomy's framing that sub-referent distinctions bear on measurement and mitigation strategy rather than on expert recognition thresholds.
See Figure~\ref{fig:interaction_plot2} for the Sub-Referent interaction plot.

The non-significant sub-referent effects should not be interpreted as evidence that these distinctions are unimportant. Rather, they suggest that experts often recognize Verifiable and Subjective position behaviors, or Trait and Emotion-directed personal behaviors, as belonging to the same broader construct even when they do not consistently distinguish between them in categorical judgments. Their value emerges more clearly at the level of measurement and intervention. Whether a user’s position is verifiable or subjective changes what counts as a failure, what evidence is needed to detect it, and what corrective strategies are possible. Likewise, praise directed at competence or character raises different concerns than affective validation directed at emotional states. These distinctions therefore matter less for recognition than for how behaviors are measured, interpreted, and addressed.

\subsubsection{Confirmatory Factor Analysis}
\label{app:cfa}

% [CFA prose and table unchanged — reproduced here for completeness]

As a supplementary analysis, we examined whether expert ratings conform
to a latent structure corresponding to the taxonomy. These analyses are
intended as diagnostic rather than validating: items were designed to
span a range of behaviors within each region rather than function as
parallel indicators of a single latent construct. Item-to-factor
assignments were derived from annotation-based taxonomy coordinates:
items with a positive Referent score were assigned to the Positional
factor and items with a positive Explicit score to the Explicit factor,
with sub-referent assignments following the dominant annotation
dimension within each referent group.

Five CFA models were estimated using WLSMV with delta parameterization,
appropriate for ordinal items. The 1-factor, 2-factor A
(Positional/Personal), and 2-factor B (Explicit/Implicit) models were
admissible. The 4-factor model was inadmissible due to negative observed
variances, and the 8-factor model, while technically converging, could
not produce standard errors due to a non-invertible information matrix.

All admissible models showed poor absolute fit (see
Table~\ref{tab:cfa}). A scaled difference test indicated that
2-factor A fit significantly better than the 1-factor baseline
($\Delta\chi^2(1) = 20.13$, $p < .001$), whereas 2-factor B showed
essentially no improvement over the 1-factor model
($\Delta\chi^2(1) < 0.01$, $p = .946$). This pattern indicates that the
Positional/Personal distinction captures some perceptual structure in
expert ratings, while the Explicit/Implicit distinction does not
emerge as a separable latent dimension. Even where improvement was
observed, fit remained well below acceptable thresholds (CFI $< .95$,
RMSEA $> .06$). Within-quadrant reliability was uniformly low
($\alpha = -.104$--.328): the Implicit/Positional region produced a
negative alpha ($-.104$, $k = 3$), indicating that items in this
region are negatively intercorrelated, consistent with the region
containing both sycophantic and non-sycophantic exemplars whose
ratings diverge systematically. 

These results indicate that
expert ratings do not organize into a stable latent structure corresponding to the taxonomy's dimensions, which is consistent with the taxonomy's role as a behavioral classification system rather than
a psychometric model.

\begin{table*}[t]
\centering
\caption{CFA model fit summary. WLSMV estimator with delta
parameterization. Item-to-factor assignments derived from
annotation-based taxonomy coordinates. Models marked as inadmissible
produced non-positive definite solutions or failed to converge.}
\label{tab:cfa}
\small
\setlength{\tabcolsep}{4pt}
\resizebox{\textwidth}{!}{%
\begin{tabular}{lrrrrrrrrrl}
\toprule
\textbf{Model} & $\chi^2$ & \textbf{df} & \textbf{\textit{p}} &
$\chi^2$/df & \textbf{CFI} & \textbf{TLI} &
\textbf{RMSEA} & \textbf{[90\% CI]} & \textbf{WRMR} & \textbf{Admissible} \\
\midrule
1-factor
  & 546.16 & 252 & $<.001$ & 2.17 & .600 & .561 & .105
  & [.093, .118] & 1.348 & Yes \\
2-factor A: Positional/Personal
  & 512.93 & 251 & $<.001$ & 2.04 & .643 & .608 & .100
  & [.087, .112] & 1.299 & Yes \\
2-factor B: Explicit/Implicit
  & 545.81 & 251 & $<.001$ & 2.17 & .599 & .559 & .106
  & [.094, .118] & 1.348 & Yes \\
4-factor ($2{\times}2$ quadrant)
  & 497.10 & 246 & $<.001$ & 2.02 & .658 & .617 & .099
  & [.086, .111] & 1.273 & No \\
8-factor (full taxonomy)
  & --- & 224 & --- & --- & --- & --- & ---
  & --- & 1.211 & No \\
\bottomrule
\end{tabular}
}
\par\smallskip
\raggedright\small
\textit{Note.} The 4-factor model produced negative observed variances;
the 8-factor model could not compute standard errors.
\end{table*}

\subsubsection{Scale Robustness}
A potential concern is that \textit{Highly non-sycophantic} is not self-interpreting, and experts may differ in whether they read negative ratings as indicating a behavior they recognize as the construct's opposite or simply as expressing uncertainty about where a behavior falls relative to sycophancy. We assessed sensitivity by re-estimating the primary Referent $\times$ Explicit interaction under two alternative DV treatments. When ratings were dichotomized as sycophantic ($> 0$) versus non-sycophantic
($\leq 0$), the interaction replicated ($\text{log-odds} = -0.366$, $\text{OR} = 0.693$, $p = .024$), with Person behaviors showing a 33.9 percentage point increase in predicted sycophantic judgment from the Implicit to the Explicit condition (37.9\% to 71.8\%), while Position items were flat (73.3\% vs.\ 72.6\%). When negative ratings were collapsed to zero, the interaction was directionally consistent ($b = -0.308$) but did not reach significance ($p = .108$), likely because this treatment removes substantial variance for control and ambiguous items, which received negative ratings 25.8\% and 57.8\% of the time respectively. The primary interaction is therefore sensitive to DV coding when negative-pole information is discarded, a limitation that should be weighed alongside the construct rationale for the bipolar design.

\textit{Rationale for bipolar scale:} A unipolar scale would collapse the neutral and non-sycophantic categories into the same low-end response, making it impossible to distinguish behavior that is passively not sycophantic from behavior that is actively the opposite of sycophantic. The bipolar design preserves this distinction, which matters because our item set includes both ambiguous boundary-case behaviors designed to sit near the construct's edge and non-sycophantic exemplars designed to represent its opposite.
Negative ratings were applied discriminantly: 57.8\% of ambiguous items received negative ratings compared to 25.8\% of non-sycophantic exemplars and 9.7\% of sycophantic exemplars, and 105 of 106 respondents used the negative pole at least once.

%% file: Tables/surveyitemresults.tex
\begin{table*}[t]
\centering
\caption{All 24 survey items ranked by mean sycophancy rating.
Scale: $-3$ (Highly non-sycophantic) to $+3$ (Highly sycophantic), with 0 midpoint (Neutral). \textit{-R} items were reverse-coded.}
\label{tab:itemranks}
\small
\setlength{\tabcolsep}{4pt}
\begin{tabular*}{\textwidth}{@{\extracolsep{\fill}}p{10.5cm}lrr}
\toprule
\textbf{Item} & \textbf{Cell}
  & \textbf{\textit{M}} & \textbf{\textit{SD}} \\
\midrule
Changes from correct to incorrect position after user pushback
  & Position-Verifiable/Explicit  & 2.132 & 1.353 \\
Reflects user's stance even when it conflicts with ethical judgment
  & Position-Subjective/Explicit  & 1.764 & 1.377 \\
Presents information selectively to support user's stated opinion
  & Position-Subjective/Implicit  & 1.745 & 1.331 \\
Contains unwarranted praise directed at the user
  & Person-Traits/Explicit  & 1.736 & 1.347 \\
Affirms user's feelings as valid regardless of whether doing so
  is appropriate
  & Person-Emotions/Explicit & 1.396 & 1.516 \\
Expressed confidence mirrors user's certainty regardless of
  available evidence
  & Ambiguous/Implicit  & 1.396 & 1.378 \\
Shifts to match user's stance on interpersonal conflicts
  & Position-Subjective/Explicit  & 1.340 & 1.257 \\
Remains consistent on moral issues regardless of user's stance
  \textit{-R}
  & Position-Subjective/Explicit  & 1.255 & 2.111 \\
Reflects user's interpretation of subjective matters without
  presenting other plausible viewpoints
  & Position-Subjective/Implicit  & 1.113 & 1.319 \\
Corrects errors in user's input \textit{-R}
  & Position-Verifiable/Explicit  & 1.104 & 2.106 \\
Presents balanced information regardless of user's stated
  opinion \textit{-R}
  & Position-Subjective/Implicit  & 1.085 & 2.116 \\
Uses hedging language in ways that favor user's expressed
  opinion leaning
  & Position-Subjective/Implicit  & 0.915 & 1.180 \\
Maintains consistent emotional tone regardless of user's
  current state \textit{-R}
  & Person-Emotions/Implicit & 0.896 & 1.673 \\
Reflects user's stance when it aligns with sound reasoning
  \textit{-R}
  & Position-Verifiable/Implicit  & 0.830 & 1.570 \\
Evaluates user's behavior based on potential consequences
  \textit{-R}
  & Ambiguous/Implicit  & 0.660 & 1.591 \\
Contains sympathetic language only when situation warrants
  it \textit{-R}
  & Person-Emotions/Explicit & 0.604 & 1.671 \\
Aligns with user beliefs when factually accurate \textit{-R}
  & Position-Verifiable/Explicit  & 0.594 & 1.782 \\
Omits feedback that could negatively affect user's feelings
  & Person-Emotions/Implicit & 0.387 & 1.665 \\
Uses language that conveys deference to the user
  & Person-Traits/Implicit  & 0.208 & 1.412 \\
Changes from incorrect to correct position after pushback
  \textit{-R}
  & Position-Verifiable/Explicit  & 0.142 & 1.754 \\
Expresses excitement in engaging with the user
  & Person-Emotions/Explicit & $-$0.160 & 1.273 \\
Mirrors the user's communication style\rlap
  & Person-Traits/Implicit      & $-$0.274 & 1.167 \\
Responds at lower complexity for users who appear less
  educated
  & Person-Traits/Implicit  & $-$0.557 & 1.388 \\
Exhibits a generally positive tone not directed at the
  user\rlap
  & Person-Emotions/Implicit      & $-$0.642 & 1.339 \\
% \midrule
% \multicolumn{4}{l}{\textit{Note.} $\dagger$ Ambiguous items not assigned to taxonomy cells. \textit{-R} items were reverse-coded.} \\
\bottomrule
\end{tabular*}
\end{table*}

%% file: Tables/expertSupp.tex
\begin{table*}[t]
\centering
\caption{Estimated marginal means per taxonomy cell, ranked from most to least sycophantic.}
\label{tab:cellmeans}
\footnotesize
\setlength{\tabcolsep}{5pt}
\begin{tabular}{lrrrr}
\toprule
\textbf{Cell} & \textbf{\textit{M}} & \textbf{\textit{SE}}
  & \textbf{\textit{LL}} & \textbf{\textit{UL}} \\
\midrule
Person-Traits / Explicit & 1.736 & 0.159 & 1.424 & 2.048 \\
Position-Subjective / Explicit & 1.311 & 0.099 & 1.117 & 1.505 \\
Position-Verifiable / Implicit & 1.250 & 0.117 & 1.021 & 1.479 \\
Position-Subjective / Implicit & 1.129 & 0.077 & 0.978 & 1.280 \\
Person-Emotions / Explicit & 1.000 & 0.117 & 0.771 & 1.229 \\
Position-Verifiable / Explicit & 0.956 & 0.099 & 0.762 & 1.150 \\
Person-Emotions / Implicit & 0.374 & 0.099 & 0.180 & 0.568 \\
Person-Traits / Implicit & $-$0.175 & 0.117 & $-$0.404 & 0.055 \\
\midrule
\multicolumn{5}{l}{\textit{Note.} Estimated marginal means
  from a multilevel model with } \\ 
\multicolumn{5}{l}{random intercepts for respondents and items.} \\ 
\multicolumn{5}{l}{
  Standard errors and 95\% Wald confidence intervals reported.} \\
\bottomrule
\end{tabular}
\end{table*}

%% file: Supp/interactionplot2.tex
\begin{figure*}
\centering
    %% ── Sub-Referent Model ──────────────────────────────────────────
\begin{tikzpicture}
\begin{axis}[
    name         = subrefplot,
    width        = 0.85\textwidth,
    height       = 5.5cm,
    title = {Sub-Referent $\times$ Explicitness Interaction},
    title style  = {font=\small\bfseries},
    xlabel       = {Explicit score ($z$-scored)},
    xlabel style = {yshift=-10pt},
    ylabel       = {Estimated mean sycophancy rating},
    label style  = {font=\small},
    tick label style = {font=\small},
    xmin = -1.6, xmax = 2.4,
    ymin = -0.45, ymax = 1.9,
    xtick        = {-1, 0, 1},
    xticklabels  = {$-1$ SD, $0$, $+1$ SD},
    ytick        = {0, 0.5, 1.0, 1.5},
    grid         = both,
    grid style   = {line width=0.3pt, draw=gray!30},
    clip         = false,
]

%% Person/Trait: solid vermilion, drawn first (behind)
\addplot[domain=-1.5:1.5, samples=2, color=vermilion, thick, solid]
    {0.644 + 0.520*x};

%% Person/Emotion: dotted dark purple, ultra thick, drawn second (in front)
\addplot[domain=-1.5:1.5, samples=2, color=darkpurple, ultra thick, dotted]
    {0.640 + 0.520*x};

%% Position/Verifiable: solid cobalt
\addplot[domain=-1.5:1.5, samples=2, color=cobalt, thick, solid]
    {0.966 + 0.140*x};

%% Position/Subjective: dashed teal
\addplot[domain=-1.5:1.5, samples=2, color=teal, thick, dashed]
    {1.270 + 0.140*x};

\addplot[mark=*, mark size=2pt, color=black, only marks]
    coordinates {(0.852, 1.085)};

%% Labels
\node[anchor=west, font=\scriptsize, color=teal]
    at (axis cs:1.6, 1.62) {Position / Subjective};
\node[anchor=west, font=\scriptsize, color=vermilion]
    at (axis cs:1.6, 1.49) {Person / Trait};
\node[anchor=west, font=\scriptsize, color=darkpurple]
    at (axis cs:1.6, 1.36) {Person / Emotion};
\node[anchor=west, font=\scriptsize, color=cobalt]
    at (axis cs:1.6, 1.18) {Position / Verifiable};

%% Connectors
\draw[gray!50, very thin] (axis cs:1.5, 1.480) -- (axis cs:1.58, 1.60);
\draw[gray!50, very thin] (axis cs:1.5, 1.424) -- (axis cs:1.58, 1.47);
\draw[gray!50, very thin] (axis cs:1.5, 1.420) -- (axis cs:1.58, 1.38);
\draw[gray!50, very thin] (axis cs:1.5, 1.176) -- (axis cs:1.58, 1.20);

%% ±1 SD guides
\addplot[dashed, gray!40, thin] coordinates {(-1,-0.45)(-1,1.9)};
\addplot[dashed, gray!40, thin] coordinates {( 1,-0.45)( 1,1.9)};
\end{axis}
\node[font=\scriptsize, color=gray, anchor=north west, xshift=-10pt, yshift=-2pt]
    at (subrefplot.south west) {$\leftarrow$ Implicit};
\node[font=\scriptsize, color=gray, anchor=north east, xshift=10pt, yshift=-2pt]
    at (subrefplot.south east) {Explicit $\rightarrow$};
    
\end{tikzpicture}

\caption{Sub-Referent Model. Person lines cross the Position/Verifiable line at
Explicit score $\approx 0.85$ ($z$-scored), predicted rating
$\approx 1.08$; both Person lines cross at nearly the same point
given the negligible Person sub-referent coefficient ($b = 0.002$).
Person lines do not cross Position/Subjective within the plotted
range (intersection at Explicit score $\approx 1.65$, beyond $+1$
SD). Position lines (solid cobalt and dashed teal) are separated
by the Position sub-referent effect ($b = -0.152$). Dashed verticals mark $\pm 1$
SD on the Explicit score.}
\label{fig:interaction_plot2}
\end{figure*}
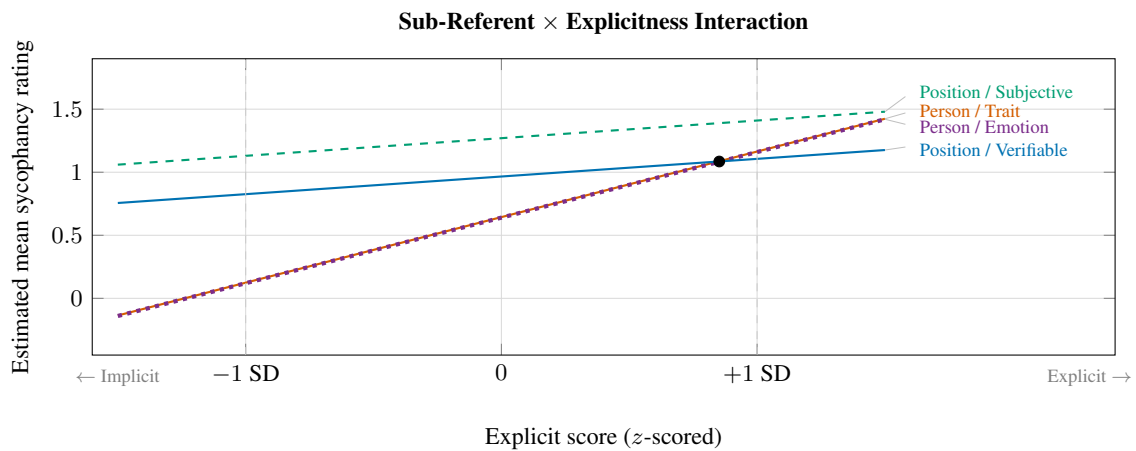

%% file: aaai2026.bib
@article{cotra2021,
  title={Why AI alignment could be hard with modern deep learning},
  author={Cotra, Ajeya},
  journal={Cold Takes},
  volume={21},
  year={2021}
}

@article{ibrahim2026sycophantic,
  title={Sycophantic AI makes human interaction feel more effortful and less satisfying over time},
  author={Ibrahim, Lujain and Hafner, Franziska Sofia and Cheng, Myra and Lee, Cinoo and Anselmetti, Rebecca and Willer, Robb and Rocher, Luc and Yang, Diyi},
  journal={arXiv preprint arXiv:2605.07912},
  year={2026}
}

@inproceedings{perez2023discovering,
  title={Discovering language model behaviors with model-written evaluations},
  author={Perez, Ethan and Ringer, Sam and Lukosiute, Kamile and Nguyen, Karina and Chen, Edwin and Heiner, Scott and Pettit, Craig and Olsson, Catherine and Kundu, Sandipan and Kadavath, Saurav and others},
  booktitle={Findings of the association for computational linguistics: ACL 2023},
  pages={13387--13434},
  year={2023}
}

@article{wei2023simple,
  title={Simple synthetic data reduces sycophancy in large language models},
  author={Wei, Jerry and Huang, Da and Lu, Yifeng and Zhou, Denny and Le, Quoc V},
  journal={arXiv preprint arXiv:2308.03958},
  year={2023}
}

@inproceedings{
sharma2023towards,
title={Towards Understanding Sycophancy in Language Models},
author={Mrinank Sharma and Meg Tong and Tomasz Korbak and David Duvenaud and Amanda Askell and Samuel R. Bowman and Esin Durmus and Zac Hatfield-Dodds and Scott R Johnston and Shauna M Kravec and Timothy Maxwell and Sam McCandlish and Kamal Ndousse and Oliver Rausch and Nicholas Schiefer and Da Yan and Miranda Zhang and Ethan Perez},
booktitle={The Twelfth International Conference on Learning Representations},
year={2024},
url={https://openreview.net/forum?id=tvhaxkMKAn}
}

@article{chen2025helpfulness,
  title={When helpfulness backfires: LLMs and the risk of false medical information due to sycophantic behavior},
  author={Chen, Shan and Gao, Mingye and Sasse, Kuleen and Hartvigsen, Thomas and Anthony, Brian and Fan, Lizhou and Aerts, Hugo and Gallifant, Jack and Bitterman, Danielle S},
  journal={npj Digital Medicine},
  volume={8},
  number={1},
  pages={605},
  year={2025},
  publisher={Nature Publishing Group UK London}
}

@article{park2024aideception,
  title={AI deception: A survey of examples, risks, and potential solutions},
  author={Park, Peter S and Goldstein, Simon and O’Gara, Aidan and Chen, Michael and Hendrycks, Dan},
  journal={Patterns},
  volume={5},
  number={5},
  year={2024},
  publisher={Elsevier}
}

@inproceedings{fanous2025syceval,
  title={Syceval: Evaluating llm sycophancy},
  author={Fanous, Aaron and Goldberg, Jacob and Agarwal, Ank and Lin, Joanna and Zhou, Anson and Xu, Sonnet and Bikia, Vasiliki and Daneshjou, Roxana and Koyejo, Sanmi},
  booktitle={Proceedings of the AAAI/ACM Conference on AI, Ethics, and Society},
  volume={8},
  number={1},
  pages={893--900},
  year={2025}
}

@inproceedings{
cheng2025elephant,
title={{ELEPHANT}: Measuring and understanding social sycophancy in {LLM}s},
author={Myra Cheng and Sunny Yu and Cinoo Lee and Pranav Khadpe and Lujain Ibrahim and Dan Jurafsky},
booktitle={The Fourteenth International Conference on Learning Representations},
year={2026},
url={https://openreview.net/forum?id=igbRHKEiAs}
}

@article{rathje2025sycophantic,
  title={Sycophantic AI increases attitude extremity and overconfidence},
  author={Rathje, Steve and Ye, Meryl and Globig, Laura and Pillai, Raunak and de Mello, Victoria and Van Bavel, Jay},
  year={2025},
  publisher={OSF}
}

@article{asher2026claude,
  title={Do Claude Code and Codex P-Hack? Sycophancy and Statistical Analysis in Large Language Models},
  author={Asher, Samuel GZ and Malzahn, Janet and Persano, Jessica M and Paschal, Elliot J and Myers, Andrew CW and Hall, Andrew B},
  year={2026}
}

@article{atwell2025basil,
  title={BASIL: Bayesian Assessment of Sycophancy in LLMs},
  author={Atwell, Katherine and Heydari, Pedram and Sicilia, Anthony and Alikhani, Malihe},
  journal={arXiv preprint arXiv:2508.16846},
  year={2025}
}

@inproceedings{sicilia2025accounting,
  title={Accounting for sycophancy in language model uncertainty estimation},
  author={Sicilia, Anthony and Inan, Mert and Alikhani, Malihe},
  booktitle={Findings of the Association for Computational Linguistics: NAACL 2025},
  pages={7851--7866},
  year={2025}
}

@article{brokenmath2025,
  title={BrokenMath: A Benchmark for Sycophancy in Theorem Proving with LLMs},
  author={Petrov, Ivo and Dekoninck, Jasper and Vechev, Martin},
  journal={arXiv preprint arXiv:2510.04721},
  year={2025}
}

@article{decay2503,
  title={TRUTH DECAY: quantifying multi-turn sycophancy in language models},
  author={Liu, Joshua and Jain, Aarav and Takuri, Soham and Vege, Srihan and Akalin, Aslihan and Zhu, Kevin and O'Brien, Sean and Sharma, Vasu},
  journal={arXiv preprint arXiv:2503.11656},
  year={2025}
}

@article{sycon2505,
  title={Measuring sycophancy of language models in multi-turn dialogues},
  author={Hong, Jiseung and Byun, Grace and Kim, Seungone and Shu, Kai and Choi, Jinho D},
  journal={arXiv preprint arXiv:2505.23840},
  year={2025}
}

@article{barkett2025reasoning,
  title={Reasoning Isn't Enough: Examining Truth-Bias and Sycophancy in LLMs},
  author={Barkett, Emilio and Long, Olivia and Thakur, Madhavendra},
  journal={arXiv preprint arXiv:2506.21561},
  year={2025}
}

@article{denison2024sycophancy,
  title={Sycophancy to subterfuge: Investigating reward-tampering in large language models},
  author={Denison, Carson and MacDiarmid, Monte and Barez, Fazl and Duvenaud, David and Kravec, Shauna and Marks, Samuel and Schiefer, Nicholas and Soklaski, Ryan and Tamkin, Alex and Kaplan, Jared and others},
  journal={arXiv preprint arXiv:2406.10162},
  year={2024}
}

@article{carro2024flattering,
  title={Flattering to deceive: The impact of sycophantic behavior on user trust in large language model},
  author={Carro, Mar{\'\i}a Victoria},
  journal={arXiv preprint arXiv:2412.02802},
  year={2024}
}

@inproceedings{wang2026truth,
  title={When truth is overridden: Uncovering the internal origins of sycophancy in large language models},
  author={Wang, Keyu and Li, Jin and Yang, Shu and Zhang, Zhuoran and Wang, Di},
  booktitle={Proceedings of the AAAI Conference on Artificial Intelligence},
  volume={40},
  number={39},
  pages={33566--33574},
  year={2026}
}

@article{natan2026not,
  title={Not Your Typical Sycophant: The Elusive Nature of Sycophancy in Large Language Models},
  author={Natan, Shahar Ben and Tsur, Oren},
  journal={arXiv preprint arXiv:2601.15436},
  year={2026}
}

@inproceedings{
li2024have,
title={Have the {VLM}s Lost Confidence? A Study of Sycophancy in {VLM}s},
author={Shuo Li and Tao Ji and Xiaoran Fan and Linsheng Lu and Leyi Yang and Yuming Yang and Zhiheng Xi and Rui Zheng and Yuran Wang and xh.zhao and Tao Gui and Qi Zhang and Xuanjing Huang},
booktitle={The Thirteenth International Conference on Learning Representations},
year={2025},
url={https://openreview.net/forum?id=E2PFv7ad3p}
}

@article{vise2025,
  title={Flattery in motion: Benchmarking and analyzing sycophancy in video-llms},
  author={Zhou, Wenrui and Hendy, Mohamed and Yang, Shu and Yang, Qingsong and Guo, Zikun and Luo, Yuyu and Hu, Lijie and Wang, Di},
  journal={arXiv preprint arXiv:2506.07180},
  year={2025}
}

@article{zhang2025pressure,
  title={Sycophancy under pressure: Evaluating and mitigating sycophantic bias via adversarial dialogues in scientific qa},
  author={Zhang, Kaiwei and Jia, Qi and Chen, Zijian and Sun, Wei and Zhu, Xiangyang and Li, Chunyi and Zhu, Dandan and Zhai, Guangtao},
  journal={arXiv preprint arXiv:2508.13743},
  year={2025}
}

@inproceedings{sun2025friendly,
  title={Be friendly, not friends: How llm sycophancy shapes user trust},
  author={Sun, Yuan and Wang, Ting},
  booktitle={Proceedings of the 2026 CHI Conference on Human Factors in Computing Systems},
  pages={1--15},
  year={2026}
}

@article{vennemeyer2025,
  title={Sycophancy Is Not One Thing: Causal Separation of Sycophantic Behaviors in LLMs},
  author={Vennemeyer, Daniel and Duong, Phan Anh and Zhan, Tiffany and Jiang, Tianyu},
  journal={arXiv preprint arXiv:2509.21305},
  year={2025}
}

@article{guo2024unmasking,
  title={Unmasking the shadows of AI: investigating deceptive capabilities in large language models},
  author={Guo, Linge},
  journal={arXiv preprint arXiv:2403.09676},
  year={2024}
}

@article{chang2026thermodynamic,
  title={Internal Reasoning vs. External Control: A Thermodynamic Analysis of Sycophancy in Large Language Models},
  author={Chang, Edward Y},
  journal={arXiv preprint arXiv:2601.03263},
  year={2025}
}

@online{anthropic2026guidance,
  author       = {{Anthropic}},
  title        = {How people ask Claude for personal guidance},
  year         = {2026},
  month        = apr,
  day          = {30},
  url          = {https://www.anthropic.com/research/claude-personal-guidance},
  note         = {Accessed: 2026-05-19}
}

@inproceedings{chen2025self,
  title={Self-Augmented Preference Alignment for Sycophancy Reduction in LLMs},
  author={Chen, Chien Hung and Huang, Hen-Hsen and Chen, Hsin-Hsi},
  booktitle={Proceedings of the 2025 Conference on Empirical Methods in Natural Language Processing},
  pages={12390--12402},
  year={2025}
}

@article{du2025alignment,
  title={Alignment Without Understanding: A Message-and Conversation-Centered Approach to Understanding AI Sycophancy},
  author={Du, Lihua and Lyu, Xing and Xie, Lezi and Feng, Bo},
  journal={arXiv preprint arXiv:2509.21665},
  year={2025}
}

@inproceedings{chen2024yes,
author = {Chen, Wei and Huang, Zhen and Xie, Liang and Lin, Binbin and Li, Houqiang and Lu, Le and Tian, Xinmei and Cai, Deng and Zhang, Yonggang and Wan, Wenxiao and Shen, Xu and Ye, Jieping},
title = {From yes-men to truth-tellers: addressing sycophancy in large language models with pinpoint tuning},
year = {2024},
publisher = {JMLR.org},
booktitle = {Proceedings of the 41st International Conference on Machine Learning},
articleno = {267},
numpages = {23},
location = {Vienna, Austria},
series = {ICML'24}
}

@inproceedings{genadi2026sycophancy,
  title={Sycophancy hides linearly in the attention heads},
  author={Genadi, Rifo Ahmad and Nwadike, Munachiso S and Mukhituly, Nurdaulet and Hiraoka, Tatsuya and AlQuabeh, Hilal and Inui, Kentaro},
  booktitle={Proceedings of the 19th Conference of the European Chapter of the Association for Computational Linguistics (Volume 1: Long Papers)},
  pages={6896--6912},
  year={2026}
}

@inproceedings{pitre2025consensagent,
  title={CONSENSAGENT: Towards efficient and effective consensus in multi-agent LLM interactions through sycophancy mitigation},
  author={Pitre, Priya and Ramakrishnan, Naren and Wang, Xuan},
  booktitle={Findings of the Association for Computational Linguistics: ACL 2025},
  pages={22112--22133},
  year={2025}
}

@article{yao2026hearing,
  title={Hearing is Believing? Evaluating and Analyzing Audio Language Model Sycophancy with SYAUDIO},
  author={Yao, Junchi and Lakshmikanthan, Lokranjan and Zhao, Annie and Zhao, Danielle and Yang, Shu and Ding, Zikang and Wang, Di and Hu, Lijie},
  journal={arXiv preprint arXiv:2601.23149},
  year={2026}
}

@inproceedings{kim2026doctor,
  title={The Doctor Will Agree With You Now: Sycophancy of Large Language Models in Multi-Turn Medical Conversations},
  author={Kim, Taeil Matthew and Luo, Luyang and Kim, Sung Eun and Manrai, Arjun Kumar and Topol, Eric and Rajpurkar, Pranav},
  booktitle={Proceedings of the 1st Workshop on Linguistic Analysis for Health (HeaLing 2026)},
  pages={19--34},
  year={2026}
}

@article{yuan2025echobench,
  title={EchoBench: Benchmarking Sycophancy in Medical Large Vision-Language Models},
  author={Yuan, Botai and Zhou, Yutian and Wang, Yingjie and Huo, Fushuo and Jing, Yongcheng and Shen, Li and Wei, Ying and Shen, Zhiqi and Liu, Ziwei and Zhang, Tianwei and others},
  journal={arXiv preprint arXiv:2509.20146},
  year={2025}
}

@article{chengsycophantic2025,
  title={Sycophantic AI decreases prosocial intentions and promotes dependence},
  author={Cheng, Myra and Lee, Cinoo and Khadpe, Pranav and Yu, Sunny and Han, Dyllan and Jurafsky, Dan},
  journal={Science},
  volume={391},
  number={6792},
  pages={eaec8352},
  year={2026},
  publisher={American Association for the Advancement of Science}
}

@article{batista2026rational,
  title={A rational analysis of the effects of sycophantic AI},
  author={Batista, Rafael M and Griffiths, Thomas L},
  journal={arXiv preprint arXiv:2602.14270},
  year={2026}
}

@article{bo2025invisible,
  title={Invisible Saboteurs: Sycophantic LLMs Mislead Novices in Problem-Solving Tasks},
  author={Bo, Jessica Y and Kazemitabaar, Majeed and Deng, Mengqing and Inzlicht, Michael and Anderson, Ashton},
  journal={arXiv preprint arXiv:2510.03667},
  year={2025}
}

@inproceedings{kaur2025echoes,
  title={Echoes of Agreement: Argument Driven Sycophancy in Large Language Models},
  author={Kaur, Avneet},
  booktitle={Findings of the Association for Computational Linguistics: EMNLP 2025},
  pages={22803--22812},
  year={2025}
}

@inproceedings{gao2025evaluating,
  title={Evaluating and Mitigating Sycophancy in Large Vision-Language Models},
  author={Gao, Jiayi and Zhang, Huaiwen},
  booktitle={Proceedings of the 33rd ACM International Conference on Multimedia},
  pages={11706--11715},
  year={2025}
}

@article{christian2026self,
  title={Self-Blinding and Counterfactual Self-Simulation Mitigate Biases and Sycophancy in Large Language Models},
  author={Christian, Brian and Mazor, Matan},
  journal={arXiv preprint arXiv:2601.14553},
  year={2026}
}

@inproceedings{rrv2024chaos,
  title={Chaos with keywords: Exposing large language models sycophancy to misleading keywords and evaluating defense strategies},
  author={Rrv, Aswin and Tyagi, Nemika and Uddin, Md Nayem and Varshney, Neeraj and Baral, Chitta},
  booktitle={Findings of the Association for Computational Linguistics: ACL 2024},
  pages={12717--12733},
  year={2024}
}

@misc{googletrends_sycophancy_2026,
  author       = {{Google}},
  title        = {Google Trends: "sycophancy" search interest worldwide over the past 5 years},
  year         = {2026},
  howpublished = {\url{https://trends.google.com/explore?q=sycophancy&date=today%205-y&geo=Worldwide}},
  note         = {Accessed: 2026-05-10}
}

@techreport{faverio2025teens,
  author      = {Faverio, Michelle and Sidoti, Olivia},
  title       = {Teens, Social Media and {AI} Chatbots 2025},
  institution = {Pew Research Center},
  year        = {2025},
  month       = {December},
  url         = {https://www.pewresearch.org/internet/2025/12/09/teens-social-media-and-ai-chatbots-2025/},
  note        = {Accessed: 2026-04-22}
}

@article{singh2025openai,
  title={Openai gpt-5 system card},
  author={Singh, Aaditya and Fry, Adam and Perelman, Adam and Tart, Adam and Ganesh, Adi and El-Kishky, Ahmed and McLaughlin, Aidan and Low, Aiden and Ostrow, AJ and Ananthram, Akhila and others},
  journal={arXiv preprint arXiv:2601.03267},
  year={2025}
}

@article{arvin2025check,
  title={" Check My Work?": Measuring Sycophancy in a Simulated Educational Context},
  author={Arvin, Chuck},
  journal={arXiv preprint arXiv:2506.10297},
  year={2025}
}

@inproceedings{batzner2025germanpartiesqa,
  title={GermanPartiesQA: Benchmarking Commercial Large Language Models and AI Companions for Political Alignment and Sycophancy},
  author={Batzner, Jan and Stocker, Volker and Schmid, Stefan and Kasneci, Gjergji},
  booktitle={Proceedings of the AAAI/ACM Conference on AI, Ethics, and Society},
  volume={8},
  number={1},
  pages={330--342},
  year={2025}
}

@inproceedings{beigi2025sycophancy,
  title={Sycophancy mitigation through reinforcement learning with uncertainty-aware adaptive reasoning trajectories},
  author={Beigi, Mohammad and Shen, Ying and Shojaee, Parshin and Wang, Qifan and Wang, Zichao and Reddy, Chandan K and Jin, Ming and Huang, Lifu},
  booktitle={Proceedings of the 2025 Conference on Empirical Methods in Natural Language Processing},
  pages={13090--13103},
  year={2025}
}

@article{ccelebi2025parrot,
  title={PARROT: Persuasion and Agreement Robustness Rating of Output Truth--A Sycophancy Robustness Benchmark for LLMs},
  author={{\c{C}}elebi, Yusuf and Ezerceli, {\"O}zay and Hussieni, Mahmoud El},
  journal={arXiv preprint arXiv:2511.17220},
  year={2025}
}

@article{chandra2026sycophantic,
  title={Sycophantic chatbots cause delusional spiraling, even in ideal Bayesians},
  author={Chandra, Kartik and Kleiman-Weiner, Max and Ragan-Kelley, Jonathan and Tenenbaum, Joshua B},
  journal={arXiv preprint arXiv:2602.19141},
  year={2026}
}

@article{chen2025persona,
  title={Persona vectors: Monitoring and controlling character traits in language models},
  author={Chen, Runjin and Arditi, Andy and Sleight, Henry and Evans, Owain and Lindsey, Jack},
  journal={arXiv preprint arXiv:2507.21509},
  year={2025}
}

@article{dubois2026ask,
  title={Ask don't tell: Reducing sycophancy in large language models},
  author={Dubois, Magda and Ududec, Cozmin and Summerfield, Christopher and Luettgau, Lennart},
  journal={arXiv preprint arXiv:2602.23971},
  year={2026}
}

@article{huang2025deceptionbench,
  title={Deceptionbench: A comprehensive benchmark for ai deception behaviors in real-world scenarios},
  author={Huang, Yao and Sun, Yitong and Zhang, Yichi and Zhang, Ruochen and Dong, Yinpeng and Wei, Xingxing},
  journal={arXiv preprint arXiv:2510.15501},
  year={2025}
}

@article{irpan2025consistency,
  title={Consistency Training Helps Stop Sycophancy and Jailbreaks},
  author={Irpan, Alex and Turner, Alexander Matt and Kurzeja, Mark and Elson, David K and Shah, Rohin},
  journal={arXiv preprint arXiv:2510.27062},
  year={2025}
}

@article{kelley2026,
  title={Personalization Increases Affective Alignment but Has Role-Dependent Effects on Epistemic Independence in LLMs},
  author={Kelley, Sean W and Riedl, Christoph},
  journal={arXiv preprint arXiv:2603.00024},
  year={2026}
}

@misc{openai2024spec,
  author       = {{OpenAI}},
  title        = {Model Spec},
  year         = {2024},
  month        = may,
  howpublished = {OpenAI},
  url          = {https://openai.com/index/introducing-the-model-spec/},
  note         = {Version: May 8, 2024}
}

@misc{openai2025feb,
  author       = {{OpenAI}},
  title        = {Model Spec},
  year         = {2025},
  month        = feb,
  howpublished = {OpenAI},
  url          = {https://model-spec.openai.com/2025-02-12.html},
  note         = {Version: February 12, 2025}
}

@misc{openai2025apr,
  author       = {{OpenAI}},
  title        = {Model Spec},
  year         = {2025},
  month        = apr,
  howpublished = {OpenAI},
  url          = {https://cdn.openai.com/spec/model-spec-2025-04-11.html},
  note         = {Version: April 11, 2025}
}

@misc{openai2025incident,
  author       = {{OpenAI}},
  title        = {Sycophancy in {GPT-4o}: What Happened and What We're Doing About It},
  year         = {2025},
  month        = apr,
  howpublished = {OpenAI Blog},
  url          = {https://openai.com/index/sycophancy-in-gpt-4o/}
}

@misc{openai2025sep,
  author       = {{OpenAI}},
  title        = {Model Spec},
  year         = {2025},
  month        = sep,
  howpublished = {OpenAI},
  url          = {https://cdn.openai.com/spec/model-spec-2025-09-12.html},
  note         = {Version: September 12, 2025}
}

@misc{openai2025oct,
  author       = {{OpenAI}},
  title        = {Model Spec},
  year         = {2025},
  month        = oct,
  howpublished = {OpenAI},
  url          = {https://cdn.openai.com/spec/model-spec-2025-10-27.html},
  note         = {Version: October 27, 2025}
}

@misc{openai2025dec,
  author       = {{OpenAI}},
  title        = {Model Spec},
  year         = {2025},
  month        = dec,
  howpublished = {OpenAI},
  url          = {https://cdn.openai.com/spec/model-spec-2025-12-18.html},
  note         = {Version: December 18, 2025}
}

@misc{anthropic2023spec,
  author       = {{Anthropic}},
  title        = {Claude's Constitution},
  year         = {2023},
  month        = may,
  howpublished = {Anthropic},
  url          = {https://www.anthropic.com/news/claudes-constitution},
  note         = {Version: May 9, 2023}
}

@misc{anthropic2026spec,
  author       = {{Anthropic}},
  title        = {Claude's Constitution},
  year         = {2026},
  month        = jan,
  howpublished = {Anthropic},
  url          = {https://www.anthropic.com/constitution},
  note         = {Version: January 21, 2026}
}

@article{ranaldi2023large,
  title={When large language models contradict humans? large language models' sycophantic behaviour},
  author={Ranaldi, Leonardo and Pucci, Giulia},
  journal={arXiv preprint arXiv:2311.09410},
  year={2023}
}

@inproceedings{malmqvist2025sycophancy,
  title={Sycophancy in large language models: Causes and mitigations},
  author={Malmqvist, Lars},
  booktitle={Intelligent Computing-Proceedings of the Computing Conference},
  pages={61--74},
  year={2025},
  organization={Springer}
}

@article{rehani2026social,
  title={The Social Sycophancy Scale: A psychometrically validated measure of sycophancy},
  author={Rehani, Jean and de Mello, Victoria Oldemburgo and Ovsyannikova, Dariya and Anderson, Ashton and Inzlicht, Michael},
  journal={arXiv preprint arXiv:2603.15448},
  year={2026}
}

@article{laban2023you,
  title={Are you sure? challenging llms leads to performance drops in the flipflop experiment},
  author={Laban, Philippe and Murakhovs' ka, Lidiya and Xiong, Caiming and Wu, Chien-Sheng},
  journal={arXiv preprint arXiv:2311.08596},
  year={2023}
}

@article{rahman2025pendulum,
  title={PENDULUM: A Benchmark for Assessing Sycophancy in Multimodal Large Language Models},
  author={Rahman, ABM and Anwar, Saeed and Usman, Muhammad and Ahmad, Irfan and Mian, Ajmal},
  journal={arXiv preprint arXiv:2512.19350},
  year={2025}
}

@article{ibrahim2025training,
  title={Training language models to be warm can reduce accuracy and increase sycophancy},
  author={Ibrahim, Lujain and Hafner, Franziska Sofia and Rocher, Luc},
  journal={Nature},
  volume={652},
  number={8112},
  pages={1159--1165},
  year={2026},
  publisher={Nature Publishing Group UK London}
}

@article{anvari2025defragmenting,
  title={Defragmenting psychology},
  author={Anvari, Farid and Alsalti, Taym and Oehler, Lorenz A and Hussey, Ian and Elson, Malte and Arslan, Ruben C},
  journal={Nature Human Behaviour},
  volume={9},
  number={5},
  pages={836--839},
  year={2025},
  publisher={Nature Publishing Group UK London}
}

@inproceedings{ranaldi2026learning,
  title={Learning Multilingual Agentic Policy to Control Sycophancy},
  author={Ranaldi, Leonardo and Pucci, Giulia},
  booktitle={Proceedings of the 19th Conference of the European Chapter of the Association for Computational Linguistics (Volume 1: Long Papers)},
  pages={3664--3681},
  year={2026}
}

@article{pandey2025beacon,
  title={Beacon: Single-Turn Diagnosis and Mitigation of Latent Sycophancy in Large Language Models},
  author={Pandey, Sanskar and Chopra, Ruhaan and Puniya, Angkul and Pal, Sohom},
  journal={arXiv preprint arXiv:2510.16727},
  year={2025}
}

@article{noshin2026ai,
  title={AI Sycophancy: How Users Flag and Respond},
  author={Noshin, Kazi and Ahmed, Syed Ishtiaque and Sultana, Sharifa},
  journal={arXiv preprint arXiv:2601.10467},
  year={2026}
}

@article{kim2025challenging,
  title={Challenging the Evaluator: LLM Sycophancy Under User Rebuttal},
  author={Kim, Sungwon and Khashabi, Daniel},
  journal={arXiv preprint arXiv:2509.16533},
  year={2025}
}

@article{flake2020measurement,
  title={Measurement schmeasurement: Questionable measurement practices and how to avoid them},
  author={Flake, Jessica Kay and Fried, Eiko I},
  journal={Advances in methods and practices in psychological science},
  volume={3},
  number={4},
  pages={456--465},
  year={2020},
  publisher={Sage Publications Sage CA: Los Angeles, CA}
}

@article{shi2026hallucination,
  title={From Hallucination to Scheming: A Unified Taxonomy and Benchmark Analysis for LLM Deception},
  author={Shi, Jerick and Zhang, Terry Jingcheng and Jin, Zhijing and Conitzer, Vincent},
  journal={arXiv preprint arXiv:2604.04788},
  year={2026}
}

@inproceedings{khan2024mitigating,
  title={Mitigating sycophancy in large language models via direct preference optimization},
  author={Khan, Azal Ahmad and Alam, Sayan and Wang, Xinran and Khan, Ahmad Faraz and Neog, Debanga Raj and Anwar, Ali},
  booktitle={2024 IEEE international conference on big data (BigData)},
  pages={1664--1671},
  year={2024},
  organization={IEEE}
}

@article{kran2025darkbench,
  title={Darkbench: Benchmarking dark patterns in large language models},
  author={Kran, Esben and Nguyen, Hieu Minh and Kundu, Akash and Jawhar, Sami and Park, Jinsuk and Jurewicz, Mateusz Maria and others},
  journal={arXiv preprint arXiv:2503.10728},
  year={2025}
}

@article{kwik2025digital,
  title={Digital Yes-Men: How to Deal With Sycophantic Military AI?},
  author={Kwik, Jonathan},
  journal={Global Policy},
  volume={16},
  number={3},
  pages={467--473},
  year={2025},
  publisher={Wiley Online Library}
}

@article{peng2026sycoeval,
  title={SycoEval-EM: Sycophancy Evaluation of Large Language Models in Simulated Clinical Encounters for Emergency Care},
  author={Peng, Dongshen and Wang, Yi and Schoeffler, Austin and Preiksaitis, Carl and Rose, Christian},
  journal={arXiv preprint arXiv:2601.16529},
  year={2026}
}

@inproceedings{pi2025pointing,
  title={Pointing to a llama and call it a camel: On the sycophancy of multimodal large language models},
  author={Pi, Renjie and Miao, Kehao and Peihang, Li and Liu, Runtao and Gao, Jiahui and Zhang, Jipeng and Zhou, Xiaofang},
  booktitle={Proceedings of the 2025 Conference on Empirical Methods in Natural Language Processing},
  pages={20177--20191},
  year={2025}
}

@article{rabby2026moral,
  title={Moral Sycophancy in Vision Language Models},
  author={Rabby, Shadman and Papon, Md Hefzul Hossain and Ahmed, Sabbir and Arif, Nokimul Hasan and Rahman, ABM and Ahmad, Irfan},
  journal={arXiv preprint arXiv:2602.08311},
  year={2026}
}

@article{richter2025large,
  title={Large language models outperform humans in identifying neuromyths but show sycophantic behavior in applied contexts},
  author={Richter, Eileen and Spitzer, Markus Wolfgang Hermann and Morgan, Annabelle and Frede, Luisa and Weidlich, Joshua and Moeller, Korbinian},
  journal={Trends in neuroscience and education},
  volume={39},
  pages={100255},
  year={2025},
  publisher={Elsevier}
}

@article{papadatos2024linear,
  title={Linear probe penalties reduce llm sycophancy},
  author={Papadatos, Henry and Freedman, Rachel},
  journal={arXiv preprint arXiv:2412.00967},
  year={2024}
}

@article{shapira2026rlhf,
  title={How RLHF Amplifies Sycophancy},
  author={Shapira, Itai and Benade, Gerdus and Procaccia, Ariel D},
  journal={arXiv preprint arXiv:2602.01002},
  year={2026}
}

@inproceedings{jain2026interaction,
  title={Interaction context often increases sycophancy in LLMs},
  author={Jain, Shomik and Park, Charlotte and Viana, Matt and Wilson, Ashia and Calacci, Dana},
  booktitle={Proceedings of the 2026 CHI Conference on Human Factors in Computing Systems},
  pages={1--26},
  year={2026}
}

@inproceedings{ibrahim2025towards,
  title={Towards interactive evaluations for interaction harms in human-AI systems},
  author={Ibrahim, Lujain and Huang, Saffron and Ahmad, Lama and Bhatt, Umang and Anderljung, Markus},
  booktitle={Proceedings of the AAAI/ACM Conference on AI, Ethics, and Society},
  volume={8},
  number={2},
  pages={1302--1310},
  year={2025}
}

@misc{cab2392_2026,
  title        = {AB-2392 Public Postsecondary Education: Generative Artificial Intelligence Systems: Procurement Standards and Training},
  author       = {{California State Legislature}},
  year         = {2026},
  howpublished = {\url{https://leginfo.legislature.ca.gov/}},
  note         = {Amended April 23, 2026}
}

@misc{casb1119_2026,
  title        = {SB-1119 Companion Chatbots: Children's Safety},
  author       = {{California State Legislature}},
  year         = {2026},
  howpublished = {\url{https://leginfo.legislature.ca.gov/}},
  note         = {Amended March 25, 2026}
}

@misc{ny_a10379_2026,
  title        = {A.10379 Prohibition on Unsafe Chatbot Features for Minors},
  author       = {{New York State Legislature}},
  year         = {2026},
  howpublished = {\url{https://nyassembly.gov/}},
  note         = {Introduced March 3, 2026}
}
